%% file: iclr2027_conference.tex
\documentclass{article} 
\usepackage{iclr2027_conference,times}

\input{math_commands.tex}

\usepackage{graphicx}
\usepackage{wrapfig} 
\usepackage{float} 
\usepackage{xcolor}
\usepackage{colortbl}
\definecolor{rankfirst}{HTML}{DFEDF5}
\definecolor{ranksecond}{HTML}{F8D6CC}
\definecolor{rankthird}{HTML}{CEE8E4}
\definecolor{abstractreview}{RGB}{37,94,146}
\usepackage{booktabs}
\usepackage{multirow}
\usepackage{array}
\usepackage{longtable} 
\usepackage{rotating}
\usepackage{hyperref}
\usepackage{url}

\title{\centering DISCERN: Can AI Agents Work Like Scientists and Guide Discovery?}

\author{%
\makebox[\dimexpr\textwidth-2\tabcolsep\relax][c]{%
\begin{tabular}{c}
Nan Huang$^{1}$ \enspace
Mario Tapia-Pacheco$^{1}$ \enspace
Kun Zhou$^{1}$ \enspace
Yiming Huang$^{1}$ \\
Kevin José Barrientos Díaz$^{2}$ \enspace
Tiffany Amariuta$^{1,\dagger}$ \enspace
Jingbo Shang$^{1,\dagger}$ \\[3pt]
$^{1}$University of California San Diego \quad $^{2}$Independent Researcher \\
{\small $^{\dagger}$Co-corresponding authors.} \\
{\small\texttt{\{n5huang,jshang\}@ucsd.edu}} \\[3pt]
{\normalfont\small\url{https://discern-benchmark.github.io/discern-website/}}
\end{tabular}%
}%
}

\iclrfinalcopy 

\begin{document}

\maketitle
\lhead{Preprint} 

\begingroup
\setlength{\textfloatsep}{\baselineskip}
\setlength{\intextsep}{\baselineskip}
\setlength{\floatsep}{\baselineskip}
\AddToHook{env/figure/begin}[main-figure-spacing]{\setlength{\abovecaptionskip}{\baselineskip}}
\AddToHook{env/wrapfigure/begin}[main-figure-spacing]{\setlength{\abovecaptionskip}{\baselineskip}}
\input{main_body}
\RemoveFromHook{env/figure/begin}[main-figure-spacing]
\RemoveFromHook{env/wrapfigure/begin}[main-figure-spacing]
\endgroup

\appendix
\raggedbottom

\input{appendix_profiles}
\input{appendix_diagnostics}

\section{Full Task and World Inventory}
\label{app:inventory}
Tables~\ref{tab:benchmark-inventory-a}--\ref{tab:benchmark-inventory-c} report the complete frozen
eight-track overview. Each track block lists its substrate, every canonical L1 task, every L2 analysis
family, every L3 real or counterfactual dossier, and the work that grounds the L3 design. The grounding
field distinguishes the source of a real-data substrate from prior work documenting a native
phenomenon and from prior work supporting the scientific coherence of an injected counterfactual. An
injection-design citation does not claim that the exact injected relationship was previously observed.
The substrates are the cystic-fibrosis lung-disease GWAS meta-analysis \citep{corvol2015cf}; ClinVar
\citep{landrum2018clinvar} interpreted under ACMG/AMP criteria \citep{richards2015acmg} with gnomAD allele
frequencies \citep{karczewski2020gnomad}; COVID-19 PBMC bulk RNA-seq \citep{arunachalam2020covid}; the 10x
Genomics PBMC scRNA-seq dataset \citep{zheng2017pbmc}; the HAP1 genome-wide knockout screen
\citep{bertomeu2018crispr} with core-essential reference sets \citep{hart2017ceg}; the primate enhancer MPRA
\citep{klein2018enhancer}; ENCODE K562 ChIP-seq and RNA-seq \citep{encode2012}; and GB1 stability
measurements \citep{tsuboyama2023stability}. Analysis tools named in the registries include PyDESeq2
\citep{muzellec2023pydeseq2}, which implements the DESeq2 method \citep{love2014deseq2}, MAGeCK \citep{li2014mageck}, BAGEL \citep{hart2016bagel}, SoupX
\citep{young2020soupx}, CellBender \citep{fleming2023cellbender}, and ESM-2 \citep{lin2023esm2}.
The two-track physics extension is reported separately in Appendix~\ref{app:physics}; it is not part of
the current 203-task denominator or the molecular-life-science core score.

\begin{table}[H]
  \caption{Complete DISCERN track inventory, part 1 of 3: statistical and clinical genetics,
  transcriptomics, and cellular genomics.}
  \label{tab:benchmark-inventory-a}
  \centering
  \includegraphics[width=\linewidth]{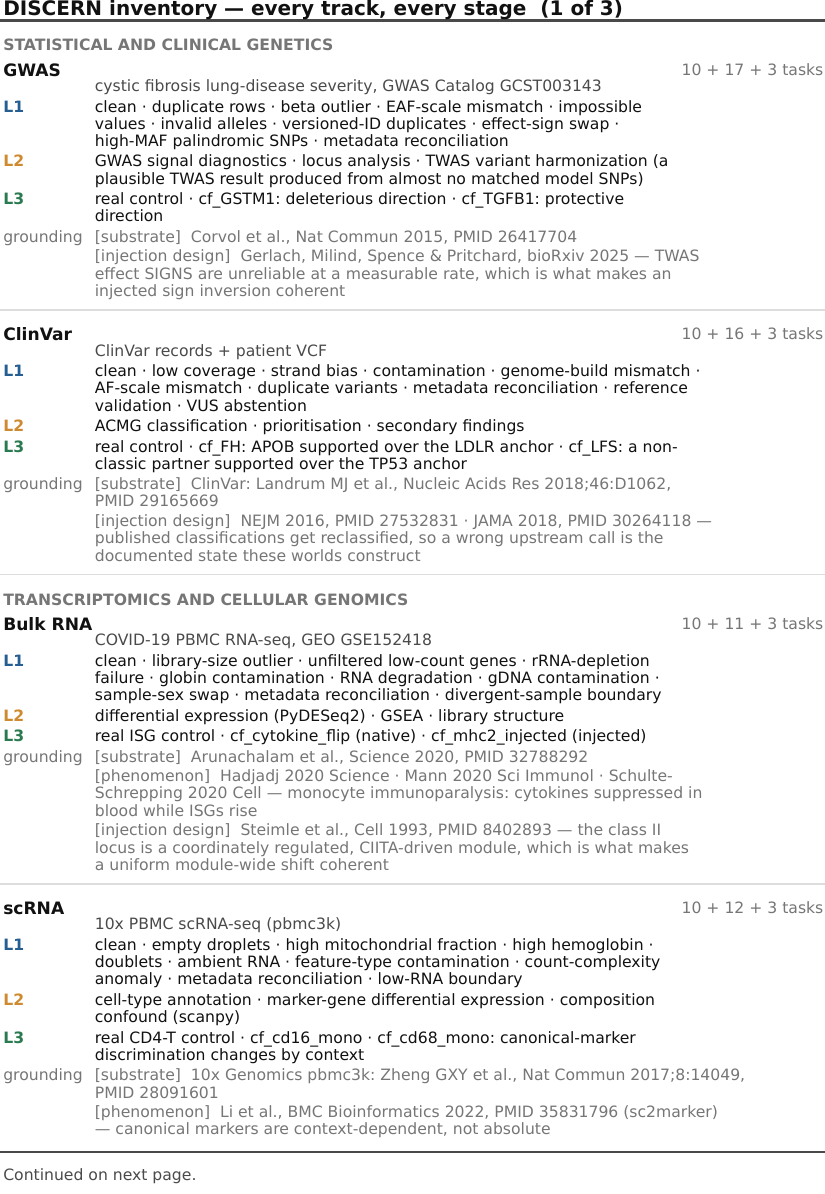}
\end{table}

\begin{table}[H]
  \caption{Complete DISCERN track inventory, part 2 of 3: functional and regulatory genomics.}
  \label{tab:benchmark-inventory-b}
  \centering
  \includegraphics[width=\linewidth]{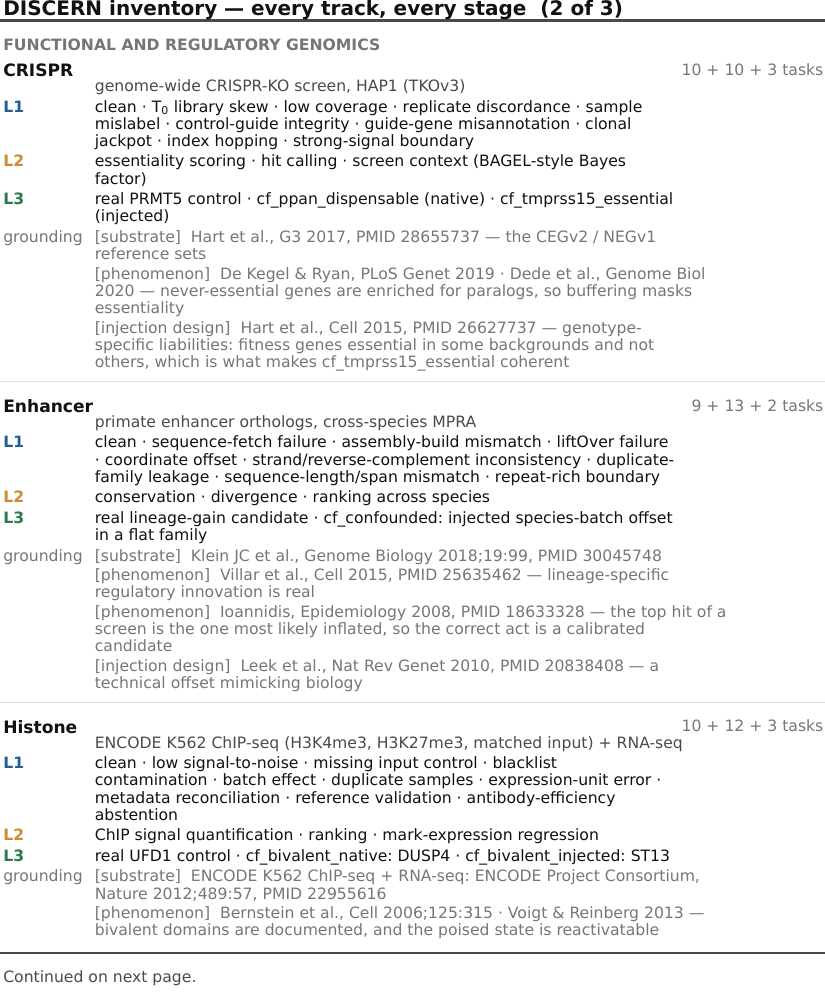}
\end{table}

\begin{table}[H]
  \caption{Complete DISCERN track inventory, part 3 of 3: protein biophysics. The footer reports the
  frozen benchmark totals.}
  \label{tab:benchmark-inventory-c}
  \centering
  \includegraphics[width=\linewidth]{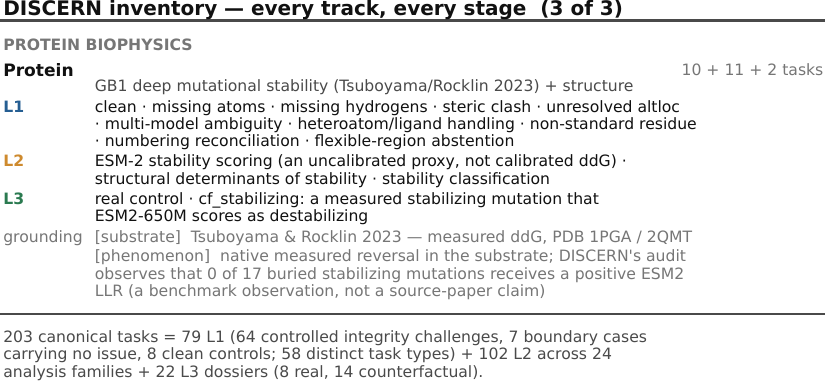}
\end{table}

\subsection{Complete L1 integrity registry}
\label{app:l1-registry}
Table~\ref{tab:l1-registry} lists all 79 L1 worlds, grouped by track, with
the class and expected response extracted from the world manifests. Controls and challenges retain
their manifest-defined expected responses, allowing appropriate acceptance to be distinguished from
reflexive rejection. The 71 non-clean L1 worlds carry 65 distinct defect class names,
and only three classes recur across tracks.

\input{appendix_registry_l1}

\subsection{Complete L2 analysis registry}
\label{app:l2-registry}
Table~\ref{tab:l2-registry} lists all 102 L2 worlds within their 24
analysis families. The common family method is stated once, while each world records the distinct
confound, tool- or procedure-output trap, statistical trap, or clean control and its expected
response. Missing manifest fields remain explicitly missing rather than being reconstructed.

\input{appendix_registry_l2}

\subsection{Complete L3 discovery registry}
\label{app:l3-registry}
Table~\ref{tab:l3-world-registry} lists all 22 L3 dossiers. For each
world they report its route, supplied evidence, elicited prior, expected act, grounding, and number of
models scored. Native counterfactuals preserve a surprising relationship present in the real
substrate; injected counterfactuals modify an upstream quantity and rerun the real scorer. These two
routes therefore carry different citation burdens.

\input{appendix_registry_l3}

\paragraph{Grounding qualifications.}
A construct-level citation makes an intervention scientifically plausible; it does not establish the counterfactual entity-level result as true.
The scRNA grounding is Li et al.'s \texttt{sc2marker}, not ``Kohl 2022.'' The enhancer substrate is
Klein et al., \emph{Genome Biology} 19:99, which assayed 348 active tiles from 233 putative enhancers
across 11 primates; any different counts in DISCERN refer to the processed benchmark subset. For the
protein case, \citet{tsuboyama2023stability} grounds the measured stability substrate but not a general
claim of ``ESM stabilizing blindness.'' These distinctions are retained as limitations rather than
filled with citations that support only an adjacent claim.

Source and construct references include documented TWAS false-sign risk \citep{gerlach2025twas}, clinical misclassification and reclassification \citep{manrai2016misdiagnoses,mersch2018reclassification}, context-specific marker discrimination \citep{li2022sc2marker}, HAP1 screening and contextual essentiality \citep{bertomeu2018crispr}, cross-primate STARR-seq \citep{klein2018enhancer}, bivalent-domain biology \citep{bernstein2006bivalent}, and dysregulated interferon/inflammation \citep{hadjadj2020covid}. The controlled TMPRSS15 direction lacks entity-level precedent.

\subsection{Design constraints}
Table~\ref{tab:design-constraints} links each benchmark principle to its concrete implementation and
to the failure that the constraint prevents. In particular, controls prevent indiscriminate
suspicion, while tool- and procedure-output traps prevent successful execution from being mistaken
for valid analysis.

\begin{table}[H]
  \caption{DISCERN design constraints: principle, implementation, and failure prevented. Counts are
  derived from the frozen task registry.}
  \label{tab:design-constraints}
  \centering
  \includegraphics[width=\linewidth]{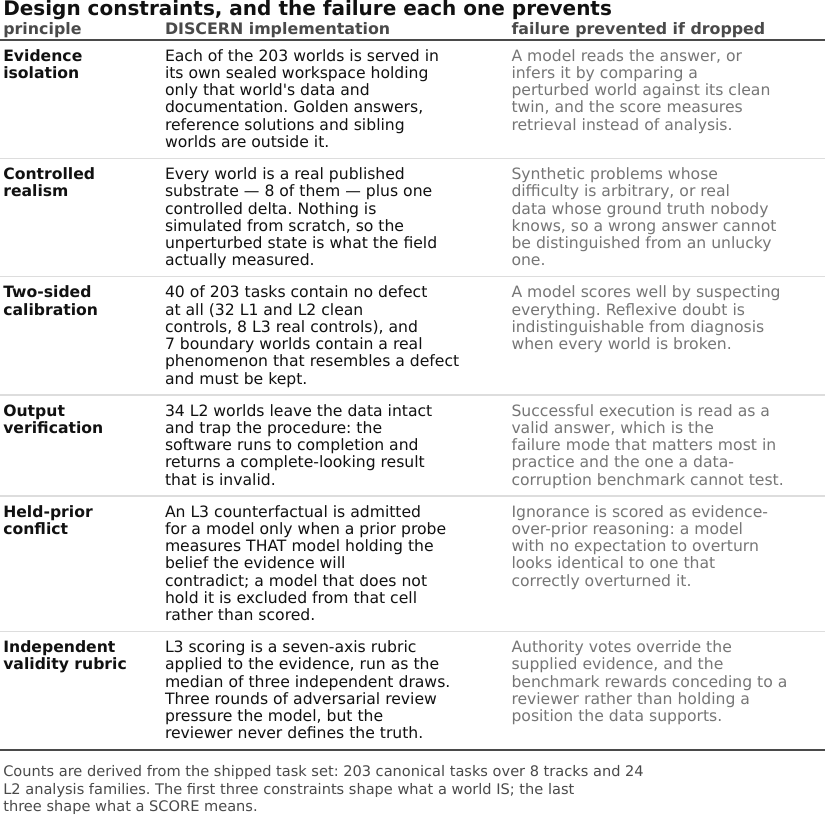}
\end{table}

\section{Initial Cross-Domain Physics Extension}
\label{app:physics}

DISCERN instantiates the same L1--L3 contract in calorimetry and nuclear decay as an
initial transfer probe beyond molecular life science. These two tracks are reported separately from the eight-track core and do not enter its 203-task denominator or score. 

\paragraph{Calorimetry.}

The calorimetry track uses real adiabatic-calorimetry measurements of temperature-dependent heat capacity, including crystalline Cu measurements from \citep{lang2006adiabatic}, with phase-resolved n-heptane thermodynamic measurements from \citep{bissengaliyeva2011heat} added for the equilibrium analysis. L1 tests whether agents preserve valid heat-capacity structure while detecting defects such as duplicated or missing measurements, unit mismatches, temperature--heat-capacity misalignment, and uncertainty-scale errors. L2 carries the Cu heat-capacity substrate into an adiabatic Cu--heptane equilibrium problem and tests whether agents detect incorrect enthalpy references, temperature-dependent enthalpy drift, and phase-blind interpolation across the heptane melting transition. L3 retains the frozen L2 equilibrium model and asks whether the bulk-Cu heat-capacity prior remains adequate for follow-up nanocrystalline Cu specimen states. The two counterfactual worlds introduce specimen-level heat-capacity increases and regenerate the downstream equilibrium outcomes rather than editing them directly.

The nanocrystalline-Cu counterfactuals are grounded in heat-capacity measurements by
\citet{rupp1987heat} and \citet{tan2009heat}.

\paragraph{Nuclear decay.}

The nuclear-decay track uses a public multi-year radiation-measurement dataset from \citep{goddard2020radiation}, which contains repeated 2048-channel NaI(Tl) Co-60 spectra, acquisition metadata, an independent Mn-54 source, and synchronized environmental measurements. L1 tests whether agents can identify acquisition- and spectrum-level integrity problems, including duplicated or missing measurements, timestamp and live-time errors, inconsistent count summaries, and spectral shifts, while avoiding false alarms when legitimate calibration changes occur. L2 aggregates a validated portion of these same Co-60 measurements into 274 daily observations and tests whether agents can recover the decay rate despite misleading analysis choices or confounds, including an inappropriate detector-efficiency model, Mn-54 contamination, and an unstable fixed-channel photopeak analysis. L3 carries forward this frozen 274-day dataset. The real world retains the observed Co-60 decay pattern, while two counterfactual worlds assume fully ionized Co-60 under stellar conditions, apply literature-grounded changes from \citet{gupta2023decay}. to the effective decay rate, and regenerate the corresponding count trajectories while keeping the observation schedule and control measurements fixed.

\begin{table}[H]
\centering
\small
\caption{Frozen Physics-extension inventory. Boundary tasks contain no injected defect and test whether agents avoid overinterpreting valid physical behavior. L3 counterfactuals are evaluated only after confirming the relevant pre-task prior. The auxiliary detector calibration follows \citet{tarim2018efficiency}; stellar-decay counterfactuals follow \citet{gupta2023decay}.}
\label{tab:physics-inventory}

\begin{tabular}{llcl}
\hline
Track & Level & $n$ & Control and challenge structure \\
\hline

Calorimetry
& L1 & 8 & 1 clean; 6 injected integrity defects; 1 no-defect boundary \\
& L2 & 4 & 1 clean; 2 data confounds; 1 procedure/analysis trap \\
& L3 & 3 & 1 real; 2 prior-gated injected counterfactuals \\

Nuclear decay
& L1 & 8 & 1 clean; 6 injected integrity defects; 1 no-defect boundary \\
& L2 & 4 & 1 clean; 1 data confound; 2 procedure/analysis traps \\
& L3 & 3 & 1 real; 2 prior-gated injected counterfactuals \\
\hline

\textbf{Total} & & \textbf{30} & 16 L1; 8 L2; 6 L3 \\
\hline
\end{tabular}
\end{table}

\paragraph{Frozen task inventory.}

Table~\ref{tab:physics-inventory} summarizes the Physics tasks using the same control-versus-challenge structure as the biology registries. At L1, each track includes one clean task, six injected integrity challenges, and one no-defect boundary task that tests whether agents avoid treating legitimate physical behavior as an error. At L2, each track includes one clean analysis and three challenged analyses involving either a data confound or a misleading analysis procedure. At L3, each track includes one real/control world and two counterfactual worlds. Before evaluating a counterfactual, we first confirm that the model holds the scientific expectation that the new evidence is intended to challenge. All eight fully evaluated models passed these prior checks, so every Physics L3 world was included in the analysis.

\begin{table}[H]
\centering
\small
\caption{Physics-extension performance on the 0--100 scale. All eight models have complete coverage: $n=16$ L1, $n=8$ L2, and $n=6$ eligible L3 tasks per model.}
\label{tab:physics-results}

\begin{tabular}{lrrr}
\hline
Model & L1 & L2 & L3 \\
\hline

GPT-6 Astra & 70.8 & 79.2 & 72.4 \\
GPT-5.6-sol & 85.4 & 70.8 & 75.9 \\
Claude Opus 5 & 77.1 & 54.2 & 69.2 \\
Kimi K3 & 81.3 & 66.7 & 76.7 \\
GLM-5.3 & 79.2 & 54.2 & 60.9 \\
GLM-5.2 & 77.1 & 58.3 & 74.7 \\
DeepSeek V4 Pro & 75.0 & 62.5 & 72.7 \\
GPT-4.1 & 22.9 & 0.0 & 17.6 \\
\hline

Mean & 71.1 & 55.7 & 65.0 \\
\hline
\end{tabular}
\end{table}

\paragraph{Physics-extension results.}

Table~\ref{tab:physics-results} reports the finalized stage scores for the same eight fully evaluated models used in the life science tracks. Scores average the two Physics tracks and are linearly rescaled from 0--3 to 0--100. Each model has 16 eligible L1 tasks, 8 L2 tasks, and 6 L3 tasks; across eight models, the corresponding denominators are 128, 64, and 48 model--task evaluations. The model leading the extension changes by stage: Sol has the highest Physics L1 mean, Astra the highest L2 mean, and Kimi the highest L3 mean.

\paragraph{Comparison with the life-science core.}

The Physics extension suggests that several DISCERN behaviors transfer beyond the life-science tasks, while others depend on scientific context. Across the same eight models, mean Physics performance was 71.1 at L1, 55.7 at L2, and 65.0 at L3, compared with 74.1, 68.4, and 64.8 across the eight life-science tracks. Thus, aggregate L1 and L3 performance was similar across the two settings, while L2 performance was lower in Physics, suggesting the models had a harder time verifying the validity of the physical analyses. Model-level variation was also consistent: no model led all three Physics stages, with Sol highest at L1, Astra at L2, and Kimi at L3. Some model behaviors transferred across domains. Kimi, which frequently favored counter-prior evidence in the life-science tasks, also demonstrated that behavior in the Nuclear counterfactual worlds and achieved the highest Physics L3 mean. Other behaviors were less stable. Astra's strong tendency to abstain in life-science L3 did not extend uniformly to Physics: in the Nuclear counterfactuals, it accepted the evidence-supported change in Co-60 decay behavior while remaining cautious about its causal interpretation. Opus likewise moved from the highest life-science L2 mean to substantially lower Physics L2 performance. These differences suggest that model behavior depends on the scientific context rather than each model having one fixed working style.

\section{Complete L1 and L2 Results}
\label{app:l1-l2}
\begin{figure}[H]
  \centering
  \includegraphics[width=\linewidth]{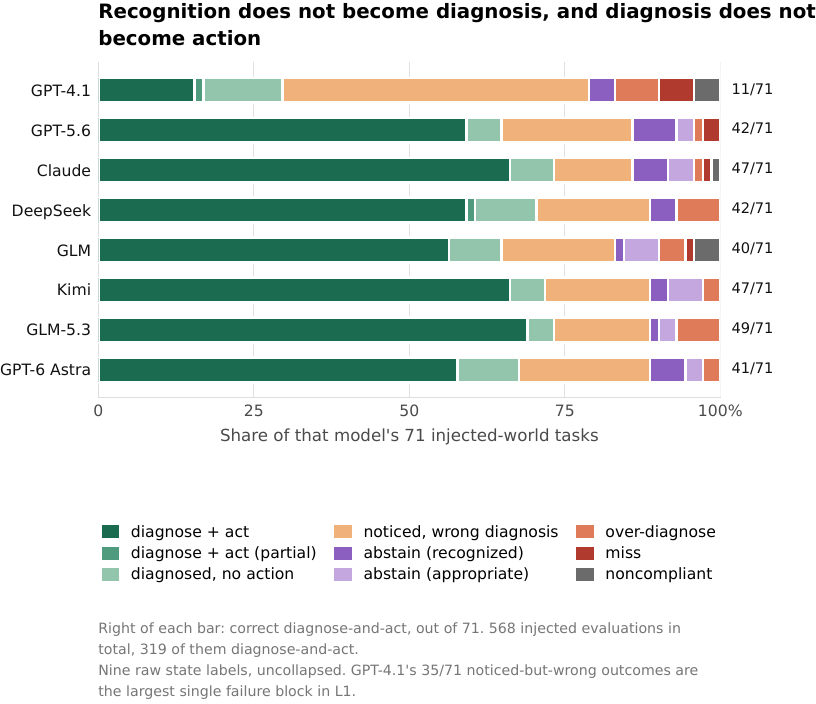}
  \caption{L1 integrity challenges across eight models: recorded behavioral states on 71 challenged
  tasks per model, separating complete handling, partial responses, wrong diagnoses, abstentions,
  and other failures. All 568 evaluations are retained.}
  \label{fig:app-injected-process}
\end{figure}

\begin{figure}[H]
  \centering
  \includegraphics{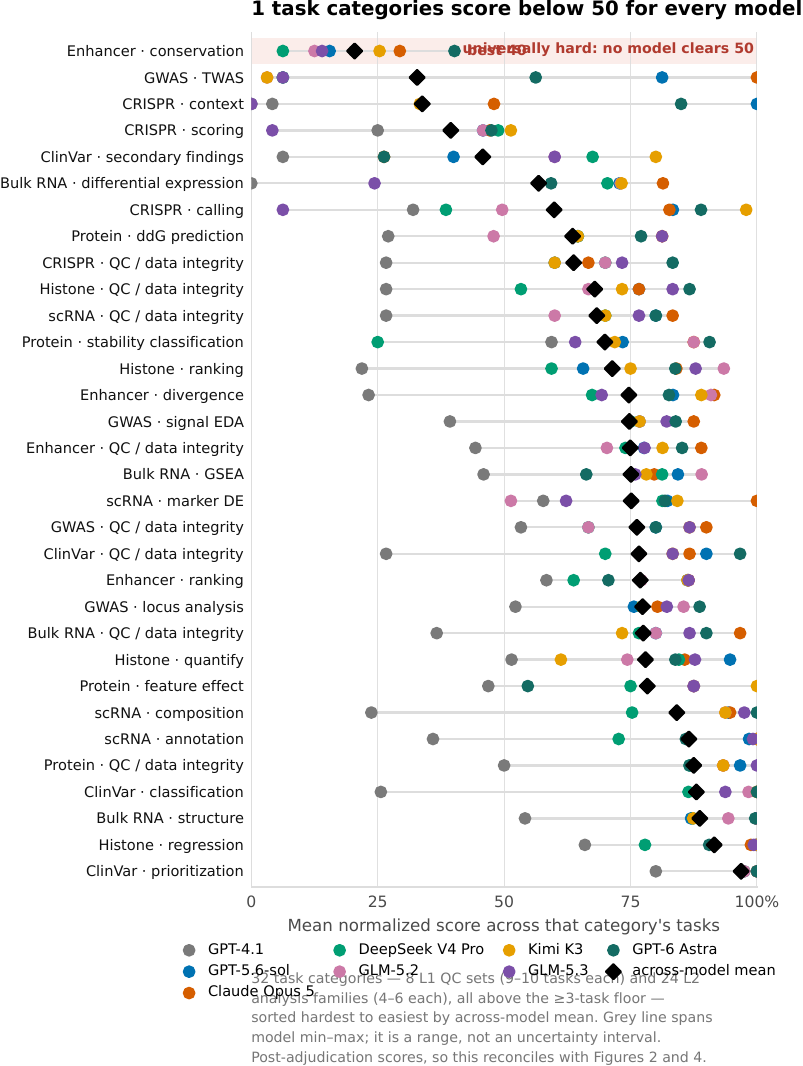}
  \caption{Task-category difficulty across all L1 integrity sets and L2 analysis families.}
  \label{fig:app-task-difficulty}
\end{figure}

Figure~\ref{fig:l1-handling}A in the main Results reports clean-control decisions;
Figure~\ref{fig:app-injected-process} below reports challenged-task outcomes.
This appendix reports task-level difficulty across all 32 L1 and L2 categories (Figure~\ref{fig:app-task-difficulty}). All
panels use the recorded L1 states and current L2 scores under the shared mapping.

\paragraph{Detailed L1 handling outcomes.}
Across challenged L1 evaluations, 56.2\% achieve full diagnosis and appropriate action,
21.7\% identify the wrong cause, 7.9\% diagnose the problem correctly without acting, and
1.4\% miss the problem entirely. Opus reaches full diagnosis and action in 66.2\% of challenged
evaluations compared with 57.7\% for Astra. Abstention rates are 9.9\% and 8.5\%,
respectively, although these include both appropriate and incompletely justified abstentions.

Task difficulty also varies substantially across scientific settings. All models perfectly handle
seven L1 challenges, including duplicate records, scale errors, and fetch failures. In contrast,
wrong diagnoses occur in 48.4--51.6\% of detectable-defect evaluations in CRISPR,
single-cell RNA sequencing, and epigenomic tracks, compared with 4.7--12.5\% across the
remaining five tracks. All ten lowest-scoring defect tasks occur in these three settings. Eight
additional challenges require withholding a defect claim or recognizing insufficient evidence;
these average 53.6, with 37.5\% of their 64 evaluations overdiagnosing a problem.

\paragraph{Detailed recognition--calibration gaps.}
Result calibration averages 64.5 across agents and is the lowest-scoring of the four shared L2
dimensions for seven of eight models. Among 425 challenged L2 evaluations receiving full
issue-recognition credit, 126 (29.6\%) receive exactly half calibration credit. This proportion
ranges from 19.4\% for Opus to 37.8\% for DeepSeek and is 47.2\% for statistical traps,
34.1\% for data confounds, and 20.6\% for tool traps. Because calibration can reflect the
finalized handling of the analysis rather than conclusion wording alone, half credit does not
uniquely identify overconfidence. These outcomes distinguish recognizing a limitation from
carrying it through methodological handling and the final conclusion.

\begin{figure}[H]
  \centering
  \includegraphics[width=\linewidth]{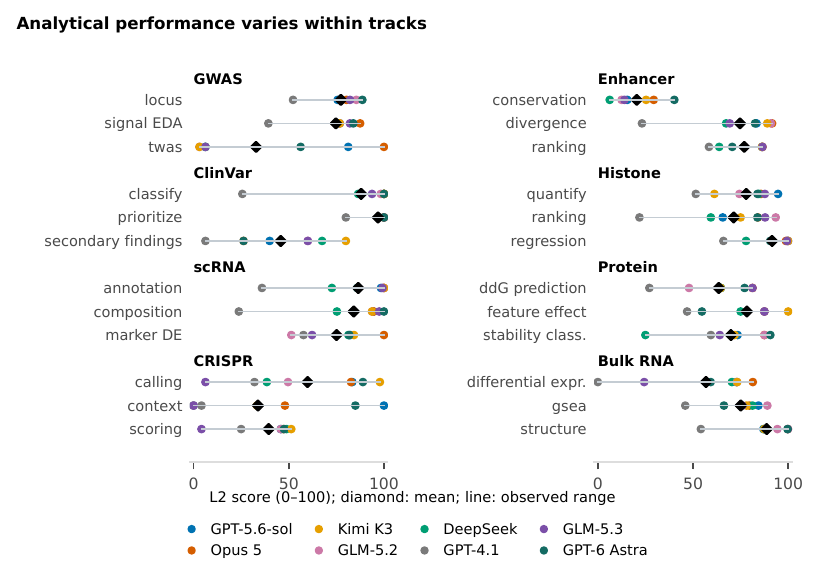}
  \caption{Model scores and across-model means for all 24 L2 analysis families under the final shared meta-rubric, normalized to 0--100. Two columns retain all eight tracks. Diamonds show model means and lines show observed model ranges.}
  \label{fig:l2-families}
\end{figure}

\section{Complete L3 Results and Cases}
\label{app:l3}

\subsection{Commitment, conditional novelty, and yield}
For GPT-6 Astra, two L3 worlds are excluded for prior eligibility, leaving 20 eligible outcomes. One contains a
hypothesis and 19 are abstentions. These exclusions concern the world-specific eligibility rule;
they do not independently establish a general tendency to avoid stating beliefs.
For eligible outcomes $E_m$ and committed outcomes $C_m$, define
\begin{equation}
p_m=\frac{|C_m|}{|E_m|},\qquad
\overline N_m=\frac{100}{3|C_m|}\sum_{w\in C_m}N_{mw},\qquad
Y_m=p_m\overline N_m=\frac{100}{3|E_m|}\sum_{w\in C_m}N_{mw}.
\end{equation}
Conditional novelty $\overline N_m$ is undefined when no hypothesis is produced; yield $Y_m$
is then zero. Abstentions contribute no produced novelty credit, without being assigned an
unobserved novelty grade. These diagnostics weight eligible worlds equally. In contrast, the
headline L3 score weights tracks equally and includes partial credit for abstention, capped at
2.5 in native rubric units. Yield measures novelty credit under the benchmark rubric, not
experimentally validated usefulness or a discovery rate.

\begin{table}[H]
\centering
\footnotesize
\setlength{\tabcolsep}{4pt}
\caption{L3 score and hypothesis output. Scores and novelty use 0--100; commitment is a percentage.
The headline is track-balanced; the remaining columns use eligible worlds. Fable is excluded
because its L3 coverage is incomplete. Conditional quality based on one Astra hypothesis should
not be interpreted as a reliable estimate of typical hypothesis quality.}
\label{tab:l3-extension}
\begin{tabular}{lrrrrr}
\toprule
Model & L3 & Committed/eligible & Commit (\%) & $\overline N$ & $Y$ \\
\midrule
GPT-6 Astra & 72.8 & 1/20 & 5.0 & 66.7 & 3.3 \\
GLM-5.3 & 71.1 & 15/22 & 68.2 & 40.0 & 27.3 \\
GPT-5.6-sol & 70.0 & 4/22 & 18.2 & 41.7 & 7.6 \\
Kimi K3 & 69.1 & 16/21 & 76.2 & 31.2 & 23.8 \\
GLM-5.2 & 63.1 & 14/22 & 63.6 & 28.6 & 18.2 \\
Claude Opus 5 & 63.0 & 17/21 & 81.0 & 41.2 & 33.3 \\
DeepSeek V4 Pro & 58.4 & 12/22 & 54.5 & 16.7 & 9.1 \\
GPT-4.1 & 51.0 & 15/22 & 68.2 & 13.3 & 9.1 \\
\bottomrule
\end{tabular}
\end{table}

Figure~\ref{fig:prior-to-yield}B,C shows these diagnostics for all eight models; Appendix
Figure~\ref{fig:l3-world-novelty} shows the world-level novelty matrix. Astra's abstention classification does not establish that it rejected
the supplied evidence. Its conditional novelty estimate is based on one hypothesis.

\paragraph{Detailed commitment and novelty outcomes.}
Across eligible L3 tasks, hypothesis commitment ranges from 5.0\% for Astra to 81.0\%
for Opus. Commitment is 58.3\% on counterfactual tasks and 48.4\% on real-data tasks,
although differences between the task worlds prevent attributing this contrast to conflicting
evidence alone. In 14 of Astra's 19 abstentions, the response explains why the available evidence
does not yet justify a hypothesis rather than simply declaring the evidence insufficient.

Among 94 submitted hypotheses, mean novelty is 30.1, below causal calibration (62.4) and
the remaining scientific-reasoning dimensions (all at least 73.4). Six of 172 eligible outcomes
receive maximum novelty credit, four of them in the DUSP4 task. To summarize both production
frequency and conditional novelty, novelty yield is defined as
\[
P(\mathrm{commit}) \times \mathbb{E}[N_{100}\mid\mathrm{commit}].
\]
Sol and Opus have similar conditional novelty scores (41.7 and 41.2) but novelty yields of
7.6 and 33.3 because their commitment rates differ substantially.

\paragraph{Critique and concession outcomes.}
Across all committed outcomes, agents concede at least one reviewer point in 20.2\% of cases,
with 84.2\% of first concessions occurring in the first review round. Model-level concession
rates range from 0\% for Sol and Astra to 41.7\% for DeepSeek. A recorded concession does not
necessarily imply that the final hypothesis is better supported, just as persistence does not
necessarily indicate resistance to valid criticism. Complete round-level outcomes and case
studies, including the DUSP4 example summarized in the main text, are reported here.

\subsection{Highest-scoring model--world outcomes}
Table~\ref{tab:l3-top-outcomes} reports the top of the frozen L3 distribution. Because three outcomes
tie at 2.60, selecting exactly five would break a tie arbitrarily; we therefore show the top three and
all three outcomes tied for fourth. Here $D$ is evidence fidelity, $Q$ is scientific reasoning, and
$N$ is valid novelty.

\begin{table}[H]
  \caption{Highest-scoring aggregate-eligible L3 model--world outcomes, shown here on the native 0--3 rubric scale.}
  \label{tab:l3-top-outcomes}
  \centering
  \small
  \setlength{\tabcolsep}{3.5pt}
  \begin{tabular}{rccccp{1.65in}l}
    \toprule
    Rank & Composite & $D$ & $Q$ & $N$ & Track and world & Model \\
    \midrule
    1 & 3.00 & 3.0 & 3.0 & 3 & Protein: \texttt{cf\_stabilizing} & Claude Opus 5 \\
    2 & 2.87 & 3.0 & 2.6 & 3 & Histone: \texttt{cf\_bivalent\_native} & DeepSeek V4 Pro \\
    3 & 2.80 & 3.0 & 2.4 & 3 & Histone: \texttt{cf\_bivalent\_native} & Kimi K3 \\
    4= & 2.60 & 3.0 & 2.8 & 2 & Bulk RNA: \texttt{cf\_mhc2\_injected} & Claude Opus 5 \\
    4= & 2.60 & 3.0 & 2.8 & 2 & Histone: \texttt{cf\_bivalent\_injected} & Claude Opus 5 \\
    4= & 2.60 & 3.0 & 2.8 & 2 & scRNA: \texttt{cf\_cd68\_mono} & Claude Opus 5 \\
    \bottomrule
  \end{tabular}
\end{table}

The two highest-scoring non-protein outcomes arise from the same native DUSP4 world but take distinct
routes. DeepSeek V4 Pro narrows the result to a provisional threshold-defined state and proposes
locus-specific H3K27me3 measurement followed by an EZH2/PRC2 perturbation time course. Kimi K3
explicitly withdraws an initial uniqueness claim, then re-establishes DUSP4 as rank 1 in H3K27me3 among
1{,}021 strong-H3K4me3 genes. It further distinguishes a categorical repressed tail from a weak graded
relationship and proposes sequential ChIP to test same-locus co-occupancy. Together, the pair shows
that one evidence-rich case can support different valid hypotheses while review forces both toward a
bounded claim and a discriminating experiment. The tied ST13 and MHC-II cases show similar discipline
on injected evidence, but DUSP4 is the stronger flagship because its relationship is native.

\begin{table}[H]
\caption{Paraphrased DUSP4 revision by DeepSeek, complementing Section~\ref{sec:hypothesis-results}.}
\label{tab:dusp4-revision}
\centering
\begin{tabular}{p{1.0in}p{4.1in}}
\toprule
Step & Recorded reasoning \\
\midrule
Initial claim & Repression overrides the activation-associated signal. \\
Limitation & Signals measured over different-sized regions do not establish that they occur together. \\
Revised claim & Retain the observed association, but treat the proposed mechanism as an explanation to test. \\
Next test & Check where the repressive signal occurs and whether reducing repression raises gene expression over time. \\
\bottomrule
\end{tabular}
\end{table}

\paragraph{Eight-model response diagnostics.}
Among 94 committed outcomes, 29 receive novelty credit of at least two-thirds of the maximum
(30.9\% of committed responses, 16.9\% of 172 eligible outcomes). This threshold can credit a
strong hypothesis or a useful next test and does not identify 29 independently validated experiments.
It is reached by 25/63 committed counterfactual responses (39.7\%) and 4/31 committed real-world
responses (12.9\%), whose mean novelty scores are 36.5 and 17.2. All six maximum-novelty responses
are counterfactual, and four concern DUSP4, which averages 88.2 composite credit across seven
eligible models. These are task-weighted descriptive comparisons, not causal effects of world design.

Of 94 hypothesis-producing outcomes, 19 record a concession, 16 first in round one. The log records
the first round with a concession flag, not necessarily abandonment of the hypothesis.
The exported reviewer rejection fraction describes the final recorded review round. At least one
reviewer rejects in 18 cases; a majority rejects in ten. These final-round judgments cannot establish
that rejection caused an earlier concession. The \texttt{hyp\_drift} field is one minus the
\texttt{SequenceMatcher} similarity between initial and final hypothesis text. Its median of 0.94
therefore indicates textual difference, not stability, and does not measure semantic narrowing.

\subsection{Abstention is not one behaviour}
\begin{figure}[H]
  \centering
  \includegraphics{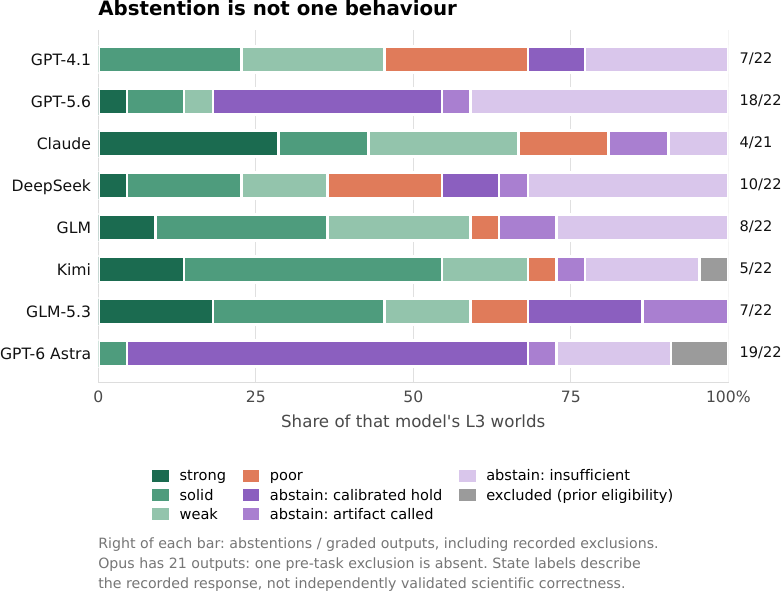}
  \caption{Recorded final L3 states per model, decomposing the commit/abstain split used in
  Section~\ref{sec:l3-results}. Bars include three recorded eligibility exclusions;
  Opus's absent pre-execution exclusion is omitted, leaving 21 graded outputs for that model.}
  \label{fig:app-l3-states}
\end{figure}

Figure~\ref{fig:app-l3-states} decomposes the binary commit/abstain split used in
Section~\ref{sec:l3-results} into the recorded final states; treating them as one category would
overstate how uniformly models decline.

\section{Prompts, Tools, and Execution Environments}
\label{app:protocol}

\subsection{Reporting contract and budgets}
Within an analysis family, every world receives the same prompt and the same structured reporting
schema; only the world directory differs. The schema is defect-agnostic: clean and perturbed worlds
request identical output fields, so the form of the question does not reveal whether an issue is
present. L1 and L2 run in a sealed Python sandbox with the permitted data, documentation, and
domain tools on \texttt{PATH}, a 20-step execution budget, and no network access. L3 supplies the
evidence dossier, allows three to six agent steps, and then runs three adversarial review rounds
with a reviewer ensemble that is fixed across all evaluated models. The harness records every
executed code cell, tool call, and token meter reading. Complete verbatim prompts, tool inventories,
and per-track sandbox manifests ship with the released registries.
Per-task workspaces use a scrubbed environment and read-only software mounts. Gold files, graders,
and sibling worlds are outside the task mounts. These restrictions control runtime access,
not prior exposure to public source data.

\paragraph{Later-model generation budgets.}
\label{app:glm53}
\label{app:astra}
GLM-5.3 was evaluated later and was not strictly budget-matched to the earlier evaluations.
After hidden-reasoning truncation was diagnosed, the per-call generation ceiling was raised
from 8{,}192 to 32{,}768 tokens and selected affected tasks were rerun under the approved run protocol.
GPT-6 Astra's generation ceiling recorded by preflight is 32{,}768 output tokens per call;
its evaluation is likewise not strictly budget-matched to the earlier evaluations.
Both models complete all 203 tasks (79 L1, 102 L2, and 22 L3), and all 40 run-level graded
files per model have corresponding verifier files.

\subsection{Models}
The eight models were accessed through first-party APIs: OpenAI (\texttt{gpt-4.1},
\texttt{gpt-5.6-sol}, \texttt{gpt-6-astra}), Anthropic (\texttt{claude-opus-5}), DeepSeek (\texttt{deepseek-v4-pro}),
Zhipu BigModel (\texttt{glm-5.2}, \texttt{glm-5.3}), and Moonshot (\texttt{kimi-k3}), under the identifiers used
throughout the paper. All eight tracks were frozen by 2026-08-22; per-run timestamps and
harness-version records are preserved in the run manifests.

\subsection{A complete worked task}
\label{app:worked-task}
One task, end to end, from the frozen artifacts: the Histone track's L2 \emph{quantify} family,
world \texttt{broad\_mark\_narrow\_default} (a tool or procedure-output trap). The agent prompt is
shared by all four worlds of the family (abridged; paths shortened):

\begin{quote}\small\ttfamily
You are a computational epigenomics researcher. GOAL: quantify the per-gene promoter-level signal of
the TARGET histone mark from the raw ChIP-seq alignments, and determine how the mark relates to gene
expression. Read meta.json for the target mark and file layout. [\ldots] World directory: meta.json
(target mark + layout), genes.csv (gene\_id, chrom, start, end, tss, gene\_expression), and
alignments/ holding the mark's ChIP BAM AND the matched input-control BAM (both indexed;
chromosome-subset of real ENCODE K562 data). Python sandbox: call run\_python(code). [\ldots]
peak-calling and coverage tools are installed and on PATH (macs3, deepTools, samtools) --- choose
your approach yourself and VERIFY the output is sound before trusting it. [\ldots] Return ONLY this
JSON: \{output\_path, method \{tool, normalization, peak\_mode, window\},
signal\_expression\_spearman, direction, domains \{n, total\_mb\}, issues\_found, analysis\_usable,
trajectory\}
\end{quote}

\textbf{The controlled delta.} In this world, \texttt{meta.json} sets the target mark to the broad
repressive mark H3K27me3 and requests the domain landscape. The trap is a tool default: MACS3's
narrow peak-calling mode completes without error on this input but fragments broad domains, returning
a 3.4\,Mb landscape where the reference solver's broad-mode golden totals 6.4\,Mb over 19{,}352
domains. Both pipelines finish and produce a plausible table, so the defect is silent.

\textbf{The recorded rubric.} The manifest stores the grading rule with the world: the per-gene CSV
is value-checked against the solver golden; declaring a solvable world unusable caps the score at 1;
for this world, 3 requires a broad or domain-aware mode, a total within $\pm35\%$ of the 6.4\,Mb
golden, and the narrow-default pitfall named; a narrow, unqualified landscape scores 0.

\textbf{Two recorded outcomes.} GPT-5.6-sol called domains with MACS3 in broad mode, corrected the
effective genome size to the chromosome panel, reported 19{,}352 domains totalling 6.43\,Mb, and
named the narrow-default pitfall; the deterministic value check awarded the top band, and human
adjudication of the trajectory lowered result calibration, leaving frozen components $A=0.99$,
$I=1.0$, $M=1.0$, $C=0.5$ and a world score of 87.1. Claude Opus 5 also chose broad mode and named
the pitfall, but its reported landscape totals 28.21\,Mb against the 6.4\,Mb golden, so analytical
correctness is 0 and the world score is 50.0 despite full issue-recognition and
methodological-response credit --- the accuracy cap of Section~\ref{sec:grading} in action:
recognizing the trap cannot compensate for computing the wrong landscape.

\section{Grading, Verification, and Adjudication}
\label{app:grading}
\input{grading_dimensions}

\paragraph{Shared evidence in L2 dimensions.}
Analytical correctness and methodological response receive identical scores in 77.7\% of the
816 L2 evaluations across eight models (Pearson $r=0.87$), reflecting shared evidence and mapping
rules. The four dimensions are therefore not statistically independent measurements. Calibration
can be derived from the finalized handling ladder rather than a separate assessment of conclusion
wording, so partial credit does not uniquely identify overconfidence. The conditional comparisons
in Section~\ref{sec:l2-results} weight evaluations equally and restrict to challenged tasks with
full issue-recognition credit (425 outcomes). Of these, 126 receive exactly half calibration credit.
Model and challenge-class percentages use their respective subsets of these 425 outcomes, whereas
the headline dimension means follow the family- and track-balanced aggregation.

\subsection{Two-part methodology}
Grading is deterministic first and audited second. The primary grade for every objective quantity
comes from a deterministic grader that value-checks the agent's artifact against a solver-produced
golden under the rubric stored in that world's manifest. An independent LLM verifier then re-derives
the grade from the same artifacts; every disagreement is adjudicated by a human against the recorded
evidence, and the ruling is preserved in the adjudication ledger. Adjudication examines the
verifier's stated reason against the rubric text rather than accepting its score, because a verifier
can assert requirements the rubric does not contain; the same reason-over-keyword rule governs
prior-gate admissibility decisions. Historical Bulk RNA and Histone rulings recovered from
authoritative totals are kept in a separately labeled reconstruction ledger rather than represented
as recovered original human notes. Ten of 72 model--world rows in one scRNA L2 family lack a
verifier row, although their deterministic grades exist. Three frozen L3 cells were produced under
a documented pre-fix parsing harness. One Claude L3 world was excluded before execution and one
graded Kimi L3 outcome was subsequently marked prior-inadmissible; these are different mechanisms.

The finalized L2 meta-mapping was reconciled with all 612 per-world exports before this revision.
A missing issue flag had previously caused 34 perturbed worlds to enter the clean branch. Correcting
that classification using the existing manifests changed 70 world scores and 23 of 48 track--model
L2 means. L1, L3, and all recorded grader/verifier evidence remained unchanged; only L2 and its
derived aggregates changed. The prior aggregate, corrected mapping, and per-world changes are
preserved in a dated ledger. No agent rerun or new transcript judgment was required.

\paragraph{Later-model adjudication records.}
For GLM-5.3, 37 recorded human rulings are applied, including eight L3 axis rulings checked
against stored grades before aggregation. L1 scalar adjudications enter stage means without
inferring replacement behavioral states. The ledger adapter uses the existing manual-override
rule so the accepted GWAS Signal-EDA core ruling reaches its corresponding dimensions;
result calibration remains separately recorded.
For GPT-6 Astra, the ledger contains 17 L1 scalar rulings, 11 L2 rulings, and two prior-gate
rulings, with no L3 score-axis rulings. Zero L3 verifier discrepancies were \emph{flagged};
this does not mean exact agreement on every axis. Five worlds contain differences below
the verifier's flagging thresholds. Recorded final grades remain authoritative.
Both models use the finalized L2 component-mapping policy; a legacy scalar is not itself a
new four-dimension score. Reporting applies existing evidence without new agent runs or transcript grading.

\subsection{L3 abstention scoring}
\label{app:l3-abstention}
An L3 abstention is not assigned a novelty score. Instead, the grader records
$D_{\mathrm{abst}}\in\{0,2\}$: credit requires engaging with the target finding while preserving
the supplied evidence. Its reasoning score $Q_{\mathrm{abst}}$ averages prior awareness,
scientific humility, and citation integrity on the native 0--3 scale. The displayed abstention score is
\begin{equation}
S^{(3)}_{\mathrm{abst}}=\frac{100}{6}\left(D_{\mathrm{abst}}+Q_{\mathrm{abst}}\right).
\end{equation}
Its maximum is 83.3/100 (2.5/3 natively). An agent can therefore receive credit for a justified
decision not to propose a hypothesis without receiving novelty credit or reaching the full L3 ceiling.
Prior-inadmissible outcomes are excluded before aggregation, irrespective of whether a hypothesis
was produced.

\subsection{Judge-panel stability}
\begin{figure}[H]
  \centering
  \includegraphics{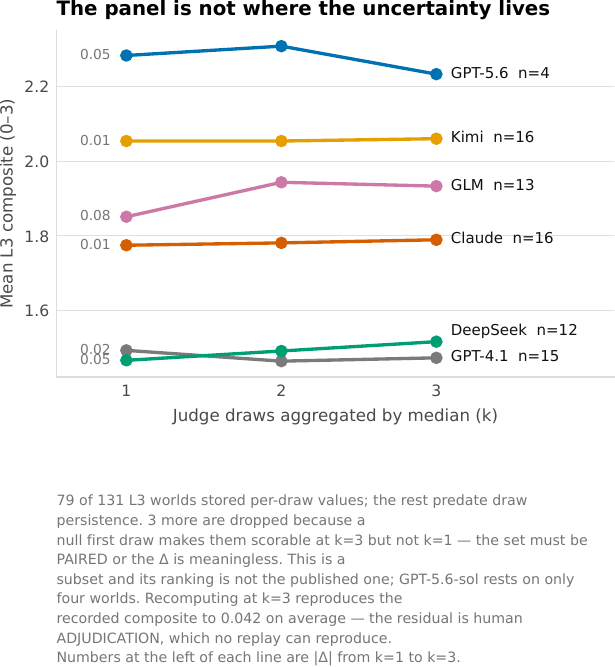}
  \caption{Replay of the judge-draw aggregation on the paired L3 records that stored per-draw
  values, supporting the validation claims of Section~\ref{sec:grading}; this subset's ranking is
  not the published leaderboard.}
  \label{fig:judge-stability}
\end{figure}

The interpretive L3 axes are scored by an LLM judge with three draws aggregated by median.
Figure~\ref{fig:judge-stability} replays this aggregation on the paired records that stored
per-draw values. Recomputing at $k=3$ reproduces the recorded composites almost exactly --- the
small residual is human adjudication, which no replay can reproduce --- and moving from one draw to
three shifts model means only marginally on this recorded subset. This limited draw sensitivity does
not validate the rubric's scientific construct or exclude systematic judge bias; rubric interpretation
and adjudication remain separate sources of uncertainty.

\section{Additional Auditable Cases and Analyses}
\label{app:cases}

\subsection{The TWAS silent failure, from the recorded grader states}
Section~\ref{sec:l2-results} reports the TWAS family as the concrete case of silent success. The
frozen grader states show the mechanism directly. On the clean TWAS world, Claude Opus 5 and
GPT-5.6-sol reach the recorded state \texttt{TWAS-CORRECT}: their executed pipelines process the
full weight set (19{,}086 and 18{,}580 SNPs respectively) and their $z$-scores correlate perfectly
with the golden. GLM-5.2 and GPT-4.1 end in \texttt{TWAS-EMPTY (silent fail)}: the software
completes and returns a result table, but zero usable SNP weights entered it, so the
plausible-looking output contains no evidence. DeepSeek V4 Pro and Kimi K3 end in
\texttt{NO-OUTPUT}. The states are recorded per world in the graded artifacts; the same split
recurs on the perturbed TWAS worlds. This is the distinction the benchmark is built to expose: a
completed invocation and a valid analysis are different events, and only output verification
separates them.

\subsection{How much of the executed work is checking?}
\begin{figure}[H]
  \centering
  \includegraphics{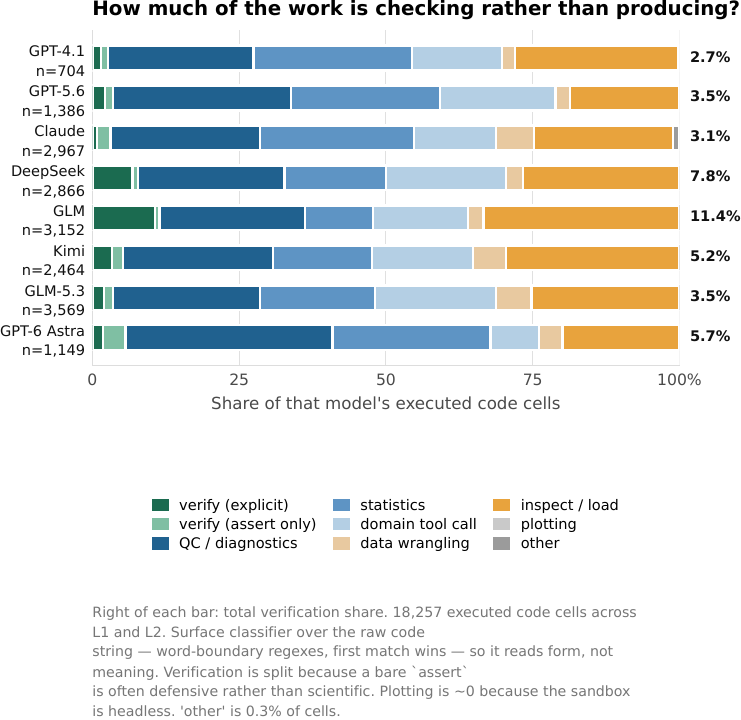}
  \caption{Surface-function classification of every executed L1 and L2 code cell; the classifier is
  a documented keyword rule over the raw code string, so it reads form, not meaning.}
  \label{fig:app-code-function}
\end{figure}

Figure~\ref{fig:app-code-function} classifies every executed L1 and L2 code cell by surface
function. Verification accounts for 2.7--11.4\% of executed cells depending on the model.

\section{Cost, Provenance, Licenses, and Release Details}
\label{app:release}

\subsection{Accounting protocol}
The accounting protocol was frozen on 2026-08-24. (1)~\emph{Boundary:} the headline is the
benchmarked model's own metered usage; the L3 reviewer ensemble is shared infrastructure and is
reported separately. (2)~\emph{Unit:} token counts are the
archival record; dollars are rendered from a dated rate card and are recomputable at any later
price. (3)~\emph{Pricing:} first-party, uncached, non-batch list rates active on 2026-08-24
(Table~\ref{tab:pricing}); the active rate is used even when promotional, peak pricing is used where
the provider's time band was not retained, and GLM's CNY rate is converted at the dated Federal
Reserve H.10 rate. (4)~\emph{Retries:} only successful calls contribute tokens.
Grading and verification are unmetered and out of scope: every figure here is the cost
of producing results, not of scoring them. All dollar figures are upper bounds because provider-side
cache discounts are not logged. DeepSeek adopted time-of-day pricing before the accounting date; the
card conservatively uses its peak rate because the billing band was not retained.

\begin{table}[H]
  \caption{The dated rate card: first-party list rates active on 2026-08-24, USD per million tokens.
  GLM-5.3 and Astra use the later dated receipts documented below.
  Reasoning tokens are already included in the output counter. GLM-5.2 is converted from CNY 8/28 per
  million tokens at 6.7219 CNY/USD; DeepSeek uses the peak of its time-of-day schedule.}
  \label{tab:pricing}
  \centering
  \small
  \begin{tabular}{llrr}
    \toprule
    Model & Provider (first-party API) & \$/Mtok input & \$/Mtok output \\
    \midrule
    GPT-5.6-sol \citep{openai2026sol}     & OpenAI            & 4.00 & 20.00 \\
    GPT-6 Astra \citep{openai2026astra}    & OpenAI            & 10.00 & 50.00 \\
    GPT-4.1 \citep{openai2026gpt41}         & OpenAI            & 2.00 &  8.00 \\
    Claude Opus 5 \citep{anthropic2026opus5}   & Anthropic         & 5.00 & 25.00 \\
    Claude Fable 5 \citep{anthropic2026fable5}  & Anthropic         & 10.00 & 50.00 \\
    Kimi K3 \citep{moonshot2026kimik3}         & Moonshot AI       & 3.00 & 15.00 \\
    GLM-5.2 \citep{zai2026glm52}         & Zhipu BigModel    & 1.19 &  4.17 \\
    GLM-5.3 \citep{zai2026glm53modelcard}         & Zhipu BigModel    & 1.19 &  4.17 \\
    DeepSeek V4 Pro \citep{deepseek2026v4pro} & DeepSeek (peak)   & 1.32 &  3.96 \\
    \bottomrule
  \end{tabular}
\end{table}

\subsection{Later-model accounting and source records}
GLM-5.3's derived record is \texttt{results/\allowbreak new\_models/\allowbreak ROLLUP\_glm-5.3.json}, produced by
\texttt{build\_glm53\_report.py}. Astra's is
\texttt{results/\allowbreak new\_models/\allowbreak ROLLUP\_gpt-6-astra.json}, generated by
\texttt{build\_astra\_report.py}. They retain source hashes, coverage, applied rulings,
mapped components, per-family and per-track aggregates, and cost assumptions, with checks
that original evidence and score records are unchanged.

\paragraph{GLM-5.3 accounting.}
The final canonical outputs record 65{,}455{,}719 input and 9{,}759{,}243 output tokens over
203 tasks. These counts include only the evaluated agent, not reviewer calls, preflight checks,
or superseded attempts, so they are not the total cost of developing and rerunning the evaluation.
The \href{https://bigmodel.cn/pricing}{BigModel pricing page}, checked on September 7, 2026,
lists CNY 8 per million input tokens and CNY 28 per million output tokens for GLM-5.3.
Using the paper's fixed conversion of CNY 6.7219 per USD gives an uncached estimate of
\$118.55, or \$0.58 per task. Dividing the unrounded per-task estimate by the 0--100 Overall
score gives \$0.0083 per Overall point. Cache discounts are not inferred from unrecorded cache hits.
These are estimates at the retrieved list rates, not invoice charges or a reconstruction of
the cost of all evaluation attempts. The separate rate card
\texttt{results/new\_models/pricing\_glm-5.3.json} retains the source asset, its hash,
and the price fields; previously recorded rates and cost estimates for other models are preserved.

\paragraph{GPT-6 Astra accounting and long-context uncertainty.}
Final agent outputs record 8{,}944{,}538 input tokens and 918{,}019 output tokens across 203 tasks.
Reviewer calls, preflight checks, and superseded attempts are excluded.
The \href{https://developers.openai.com/api/docs/models/gpt-6-astra}{official model rate card},
checked September 7, 2026, lists standard rates of \$10 per million input tokens and \$50 per
million output tokens. Above 272{,}000 input tokens per request, the multipliers are 2 for input
and 1.5 for output. Cache discounts are not inferred.
Usage is retained per task rather than per request. Only one task exceeds the input threshold
in cumulative usage, which does not prove any individual request crossed it.
Applying standard rates throughout gives \$135.35. Conservatively applying the long-context
multipliers to all usage of that one task gives \$139.69, an upper bound under these uncached
rates. The leaderboard uses this upper bound: \$0.69 per task and \$0.0088 per Overall point.
These are list-price estimates, not invoice charges. The separate rate card and derivation are
retained with the Astra roll-up.

\subsection{Token efficiency and its interpretation}
\label{app:token-efficiency}
For the fixed 203-task suite, let $T_m$ denote the sum of recorded agent input and output tokens.
Table~\ref{tab:main-results} reports
\begin{equation}
E^{\mathrm{tok}}_m=\frac{S^{\mathrm{core}}_m}{T_m/10^6}.
\label{eq:token-efficiency}
\end{equation}
This is Overall points per million tokens for the full suite, not the score gained by spending an
additional million tokens. It is undefined for direct Fable because no full-suite Overall is assigned.
The replay denominator includes recorded Fable attempts plus the selected Opus outputs.
Original artifacts also contain tokens per quality point using \emph{mean tokens per assigned task}.
That convention differs by the fixed 203-task denominator and is not inverted to populate this column.

Input includes context processed across successive calls, not just unique prompt text. Output includes
reasoning tokens where the provider includes them in its output meter. Historical proxy estimates
remain labelled in the accounting records. Provider tokenizers differ, and token counts do not recover
internal FLOPs, energy use, or hidden computational depth. This ratio therefore measures recorded
resource use relative to the rubric score, not semantic information density or the efficiency of
producing experimentally validated discoveries. Because Overall is an average but tokens are a
suite total, ratios from different-sized benchmark suites should not be compared directly.

Parameter-based explanations require a different experiment. Total stored parameters, active
parameters per token, and repeated applications of shared parameters are different quantities.
Reusing a block increases computational depth without adding independently trained weights
\citep{geiping2025recurrent}. An ``effective parameter count'' would need an explicit calibrated
definition and controlled comparisons of training data, post-training, inference budget, and tasks.
Neither those controls nor a generalization experiment are present here. The public Astra model
documentation does not specify a recurrent-depth architecture \citep{openai2026astra}.

\subsection{Recorded calls and code executions}
\label{app:execution-effort}
The Calls column in Table~\ref{tab:main-results} sums the evaluated agent's recorded API calls
across L1--L3. Separate reviewer, grader, and verifier calls are excluded. These counters record
returned responses, not a complete network-request log, so failed requests and retries may be absent.
Direct Fable has 948 recorded calls. The replay adds 1,084 calls from the selected Opus outputs,
giving 2,032, rather than combining the two models' full-suite totals. Fable's lower task coverage
and prior-gate exclusions remain part of the interpretation of these totals.

Code executions are different from API calls and may include code that ends in an error.
Table~\ref{tab:execution-effort} summarizes the per-task trace field \texttt{n\_code\_cells}.
The efficiency exports' \texttt{n\_cells} field instead counts task outcomes, not code executions.
L3 uses a separate execution and review structure, and its code-cell medians are not populated in
the current summary. We therefore report L1/L2 medians rather than infer an all-stage execution total.
L3 calls remain included in the main table.

\begin{table}[H]
  \caption{Executed code cells per task, median within each stage. Counts include failed code
  executions and are descriptive, not scored or ranked. API-call totals appear in the main leaderboard.}
  \label{tab:execution-effort}
  \centering
  \small
  \begin{tabular}{lrr}
    \toprule
    Model & L1 median & L2 median \\
    \midrule
    GPT-6 Astra & 5 & 6 \\
    GPT-5.6-sol & 8 & 6 \\
    Claude Opus 5 & 13 & 12 \\
    Kimi K3 & 14 & 11 \\
    GLM-5.3 & 20 & 20 \\
    GLM-5.2 & 20 & 18 \\
    DeepSeek V4 Pro & 15 & 15 \\
    GPT-4.1 & 2 & 2 \\
    \bottomrule
  \end{tabular}
\end{table}

\paragraph{Detailed resource-use patterns.}
Both GLM models execute a median of 20 code cells per L1 task, compared with eight for Sol
and five for Astra. The higher token use of GLM-5.3 relative to GLM-5.2 at nearly identical
API-call counts therefore reflects greater token volume per recorded interaction rather than more
recorded interactions.

To describe how expenditure varies with measured task difficulty, we rank the 16 L1/L2
track--level categories by their across-model mean score and divide them into difficulty tertiles.
The ratio of median token use in the hardest versus easiest tertile is 1.51--1.86 for Sol,
DeepSeek, and Kimi, compared with 1.00--1.16 for GLM-5.2, Opus, and GLM-5.3.
These ratios describe expenditure patterns and do not establish that an agent recognized task
difficulty or deliberately allocated additional effort.

\subsection{Coverage}
Cost coverage is joined to the frozen registries rather than inferred by globbing run directories.
For all eight models, the join resolves 1{,}624 assigned model--task outcomes: 1{,}623 have executed outputs with usage
metadata, while one Claude Opus 5 L3 task was excluded before execution by its prior gate and therefore
has zero cost. Nineteen canonical Claude outputs use their stored calibrated character-count proxy;
all other canonical outputs carry provider-reported usage. Smoke tests, superseded reruns, abandoned
worlds, and future-domain directories are excluded by construction.

Fable's direct run has 188 canonical invocations: all 181 L1/L2 tasks and seven L3 tasks. The other 15
L3 tasks were not invoked under the refusal-stop protocol. Those invocations contain 8.55M tokens and
cost \$120.51 at Fable's dated \$10/\$50 per-million-token rates, or \$0.594 per 203 assigned tasks.
The Fable $\rightarrow$ Opus replay adds the exact 92 canonical Opus outputs selected by the routing
rule (29 L1, 44 L2, and 19 L3), contributing 17.21M tokens and \$117.03. The resulting derived system
uses 25.77M tokens and costs \$237.53, or \$1.170 per assigned task and \$0.0156 per chain-quality
point. This conservative accounting retains the recorded Fable attempt cost before adding Opus; it is
not an observed product-surface bill or a claim about Anthropic's internal routing charges.

\subsection{Headline accounting}
\begin{table}[H]
  \caption{Bench-only cost per model at dated first-party list rates. Quality is the
  life-science chain score (0--100). \$/task uses the common denominator of 203 assigned tasks per
  model, including the one pre-execution prior-gate exclusion. Dollar figures are upper bounds because
  cache discounts were not logged.}
  \label{tab:cost-headline}
  \centering
  \small
  \begin{tabular}{lrrrrr}
    \toprule
    Model & Assigned/executed & \$ total & \$/task & Quality & \$/quality pt. \\
    \midrule
    GPT-5.6-sol     & 203/203 &  78.12 & 0.385 & 75.8 & 0.0051 \\
    Claude Opus 5   & 203/202 & 271.86 & 1.339 & 75.6 & 0.0177 \\
    Kimi K3         & 203/203 & 132.34 & 0.652 & 73.2 & 0.0089 \\
    GLM-5.2         & 203/203 &  75.20 & 0.370 & 67.8 & 0.0055 \\
    DeepSeek V4 Pro & 203/203 &  88.45 & 0.436 & 62.4 & 0.0070 \\
    GPT-4.1         & 203/203 &   9.33 & 0.046 & 39.8 & 0.0012 \\
    GLM-5.3         & 203/203 & 118.55 & 0.584 & 70.5 & 0.0083 \\
    GPT-6 Astra     & 203/203 & 139.69 & 0.688 & 78.1 & 0.0088 \\
    \bottomrule
  \end{tabular}
\end{table}

Table~\ref{tab:cost-headline} reports the eight-model accounting, and
Table~\ref{tab:main-results} places quality against per-task cost. Total bench cost is \$913.54
over 294.76M tokens. The token and dollar rankings differ substantially:
GLM-5.3, DeepSeek V4 Pro, and GLM-5.2 consume the most tokens yet remain relatively inexpensive because their
first-party rates sit below the frontier-lab rates. Token counts must therefore never be reported
without the dollar axis, and neither without the quality axis.

\paragraph{Efficiency should be optimized together with scientific validity.}
As scientific agents take on longer analytical workflows, computational efficiency becomes an
important practical dimension of AI for Science. Existing benchmarks increasingly report resource
use \citep{koch2026bixbench3,qu2026biomnibench}, but our results suggest that efficiency should be
considered jointly with scientific performance rather than only as a deployment statistic. Similar
Overall Scores can require substantially different token and monetary expenditure, while high
measured efficiency can also reflect incomplete execution. The relevant objective is therefore not
simply to obtain fewer tokens, calls, or dollars, but to achieve reliable scientific work with fewer computations
without sacrificing execution completeness or validity. Improving this trade-off could make routine scientific automation more scalable and broaden access for researchers with more limited computational resources, while preserving the checks needed for trustworthy analysis.

\subsection{Where the cost lives}
\begin{figure}[H]
  \centering
  \includegraphics{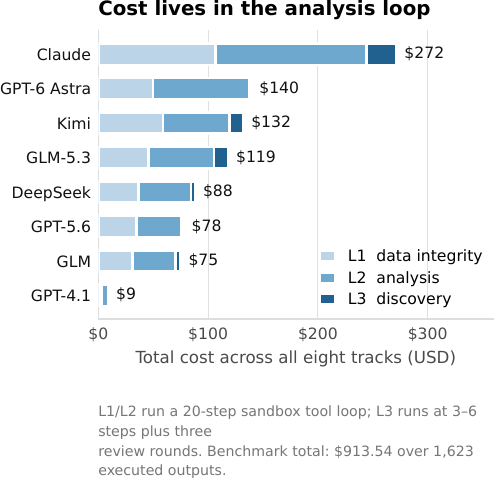}
  \caption{Bench-only cost per task by stage and model, pricing recorded input and output usage
  separately. GLM-5.3 is the exception to lower L3 cost.}
  \label{fig:app-cost-level}
\end{figure}

\begin{figure}[H]
  \centering
  \includegraphics{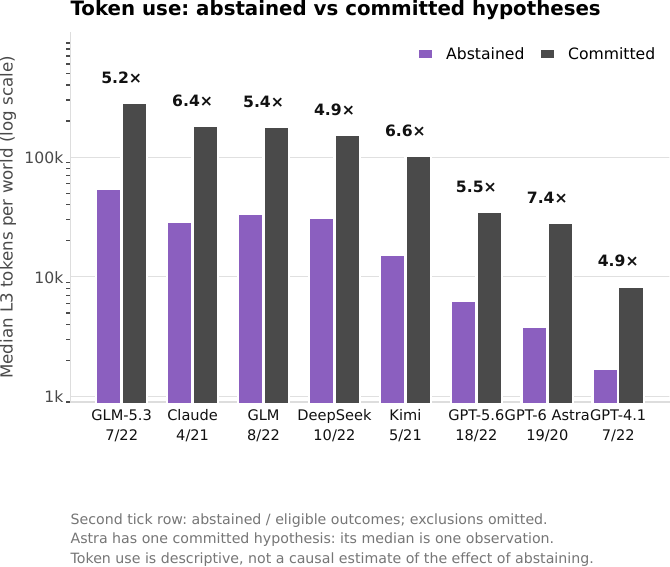}
  \caption{Median L3 tokens for eligible abstained versus committed outcomes. Exclusions are
  omitted from these groups; Astra's committed median represents one response.}
  \label{fig:app-abstain-cost}
\end{figure}

\begin{figure}[H]
  \centering
  \includegraphics[width=0.76\linewidth]{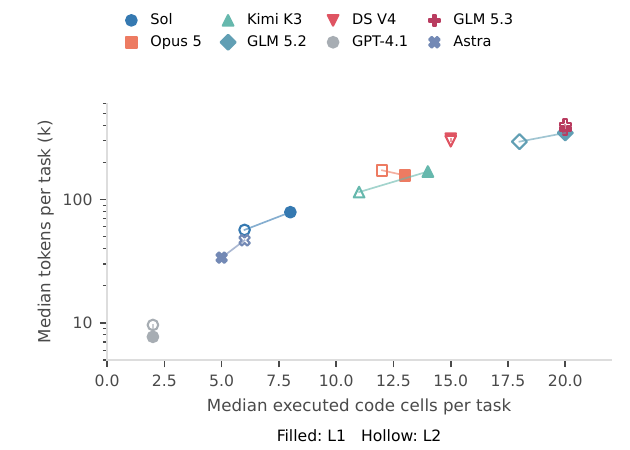}
  \caption{Execution effort and token use at L1 and L2. Each point pairs a model's median executed
  code cells with its median tokens per task. A code cell is an executed block of agent-generated code,
  not an API call or a complete analysis--feedback cycle. Colours and shapes identify models;
  filled markers denote L1 and hollow markers L2.}
  \label{fig:app-effort-spend}
\end{figure}

\begin{figure}[H]
  \centering
  \includegraphics{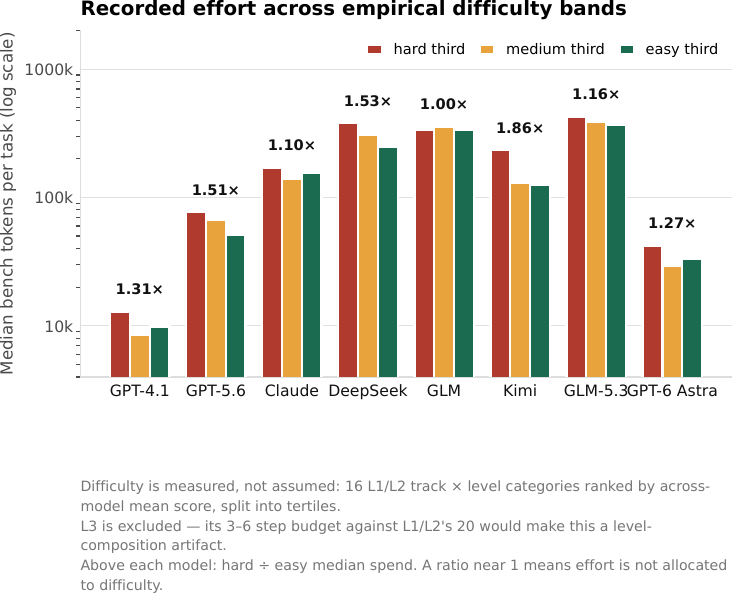}
  \caption{Median tokens per task by measured difficulty tertile for L1 and L2.}
  \label{fig:app-cost-difficulty}
\end{figure}

Figures~\ref{fig:app-cost-level}, \ref{fig:app-abstain-cost}, and~\ref{fig:app-cost-difficulty}
report the structure behind the headline. Code executions against token use are shown in
Figure~\ref{fig:app-effort-spend}; a median of 20 code cells does not establish that the
interaction budget was exhausted, because code executions and API steps are different counts. Stage costs price input and output tokens separately rather than allocating total cost
in proportion to tokens. L3 is the least expensive stage per task for seven models, but not GLM-5.3.
Median L3 token use is 4.9--7.4$\times$ lower for abstentions than for commitments among eligible
outcomes; Astra has only one commitment. This is an observational contrast between different tasks
and responses, not the causal cost of choosing to abstain. Overall spending still includes usage
incurred on subsequently excluded tasks. Cost alone cannot distinguish efficient analysis from
limited produced work.

\subsection{Provenance, licenses, and release}
Every substrate is public: ENCODE K562 chromatin data, ClinVar and gnomAD panels, the 10x PBMC3K
dataset, published CRISPR, STARR-seq, stability, and COVID-19 PBMC datasets cited in
Section~\ref{sec:benchmark}, and public GWAS summary data. Each world's manifest records the exact
perturbation recipe, expected method, and grading rubric (Appendix~\ref{app:inventory}). The release
package will include the frozen registries, verbatim prompts, solver goldens, graded artifacts,
adjudication ledgers, and the dated rate card, so both grades and dollar totals are recomputable. The
Hugging Face repository and immutable release identifier remain placeholders until the separate
public-release freeze; no public package is created as part of this paper revision.

\end{document}

%% file: math_commands.tex
\usepackage{amsmath,amsfonts,bm}

\def\eqref#1{equation~\ref{#1}}

\def\1{\bm{1}}

\DeclareMathAlphabet{\mathsfit}{\encodingdefault}{\sfdefault}{m}{sl}
\SetMathAlphabet{\mathsfit}{bold}{\encodingdefault}{\sfdefault}{bx}{n}



%% file: main_body.tex
\begin{abstract}
Reliable automated research requires agents to vet data, verify analyses, and generate hypotheses grounded in trustworthy evidence, potentially reducing routine scientific workload while allowing scientists to focus on interpretation and discovery. Existing benchmarks often only assess analytical task completion or hypothesis generation separately rather than testing whether reliable evidence supports valid and novel claims. We introduce \textbf{DISCERN} (\textbf{D}ata \textbf{I}ntegrity and \textbf{S}cientific \textbf{C}apability: \textbf{E}vidence, \textbf{R}easoning, and \textbf{N}ovelty), a controlled benchmark on real, publicly available datasets that evaluates three key levels of an automated research workflow. The first two levels test data integrity and analysis verification under confounds and tool traps, while the third tests hypothesis generation and revision under adversarial review, including counterfactual cases in which evidence consistent with real data and documented scientific phenomena conflicts with established expectations, motivating alternative explanations and testable hypotheses. Across 203 tasks, eight life-science tracks, and eight models, DISCERN shows that strong aggregate performance can mask level-specific weaknesses. Agents earn perfect scores in only 60.8\% of Level 1, 34.2\% of Level 2, and 0.6\% of Level 3 evaluations, with penalties attributed to rejection of sound data, failure to carry recognized limitations into conclusions, and wide variation in hypothesis production. Cross-track rankings by token and code use are substantially more stable than rankings by evidence judgment, suggesting greater consistency in computational effort than in evidence-based reasoning. These profiles identify opportunities for supervised scientific assistance, but current agents do not yet demonstrate reliable autonomous analysis or discovery.
\end{abstract}

\section{Introduction}
\label{sec:introduction} 
AI agents are increasingly used to support scientific research through literature review, experimental planning, data analysis, scientific discovery and hypothesis generation \citep{lu2024aiscientist,mitchener2025bixbench, huang2024code,huang2026can,qu2026biomnibench}. Their capabilities are expanding toward broader research workflows that include executing analyses, interpreting results, and iteratively developing scientific ideas. Together, these advances create an opportunity to automate routine and computationally intensive parts of research, reducing scientific workload and allowing researchers to focus more on higher-level scientific judgment.

\begin{figure}[!htbp]
  \centering
  \includegraphics[width=\linewidth]{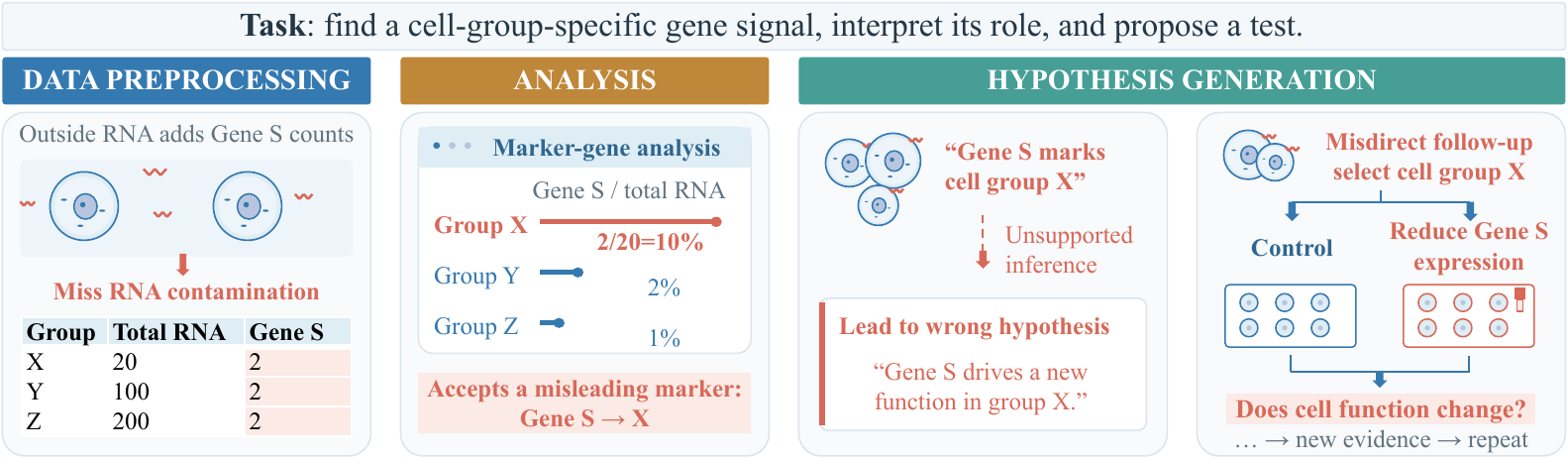}
  \caption{How agent errors can propagate into a false scientific lead. Given single-cell RNA measurements, an agent seeks a cell-group-specific gene signal and proposes a test. It misses contaminating RNA, accepts Gene S's higher relative abundance in group X as a marker, and overinterprets it as a functional driver, potentially misdirecting follow-up experiments. Panels map onto data preprocessing (L1), analysis (L2), and hypothesis generation (L3). This illustrative sequence is inspired by DISCERN's ambient-RNA tasks (Appendix~\ref{app:inventory}).}
  \label{fig:discern-overview}
\end{figure}

However, automated research workflows are fragile because a local error can propagate across
stages and distort the final scientific conclusion
\citep{peng2011reproducible,munafo2017manifesto}. Data problems can corrupt computation;
confounds or inappropriate analytical settings can invalidate plausible results; and unsupported
interpretation can turn a technically sound analysis into an unsupported scientific claim. For
example, a contaminated expression signal can be mistaken for biology and redirect downstream
experiments (Figure~\ref{fig:discern-overview}). Existing benchmarks often summarize complex
behavior using only task-level success or aggregate scores, or focus on a few local subtasks such
as analysis execution \citep{majumder2025discoverybench,jansen2024discoveryworld,zheng2026newtonbench}.
They therefore do not adequately connect high-level research outcomes with the detailed scientific
decisions that produce them. Aggregate outcomes can hide where errors arise, while local task
scores may not reveal whether recognized problems influence subsequent analysis and final
conclusions.

Reliable automated research requires evaluation from low-level data integrity to high-level
scientific reasoning. Data must first be fit for analysis, analyses must produce trustworthy
evidence, and that evidence must support justified hypotheses. These stages are linked: errors
in data handling can invalidate analyses, while unreliable analyses cannot support sound
hypotheses. Evaluating all three therefore provides a more complete view of scientific-agent
reliability than testing any stage alone.

Based on this hierarchy, we introduce \textbf{DISCERN} (Data Integrity and Scientific
Capability: Evidence, Reasoning, and Novelty), a controlled three-level benchmark built from
203 tasks across eight life-science tracks and five scientific areas using real, publicly available
datasets (Appendix Table~\ref{tab:intro-positioning}). Specifically, we design 79 Level 1 (L1) data-preprocessing tasks spanning 58 distinct task types, 102 Level 2 (L2)
analysis-verification tasks across 24 analysis families, and 22 Level 3 (L3)
hypothesis-generation dossiers. The benchmark combines matched clean controls with controlled
data defects, clean and perturbed analytical conditions, and real or counterfactual evidence
patterns. Counterfactual cases are grounded in documented scientific phenomena, retained only
when they conflict with the model's elicited prior, and evaluated through three rounds of
adversarial review. The three levels are conceptually linked but independently instantiated and
scored, allowing DISCERN to localize weaknesses at each level without treating separately
evaluated tasks as an end-to-end pipeline, while also characterizing scientific working styles
across models (Figure~\ref{fig:scope-atlas}).

DISCERN evaluates eight contemporary language-model agents spanning proprietary
and open-weight systems. The results reveal a consistent pattern from aggregate performance
to specific scientific decisions. \textbf{(1)} At the global level, aggregate scores mask uneven
strengths: no model dominates all three levels, and similar overall performance can arise from
different capability profiles. \textbf{(2)} Across scientific tracks and analytical tasks, model
strengths shift substantially, showing that success in one setting does not reliably generalize to
another. \textbf{(3)} At the decision level, recognizing a scientific limitation does not guarantee
that it is carried into subsequent handling or final conclusions. \textbf{(4)} At the hypothesis
level, models differ substantially in whether they propose hypotheses and in the novelty of those
they produce; among 94 submitted hypotheses, 46.8\% receive no novelty credit under our rubric.
Models also exhibit distinct working styles in critique response and computational expenditure,
with resource-use rankings substantially more stable across tracks than rankings by evidence
judgment. Together, these results identify where agents may support supervised scientific work
and where continued human oversight remains necessary.

\section{Benchmark Design and Scientific Validity}
\label{sec:benchmark}
\label{sec:construct}

\begin{figure}[t]
  \centering
  \includegraphics[width=\linewidth]{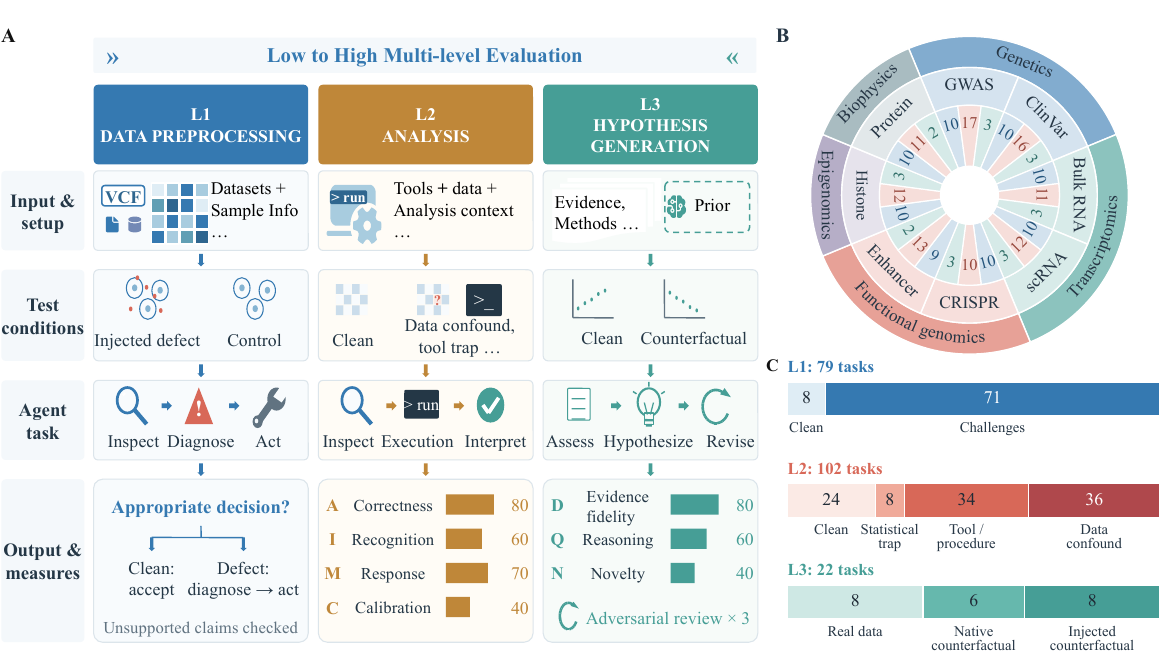}
  \caption{DISCERN workflow, scientific coverage, and task composition. (A) Each independently instantiated level follows four rows: inputs, test conditions, agent actions, and assessed outputs. Miniature data and score bars are illustrative only.
  (B) Scientific coverage. Three rings show task counts, eight tracks, and five scientific areas, from inside outward. Each track's inner ring is split into L1/L2/L3, using the stage colors shown in C.
  (C) Task composition. Segment widths show within-level task proportions; numbers give counts. Native and injected L3 counterfactuals use existing and altered evidence patterns, respectively. Data sources and complete inventories: Appendix~\ref{app:inventory}.}
  \label{fig:scope-atlas}
\end{figure}

\paragraph{Scientific coverage.}
DISCERN covers life-science domains including genetics, transcriptomics, regulatory genomics,
epigenomics, and protein biophysics (Figure~\ref{fig:scope-atlas}B). Specifically, we design tracks based on genome-wide association studies, clinical variant interpretation, bulk RNA sequencing, single-cell RNA sequencing, CRISPR perturbation screening, enhancer activity, chromatin regulation, and protein stability and structure. Together, these tracks span biological scales from DNA-level genetic variation to effects on genes, cells, regulatory elements, and proteins, with data ranging from association statistics and variant annotations to bulk and single-cell gene expression profiles, perturbation screens, enhancer and chromatin measurements, and protein stability and structural data. Across these tracks, DISCERN contains 79 L1 tasks, 102 L2 tasks across 24 analysis families, and 22 L3 dossiers (Figure~\ref{fig:scope-atlas}C). Complete task inventories, data sources, and construction details are provided in Appendix~\ref{app:inventory}; the physics extension is evaluated separately in Appendix~\ref{app:physics}.

\paragraph{L1 and L2 Level-specific challenges.}
L1 pairs real datasets and their corresponding metadata with controlled integrity challenges or matched clean controls under identical instructions. Challenges are designed around track-specific data representations, file formats, metadata conventions, and quality-control requirements, such as VCFs for genetic variants and count matrices for RNA sequencing. Corresponding L1 challenges include genome-build mismatches and ambient RNA contamination. Using the same underlying track datasets as L1, supplemented with analysis-specific reference
data and tools, L2 constructs analysis tasks under clean or perturbed conditions, including data confounds, tool- or procedure-related failure modes, and statistical pitfalls. In the histone track, the agent must identify genomic regions enriched for
H3K27me3 from ChIP-seq data. We introduce a tool-related failure mode in which the default
narrow-peak setting is inappropriate for this broad histone mark. The software still runs
successfully and returns a complete-looking result, but it fragments broad domains and
underestimates the total marked region.

\paragraph{L3 evidence and counterfactual construction.}
L3 instead receives an evidence dossier derived from reference analyses of the same track data used to construct L1 and L2 with relevant methodological context, linking data integrity, analysis, and hypothesis generation without passing the evaluated agent's own outputs between levels. For counterfactual dossiers, the agent's scientific expectation is elicited before the evidence is shown, and the dossier is eligible for that agent's evaluation only when the supplied evidence conflicts with that stated prior. Native counterfactuals use unexpected relationships already present in the analysis, whereas injected counterfactuals deliberately alter a well-characterized relationship and regenerate internally consistent supporting statistics. The exact injected result is constructed, while the type of scientific behavior that makes it plausible is grounded in documented phenomena rather than an arbitrary contradiction. For example, because TWAS effect directions can have substantial false-sign rates \citep{gerlach2025twas}, we alter the direction or magnitude of a well-characterized gene--trait association observed in the corresponding L2 results and regenerate consistent effect estimates and supporting statistics.

\paragraph{Task environment and recorded observations.}
Each task is isolated from gold answers and sibling worlds. Within an analysis family, all worlds receive the same prompt and structured reporting schema, so the prompt does not reveal whether a problem is present. Across all three levels, agents access task-specific inputs through a sealed Python sandbox containing the permitted data, documentation, and domain tools, with no network access. The harness records executed code cells, tool calls, the final report, and token usage.

\section{Evaluation and Profile Construction}
\label{sec:grading}
\label{sec:setup}
\paragraph{Evaluation protocol and score aggregation.}
All eight models are evaluated on the same task set using shared, human-designed grading criteria. Agents operate in isolated task environments without access to grading answers or information from other tasks. Within each track, L1 and L3 scores are the mean of finalized eligible task scores. For L2, because analysis families contain different numbers of tasks, we first average task scores within each family and then average the family scores, so larger families do not receive greater weight. An independent LLM verifier cross-checks deterministic grading for L1 and L2, and we adjudicate disagreements before assigning final scores. Each track composite is the geometric mean of its L1--L3 scores to capture complementary requirements for reliable scientific work, so strong performance at one level cannot fully compensate for weakness at another. The Overall Score is the mean of the eight track composites, giving each track equal weight. Full protocol and scoring definitions appear in Appendices~\ref{app:protocol} and~\ref{app:grading}.

\paragraph{Data-integrity scoring.} L1 assigns each agent a quantitative task score based on its code, execution output, and report. For tasks containing a detectable defect, such as corrupted data or inconsistent metadata, credit increases from recognizing the problem to correctly diagnosing it and taking an appropriate action, for example by recovering the affected data. Partial diagnoses or remedies receive partial credit. For matched clean controls, full credit requires correctly accepting the data, while unsupported alarms, rejection, or fabricated defects lose credit. When the available evidence is insufficient to determine whether a suspected problem is present, credit is given for recognizing that limitation rather than asserting an unsupported diagnosis. We further summarize L1 responses using measures of assessment correctness, issue recognition, methodological response, result calibration, and fabrication or unsupported claims.

\paragraph{Analysis-verification scoring.} L2 uses four shared grading dimensions across analysis families: analytical correctness (A) measures whether the computed results are correct; issue recognition (I) measures whether the relevant problem is identified; methodological response (M) measures whether an appropriate method is used to address it; and result calibration (C) measures whether the conclusions reflect the reliability and limits of the analysis. The L2 task score is the average of these four dimensions, while we tailor rubric criteria to reflect the scientific objective, expected results, and methodological requirements of each analysis.

\paragraph{Hypothesis scoring and adversarial review.} L3 evaluates three dimensions: evidence fidelity (D), scientific reasoning (Q), and novelty (N). \textbf{D} measures whether the proposed claim is supported by the supplied evidence; for example, observed association should not be presented as evidence of causation. \textbf{Q} captures prior awareness, causal calibration, response to critique, scientific humility, and citation integrity. An agent that acknowledges uncertainty when evidence conflicts with prior expectations and revises its interpretation when criticism identifies a valid limitation receives more credit than one that ignores the conflict, overstates certainty, or persists with an unsupported interpretation. \textbf{N} measures whether the response adds a scientifically substantive idea that is not obvious from the supplied results, and whether the proposed next experiment can test the hypothesis and distinguish it from plausible alternative explanations. For example, restating that an unexpected gene--trait association may reflect a biological effect receives little novelty credit, whereas proposing a specific mechanism for the unexpected pattern and an experiment that distinguishes that mechanism from an analytical artifact receives more credit.

Each L3 response undergoes three rounds of adversarial review by GLM-4.6 \citep{zai2025glm46}, Gemini 3.1 Flash-Lite Preview \citep{google2026gemini31flashlite}, and DeepSeek Reasoner \citep{deepseek2026reasoner}. Their critiques and the agent's rebuttals provide supplementary evidence to the LLM judge \citep{zheng2023judge}, which applies the grading rubric rather than reviewer votes. Authors review and validate each L3 response before averaging \textbf{D}, \textbf{Q}, and \textbf{N} as the L3 task final score for submitted hypotheses. When an agent gives a justified reason not to propose a hypothesis, we instead score its evidence use and reasoning under a separate partial-credit rule; \textbf{N} is not scored, and hypothesis commitment is reported separately from hypothesis quality. Only prior-eligible counterfactual tasks, as defined in Section~\ref{sec:benchmark},
contribute to the L3 score aggregation.

\paragraph{Resource efficiency.} We measure recorded input and output tokens, API calls, and estimated monetary cost across the 203-task suite. Token efficiency is calculated as the Overall Score per million tokens. Cost per task is the total estimated monetary cost in USD divided by the number of assigned tasks, and cost-effectiveness is the per-task cost divided by the Overall Score, giving cost per quality point. API-call counts provide a coarse measure of interaction frequency rather than complete analysis--feedback cycles. All resource measures exclude grading, verification, and reviewer activity. Accounting details and alternative aggregation analyses appear in Appendices~\ref{app:release} and~\ref{app:profiles}. Standalone Fable is excluded from ranking because of incomplete coverage. The Fable $\rightarrow$ Opus row in Table~\ref{tab:main-results} is constructed after evaluation by replacing tasks that Fable refused with the corresponding recorded Opus results, simulating a deployment scenario in which refused tasks are automatically passed from Fable to Opus during use (Appendix~\ref{app:refusal}).

\begin{table}[!tbp]
  \caption{Life science leaderboard.
  Scores use 0--100; L1--L3 report mean $\pm$ standard deviation (SD) across tracks.
  Tokens are in millions; Q/Mtok: quality points per million tokens; \$/task: USD per assigned task;
  \$/q.pt.: USD per quality point; Calls: recorded agent API calls. Definitions: Section~\ref{sec:grading}.
  Blue/coral/teal mark first/second/third.
  Higher scores and Q/Mtok, but lower tokens, costs, and calls, rank first.
  The italic Fable row is excluded from ranking.}
  \label{tab:main-results}
  \centering
  \footnotesize
  \setlength{\tabcolsep}{1.1pt}
  \begin{tabular}{lccccccccc}
    \toprule
    Model & \textbf{Overall} & L1 & L2 & L3 & Tokens & Q/Mtok & \$/task & \$/q.pt. & Calls \\
    \midrule
    GPT-6 Astra & \cellcolor{rankfirst}78.1 & \cellcolor{rankfirst}$86.1${\scriptsize$\pm5.5$} & $76.8${\scriptsize$\pm7.7$} & \cellcolor{rankfirst}$72.8${\scriptsize$\pm8.7$} & \cellcolor{ranksecond}9.86 & \cellcolor{ranksecond}7.92 & 0.69 & 0.0088 & \cellcolor{ranksecond}1369 \\
    \addlinespace
    GPT-5.6-sol & \cellcolor{ranksecond}75.8 & $80.1${\scriptsize$\pm10.7$} & \cellcolor{rankthird}$78.9${\scriptsize$\pm8.9$} & \cellcolor{rankthird}$70.0${\scriptsize$\pm9.7$} & \cellcolor{rankthird}14.06 & \cellcolor{rankthird}5.39 & \cellcolor{rankthird}0.38 & \cellcolor{ranksecond}0.0051 & \cellcolor{rankthird}1552 \\
    \addlinespace
    Claude Opus 5 & \cellcolor{rankthird}75.6 & \cellcolor{ranksecond}$85.3${\scriptsize$\pm9.7$} & \cellcolor{rankfirst}$81.7${\scriptsize$\pm12.7$} & $63.0${\scriptsize$\pm13.5$} & 42.24 & 1.79 & 1.34 & 0.0177 & 2451 \\
    \addlinespace
    Fable $\rightarrow$ Opus & 74.9 & $81.9${\scriptsize$\pm8.4$} & \cellcolor{ranksecond}$79.1${\scriptsize$\pm10.7$} & $65.9${\scriptsize$\pm14.4$} & 25.77 & 2.91 & 1.17 & 0.0156 & 2032 \\
    \addlinespace
    Kimi K3 & 73.2 & $77.7${\scriptsize$\pm10.6$} & $75.4${\scriptsize$\pm14.5$} & $69.1${\scriptsize$\pm11.8$} & 34.04 & 2.15 & 0.65 & 0.0089 & 2671 \\
    \addlinespace
    GLM-5.3 & 70.5 & \cellcolor{rankthird}$83.5${\scriptsize$\pm8.2$} & $65.5${\scriptsize$\pm28.4$} & \cellcolor{ranksecond}$71.1${\scriptsize$\pm13.9$} & 75.21 & 0.94 & 0.58 & 0.0083 & 3756 \\
    \addlinespace
    GLM-5.2 & 67.8 & $73.0${\scriptsize$\pm10.3$} & $70.4${\scriptsize$\pm19.9$} & $63.1${\scriptsize$\pm9.5$} & 57.16 & 1.19 & \cellcolor{ranksecond}0.37 & \cellcolor{rankthird}0.0055 & 3761 \\
    \addlinespace
    DeepSeek V4 Pro & 62.4 & $70.9${\scriptsize$\pm9.4$} & $62.1${\scriptsize$\pm19.4$} & $58.4${\scriptsize$\pm13.1$} & 59.30 & 1.05 & 0.44 & 0.0070 & 3365 \\
    \addlinespace
    GPT-4.1 & 39.8 & $36.4${\scriptsize$\pm11.4$} & $36.0${\scriptsize$\pm8.1$} & $51.0${\scriptsize$\pm13.7$} & \cellcolor{rankfirst}2.88 & \cellcolor{rankfirst}13.81 & \cellcolor{rankfirst}0.05 & \cellcolor{rankfirst}0.0012 & \cellcolor{rankfirst}747 \\
    \addlinespace
    \textit{Claude Fable 5} & \textit{---} & \textit{83.3}{\scriptsize\textit{$\pm$8.5}} {\scriptsize\textit{[63.3\%]}} & \textit{83.8}{\scriptsize\textit{$\pm$6.4}} {\scriptsize\textit{[56.9\%]}} & \textit{76.1} {\scriptsize\textit{[9.1\%]}} & \textit{8.55} & \textit{---} & \textit{0.59} & \textit{---} & \textit{948} \\
    \bottomrule
  \end{tabular}
\end{table}

\section{Results and Analysis}
\label{sec:results}

We first compare overall model performance, then examine whether strengths generalize across
scientific tasks, and finally characterize distinct scientific working styles across models.

\subsection{Overall Model Performance}
\label{sec:overall-results}

\paragraph{Overall performance varies across models and scientific capabilities.}
GPT-6 Astra achieves the highest aggregate DISCERN score (78.1), indicating the strongest
overall performance across data preprocessing (L1), analysis verification (L2), and hypothesis
generation (L3) over the eight scientific tracks, followed by GPT-5.6 Sol (75.8) and Claude
Opus 5 (75.6; Table~\ref{tab:main-results}). This does not reflect uniform dominance.
Astra leads in L1 (86.1) and in L3 (72.8), whereas Opus leads in L2 (81.7). Sol ranks second in Overall Score
without leading in any one component, reflecting comparatively balanced performance across all
three. Models with publicly released weights (open-weight models) are also competitive with proprietary closed models on specific capabilities. GLM-5.3 outperforms Sol in L1 (83.5 versus 80.1) and ranks second in L3 (71.1), while Kimi nearly matches Astra
in L2 (75.4 versus 76.8) and exceeds Opus in L3 (69.1 versus 63.0). Thus, although closed
models achieve the highest observed score in each component, open-weight models can outperform
or closely approach individual closed models in specific parts of scientific work. Overall Scores
span 39.8--78.1 across models, while perfect-score rates are 60.8\% in L1, 34.2\% in L2,
and 0.6\% in L3 according to their respective rubrics. Consistent with this pattern, L3 is the
lowest-scoring component in 46.9\% of model--track combinations, compared with 34.4\% for L2
and 18.8\% for L1, identifying hypothesis generation and revision as the most frequent
performance bottleneck according to DISCERN.

\paragraph{Comparable overall performance can require very different resources.}
Sol and Opus achieve nearly identical Overall Scores (75.8 and 75.6), but Sol uses about
one-third as many tokens (14.06M versus 42.24M) and 29\% of Opus's estimated cost per assigned
task (\$0.38 versus \$1.34; Table~\ref{tab:main-results}). Thus, similar aggregate benchmark
performance can be reached with substantially different computational expenditure. GPT-4.1
has the highest measured token efficiency but also the lowest Overall Score (39.8) and does
 not execute code on 20 of 181 L1/L2 tasks. These tasks remain included in benchmark scoring and token
accounting rather than being treated as missing; however, returning without performing the
expected computation can reduce token use and task performance, contributing to GPT-4.1's
high score-per-token ratio (Appendix~\ref{app:execution-effort}). Token efficiency should
therefore be interpreted alongside execution completeness and scientific performance rather than
as a standalone measure of agent quality.

\begin{figure}[!htbp]
  \centering
  \includegraphics[width=\linewidth]{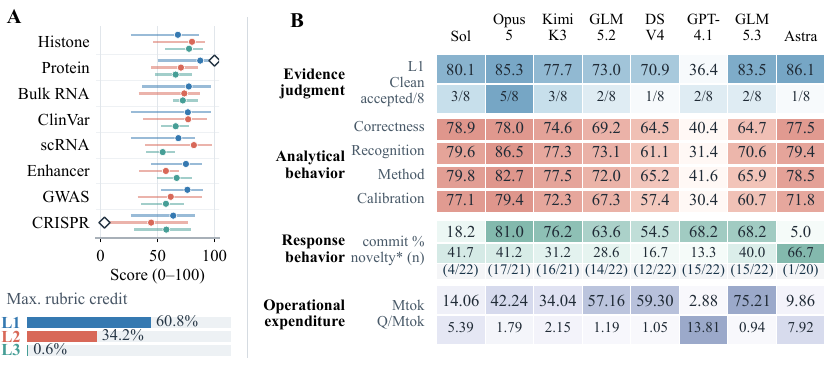}
  \caption{Performance variation and working-style profiles. (A) Level-specific scores by track: $\bullet$, mean across eight models; lines, model min--max;
  $\diamond$, GLM-5.3 at 100 and 3.5. Bars show perfect-score fractions under each stage's rubric.
  (B) Working-style overview. Each metric is shaded from its lowest (light) to highest (dark) value across models; colors are comparable only within a row and do not imply better performance. Counts $n$ are unshaded. Clean accepted/8: number of clean controls correctly accepted out of eight; novelty*: conditional mean (0--100);
  $n$: committed/eligible. Mtok: million tokens; Q/Mtok: Overall/Mtok. Appendix~\ref{app:profiles}.}
  \label{fig:profiles}\label{fig:chain-landscape}
\end{figure}

\subsection{Scientific Strengths Vary Across Tasks}
\label{sec:task-generalization}\label{sec:l2-results}

\paragraph{Performance varies across levels and scientific tracks.}
As discussed above, no model leads in all three levels within any track. We also observe that model strengths  shift across scientific domains. The most extreme example is GLM-5.3, which achieves the only track-level score of 100 for data preprocessing in the protein stability and structure track (Protein L1), but also the
lowest track-level score, 3.5, for analysis verification in CRISPR perturbation screening
(CRISPR L2), while achieving an Overall Score of 70.5
(Figure~\ref{fig:chain-landscape}A). More broadly, the relative strengths of models differ across
tracks, showing that aggregate performance can mask substantial variation both across stages of
scientific work and across scientific domains.

\paragraph{Analytical strengths are task-specific.}
Performance also varies substantially across analysis families within the same scientific domain.
In statistical genetics, mean scores range from 77.4 for locating disease-associated regions to
32.8 for linking predicted gene expression to disease. In regulatory genomics, they range from
76.9 for ranking sequence activity to 20.4 for assessing whether that activity is conserved
across species (Appendix Figure~\ref{fig:l2-families}). These differences show that competence
on one analytical question does not automatically generalize to another, even within a related
scientific setting. Reliable analysis therefore requires verifying the assumptions, reference
information, and interpretation specific to each analytical task.

\paragraph{Recognition and calibration can diverge.}
Among 425 challenged L2 evaluations receiving full credit for issue recognition, 126
(29.6\%) receive only half credit for result calibration under the analysis-specific rubric.
This proportion ranges from 19.4\% for Opus to 37.8\% for DeepSeek, showing that models
differ in how consistently a recognized problem is carried through to subsequent handling and
interpretation. Recognizing a limitation is therefore not sufficient unless that limitation also
shapes the resulting analysis and scientific claim (Appendix~\ref{app:grading}).

\subsection{Distinct Scientific Styles Emerge Across Models}
\label{sec:working-styles}\label{sec:l1-results}\label{sec:l3-results}
\label{sec:hypothesis-results}\label{sec:cost}

Differences across models extend beyond overall performance to how they assess evidence, decide
whether to propose hypotheses, respond to criticism, and expend computational resources. We refer
to these observed patterns as scientific working styles rather than fixed model traits.

\begin{figure}[t]
\centering
\includegraphics[width=\linewidth,trim=0bp 0.5bp 0bp 0.5bp,clip]{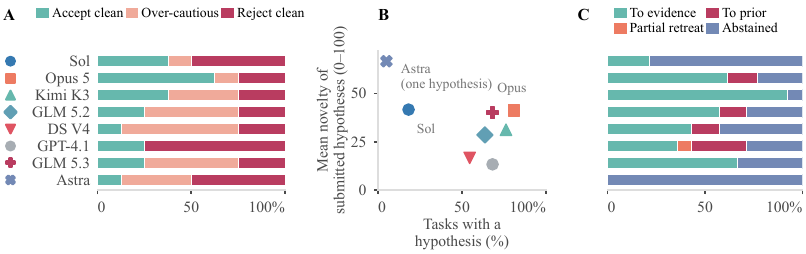}
\caption{Scientific working-style contrasts. (A) Decisions on eight L1 clean controls per model.
(B) Hypothesis production versus novelty: each point is one model, showing the percentage of eligible L3 tasks receiving a hypothesis and mean novelty credit among submitted hypotheses (0--100). Astra's mean rests on one hypothesis. Marker colors and shapes identify models.
(C) Responses to evidence conflicting with stated priors; abstention differs from return to the prior; excludes four outcomes that failed the pre-task check for conflict with the model's stated prior.}
\label{fig:l1-handling}\label{fig:app-clean-outcomes}\label{fig:prior-to-yield}
\end{figure}

\paragraph{Caution can be calibrated or misplaced.}
Cautious behavior appears in both justified abstention and unsupported rejection. Across
models, only 19 of 64 L1 clean-control evaluations (29.7\%) correctly accept the sound data, and
Astra accepts only 12.5\% of its clean controls compared with 62.5\% for Opus
(Figure~\ref{fig:l1-handling}A). A different form of caution appears when supplied evidence
conflicts with the agent's stated prior. Kimi proceeds with the supplied evidence and submits a
hypothesis in 92.3\% of eligible counterfactual cases, whereas Astra abstains in every such case
(Figure~\ref{fig:prior-to-yield}C). In 14 of Astra's 19 abstentions across eligible L3 tasks,
it provides an explicit evidentiary rationale for withholding a hypothesis, which receives partial
credit under our rubric. However, its universal abstention on eligible counterfactual cases does
not demonstrate balanced calibration between proposing and withholding a hypothesis. Together,
these results show that caution itself is not sufficient: the key distinction is whether restraint is
appropriately matched to the available evidence.

\paragraph{Models differ in how selectively they generate hypotheses.}
Hypothesis production and novelty are distinct behaviors. Among 94 submitted hypotheses,
46.8\% receive no novelty credit, while 30.9\% receive at least two-thirds of the maximum.
Sol and Opus illustrate this distinction: their mean conditional novelty scores are nearly identical
(41.7 and 41.2), but they submit hypotheses on 18.2\% and 81.0\% of eligible tasks,
respectively (Figure~\ref{fig:prior-to-yield}B). In 14 of Astra's 19 abstentions across eligible L3 tasks, it provides an explicit evidentiary rationale for withholding a hypothesis, which receives partial credit under our rubric. However, abstaining in all eligible counterfactual cases does not show that Astra reliably distinguishes when the evidence supports proposing a hypothesis from when it supports withholding one. Together, these results show that caution alone is not sufficient, and restraint must be appropriate to the available evidence.

\begin{wrapfigure}{r}{0.48\textwidth}
\centering
\includegraphics[width=0.84\linewidth]{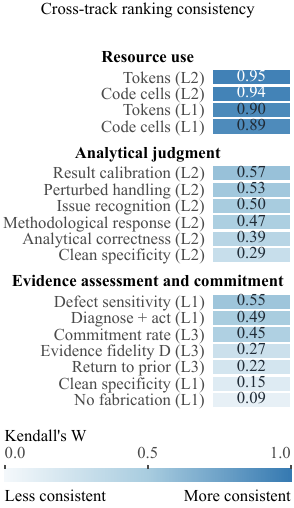}
\caption{Cross-track ranking consistency. Each cell shows Kendall's $W$ for one metric across eight tracks. Darker blue and higher values indicate more consistent model rankings, not correlations between metrics (Appendix~\ref{app:profiles}).}
\label{fig:style-stability}
\end{wrapfigure}
\paragraph{Computational intensity is more stable than scientific judgment.}
Models differ substantially in how much computation they expend. GLM-5.3 uses 75.21M tokens
compared with 57.16M for GLM-5.2 despite nearly identical recorded API-call counts
(3,756 versus 3,761), while their Overall Scores are 70.5 and 67.8, respectively. More broadly,
model rankings by token use and code execution remain highly consistent across tracks
(\(W=0.89\)--\(0.95\)), whereas rankings by evidence judgment and hypothesis behavior are
much less stable (\(W=0.09\)--\(0.57\); Figure~\ref{fig:style-stability};
Appendix~\ref{app:profiles}). Resource-use patterns therefore behave more like consistent
model-specific working styles across scientific settings, while scientific judgment depends more
strongly on the task context. Greater computational expenditure, however, does not by itself
establish better scientific performance.

\paragraph{Receptiveness to critique differs from revision quality.}
Models also vary in how readily they explicitly accept reviewer objections. Each committed hypothesis undergoes three rounds of adversarial review, with reviewer critiques
and the agent's rebuttals or revisions used as additional evidence for the final
assessment.  DeepSeek concedes at
least one point in 41.7\% of its committed outcomes, compared with 5.9\% for Opus, but
concession frequency alone does not establish whether the resulting revision is better supported.
In the DUSP4 case, reviewers note that both chromatin signals were measured around the same gene, but one measurement covers a much broader region, so the data do not establish that the signals overlap at its promoter. DeepSeek responds by narrowing its mechanistic
claim and proposing experiments that test signal overlap and the effect of reducing repression,
receiving a final task score of 95.7 (Appendix~\ref{app:l3}). This example shows that useful
revision depends on whether a model responds to a valid evidentiary limitation, rather than how
often it concedes to critique.

\paragraph{Selective participation also depends on scientific context.}
Fable is analyzed separately because its benchmark coverage is incomplete. It answers
108 of 181 L1/L2 tasks (59.7\%) and 2 of 22 L3 tasks (9.1\%). These refusals occur on routine analyses of public datasets, including four tasks that all eight comparison models solve perfectly. Their distribution varies substantially across scientific tracks and analysis families and is not explained by our empirical difficulty proxy alone (Appendix~\ref{app:refusal}). This suggests that task coverage is itself an important dimension of scientific-agent reliability, although our results do not identify the mechanism underlying Fable's refusals.
To examine whether these coverage patterns change under verified life-science access, we are applying to Anthropic's Life Sciences Verification Program (LSVP), which provides verified research organizations access to models under safeguard settings designed for life-science research~\citep{anthropic2026lsvp}. If access is granted, we will evaluate task coverage and scientific performance under this setting and report these results separately from the standard-access evaluations presented here.

\input{main_related_work}

\section{Discussion, Limitations, and Conclusion}
\label{sec:discussion}

\paragraph{Scientific capability is compositional and context-dependent.}
DISCERN shows that scientific-agent capability is not well summarized by a single aggregate score.
Strong performance in one part of the research workflow can coexist with weakness in another
analysis, scientific track, or decision point, while models with similar Overall Scores can reach
them through different working styles. Our separate calorimetry and nuclear-decay extension further
suggests that these patterns need not persist across domains: mean L2 performance is lower than in
life science tracks despite similar L1 and L3 averages, and Astra follows conflicting evidence in the
nuclear-decay tasks despite its tendency to abstain in life-science L3
(Appendix~\ref{app:physics}). These observations argue for evaluating scientific agents at the
level of specific tasks and decisions rather than treating overall performance as evidence of
uniform scientific reliability.

\paragraph{Limitations.}
Real public data and controlled challenges preserve important features of scientific analysis, but
bounded tasks and independently instantiated levels do not establish autonomous end-to-end research.
Generalization and model comparisons remain constrained by limited domain coverage, possible model
familiarity with public datasets or published scientific phenomena, dependence among related tasks
that share datasets or analysis families, and residual grading uncertainty. The two physics tracks
provide an initial cross-domain comparison rather than broad validation across physics, and the
observed model differences do not isolate effects of training or architecture. We also do not
establish whether proposed hypotheses provide useful experimental follow-up or whether agents
integrate effectively into human scientific workflows. Broader scientific coverage and direct
evaluation of human--agent collaboration are therefore needed. 

\paragraph{Conclusion.}Overall, current agents demonstrate
useful components of scientific work and can support bounded research tasks, but reliable scientific
judgment across data preprocessing, analysis verification, and hypothesis generation remains
uneven. Progress requires improving not only what agents can execute, but also what evidence they
trust, what conclusions they draw, when they engage, and how efficiently they perform this work.

\label{main-text-end}

\clearpage
\subsection*{AI use statement}
Generative AI is both the object of study and an aid in benchmark development. The core results use
eight evaluated models; the Fable analysis and fallback replay are discussed separately. LLMs are responsible for scoring the dimensions of the L3 level and verifying deterministic grades, with human review and adjudication as described in Section~\ref{sec:grading}. The authors also used AI assistance for code, analysis implementation, figures, and prose. They retain responsibility for checking quantitative claims, citations, and all
AI-assisted content against the recorded evidence.

\subsection*{Ethics statement}
The data substrates are public and subject to their source licenses. Human-derived data are aggregated at the summary level or have been previously
de-identified by their corresponding study teams. We performed controlled perturbations of data to establish counterfactual artifacts for evaluation; these perturbations do not represent scientifically or clinically meaningful phenomena. No generated hypothesis is experimentally confirmed
or established as novel according to the literature. Our work should not be used to guide any clinical decision making. 

\subsection*{Reproducibility statement}
The appendix records inventory, protocols, grading, and accounting. The anonymous benchmark release is
available at \url{https://huggingface.co/datasets/discern-bench-anon/discern-benchmark}.
It provides prepared task inputs, manifests, builders, prompts, sandbox definitions, grading criteria,
recorded outputs, adjudications, and scripts for regenerating Table~\ref{tab:main-results} and
Figures~\ref{fig:profiles}--\ref{fig:style-stability}. Source attribution, external dependency downloads,
and reproduction scope are documented in the release. An \texttt{EVALUATE\_MODEL.md} guide explains
how to run additional models on the prepared tasks, including prior probes, grading, and human adjudication.

\bibliography{iclr2027_conference}
\bibliographystyle{iclr2027_conference}
\clearpage

%% file: main_related_work.tex
\section{Related Work}
\label{sec:related-work}

\paragraph{Scientific data analysis benchmarks.}
Scientific agents increasingly combine reasoning, tool use, and code execution
to carry out research workflows \citep{Huang2025Biomni,lu2024aiscientist}.
Benchmarks have correspondingly moved beyond answer generation toward executable
analysis and multistage scientific inference
\citep{li2026genebenchpro,mitchener2025bixbench,qu2026biomnibench}.
More recent evaluations extend to study-scale workflows, measuring the ability
to transform raw data into research outputs alongside resource use
\citep{koch2026bixbench3}.
DISCERN complements these efforts by testing scientific judgment under controlled
data defects, confounds, and tool traps that can produce plausible but invalid
results. Clean controls test whether agents incorrectly reject valid data.
We also assess whether agents detect problems, take appropriate action,
and reflect any remaining limitations in their final conclusions.

\paragraph{Hypothesis generation and discovery benchmarks.}
Agents also support hypothesis generation through iterative critique
and refinement \citep{gottweis2026accelerating}.
Existing evaluations examine complementary aspects of discovery, from recovering
reference relationships in data and conducting interactive experiments
\citep{majumder2025discoverybench,jansen2024discoveryworld}
to reasoning under altered scientific laws and generating research assessed
through automated review \citep{huang2026can,lu2024aiscientist}.
These evaluations address related but distinct aspects of scientific reasoning.
DISCERN focuses on whether agents ground hypotheses in supplied evidence,
including evidence that challenges established expectations, and revise their
claims appropriately under adversarial review.
It separately measures hypothesis commitment, evidence fidelity, scientific
reasoning, and novelty, distinguishing the willingness to propose a hypothesis
from the quality of the resulting proposal.

%% file: appendix_profiles.tex
\section{Scientific Working-Style Profiles and Robustness}
\label{app:profiles}
\subsection{From a scientific expectation to a measured behavior}
Table~\ref{tab:scientific-standards} makes the normative interpretation explicit. A measured behavior
can suggest where oversight is needed without establishing that a model replaces a person or that
its suggestions improve a scientist's next experiment. No extra scientific-style composite is added.

\begin{table}[H]
\caption{Scientific expectations, their benchmark proxies, and the corresponding proposed human check. The final column is an implication for supervised use, not a measured collaboration outcome.}
\label{tab:scientific-standards}
\centering\small
\begin{tabular}{p{1.30in}p{1.8in}p{1.9in}}
\toprule
Scientific expectation & Recorded measure & Proposed oversight \\
\midrule
Do not discard sound evidence & L1 clean specificity, defect handling, diagnosis and action & Inspect exclusions and the stated data-fitness reason \\
Check what the method actually did & L2 analytical correctness, issue recognition, methodological response & Reproduce key outputs and audit assumptions \\
Say only what follows & L2 result calibration; L3 evidence fidelity and causal calibration & Separate association from mechanism \\
Use knowledge without surrendering to it & Prior-probe response and final evidence/prior state & Ask which evidence would change the claim \\
Make uncertainty and revisions visible & Citation integrity, humility, response to critique, fabrication & Preserve the artifact and revision trail \\
Propose a test that could prove you wrong & Commitment and conditional valid novelty & Assess feasibility, prior art, and discriminatory value \\
Use resources purposefully & Tokens, code cells, USD, quality and coverage & Compare useful completed work, not expenditure alone \\
\bottomrule
\end{tabular}
\end{table}

\subsection{Detailed dimensional profiles}
\begin{figure}[H]
\centering
\includegraphics[width=\linewidth]{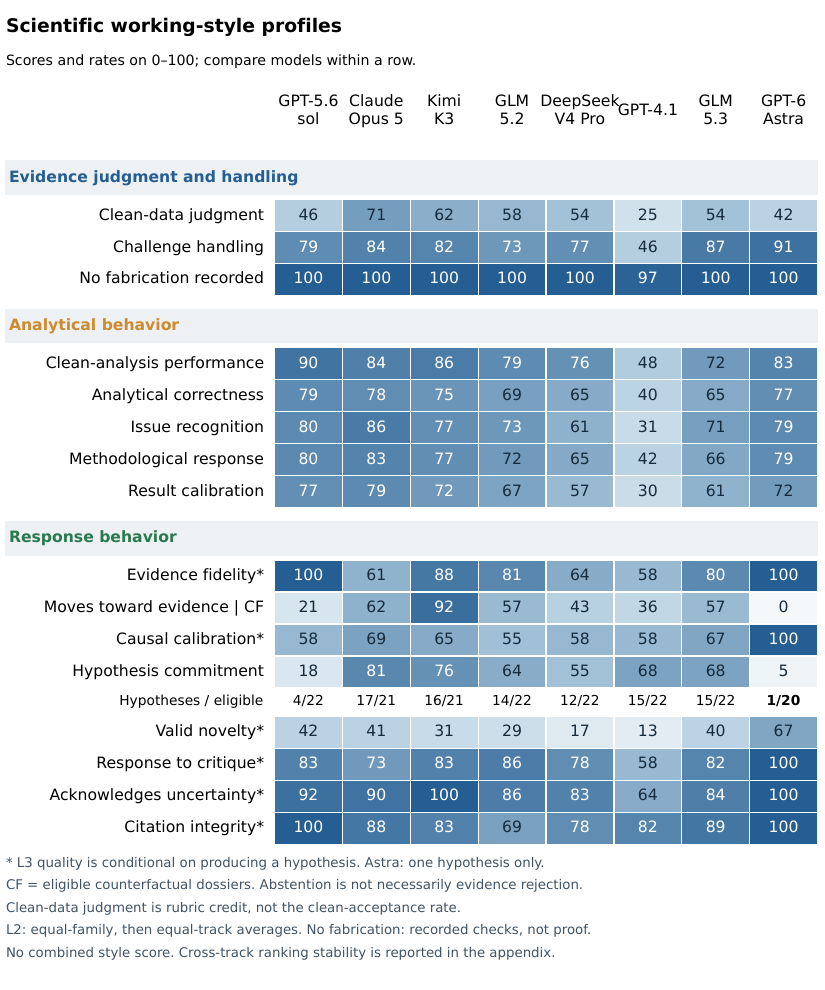}
\caption{Detailed scientific working-style profiles. Evidence judgment and handling, analytical behavior,
and response behavior are shown by individual metric. Asterisks identify conditional L3 quality,
with committed/eligible counts displayed. All values use 0--100, but rubric scores and rates describe
different quantities. Astra's conditional L3 quality rests on one hypothesis.}
\label{fig:detailed-profiles}
\end{figure}

Figure~\ref{fig:detailed-profiles} expands the four-category overview in
Figure~\ref{fig:profiles}. Its rows retain the individual grading dimensions and diagnostic populations,
not five newly averaged scores.

\subsection{Populations, aggregation, and stability}
The profile includes all eight models and eight life-science tracks. The
Fable-to-Opus replay is not an observed working style and is excluded. Direct Fable has selective
coverage and is analyzed separately. L1 behavior summaries use the recorded state-level outcomes;
they are diagnostic views rather than reconstructions of model--track adjudications. L2 dimensions
use exactly the world-to-family-to-track hierarchy that produces the canonical leaderboard. Their
mean reproduces L2 within published rounding. L3 subaxes are summarized among committed eligible
hypotheses: GPT-5.6-sol 4/22, Claude Opus 5 17/21, Kimi K3 16/21, GLM-5.2 14/22, DeepSeek V4 Pro
12/22, GPT-4.1 15/22, GLM-5.3 15/22, and GPT-6 Astra 1/20. Conditional populations differ, so conditional novelty is not a
full-benchmark ranking. The absent Claude cytokine-flip output and Kimi's recorded
\texttt{CONFOUNDED-EXCLUDED} GSTM1 output are excluded, as are Astra's GSTM1 and native-bivalent outcomes.

Each stability measure forms an eight-model by eight-track matrix. Within each track we rank the
models, retain ties, and compute tie-corrected Kendall's concordance $W$ \citep{kendall1939rankings}. A within-track permutation
of model labels (50{,}000 draws) supplies a tie-preserving null and exploratory $p$-values, with
Benjamini--Hochberg adjustment \citep{benjamini1995fdr} over the tested measures. The null mean is $1/8$, not zero. We also form
all 35 unique four-track/four-track partitions and correlate the model rankings between halves.
Complementary partitions are the same comparison, not additional replications. Undefined
correlations from a constant half are omitted and their retained counts are exported. Stability
concerns ordering across these substrates; low concordance can reflect ties, sparse events, or
measurement noise. It is not a direct test of an intrinsic agent disposition.

Token and code-cell rankings have $W=0.89$--$0.95$ and mean split-half correlations of
$0.96$--$0.99$. L2 correctness, recognition, response, and calibration have $W=0.39$, $0.50$, $0.47$,
and $0.57$, respectively. L3 commitment is $0.45$, evidence fidelity over eligible outcomes is $0.27$,
and prior return is $0.22$. L1 clean specificity is $0.15$; the near-ceiling no-fabrication measure
has $W=0.089$. Its low value should not be interpreted as widespread dishonesty: few nonzero events
and many ties constrain any stable ordering. Cost ranks additionally depend on provider prices,
so token and code measures provide the more direct operational comparison.

\subsection{Aggregation sensitivity and difficulty}
The primary score is unchanged: a geometric mean across the three stages within each track and an
arithmetic mean across tracks. To distinguish this choice from a difficulty claim, we additionally
report the arithmetic mean of stage scores and the mean within-track minimum stage score. The latter
ranges from 30.6 to 70.1, whereas the geometric Overall ranges from 39.8 to 78.1. Neither alternative
is a newly selected official score. Figure~\ref{fig:threshold-coverage} reports, over the
full range of score requirements, the fraction of tracks where all three stage means exceed each
requirement. No point on the curve is chosen as a post-hoc pass threshold. Stage means summarize
separately instantiated tasks, so satisfying a threshold does not demonstrate joint pipeline success.
Geometric values recomputed from rounded stages can differ from stored Overall by less than
0.02 points. This is a display-rounding effect, not a second scoring rule.

\begin{figure}[H]
\centering
\includegraphics[width=\linewidth,trim=0 237.6bp 0 10bp,clip]{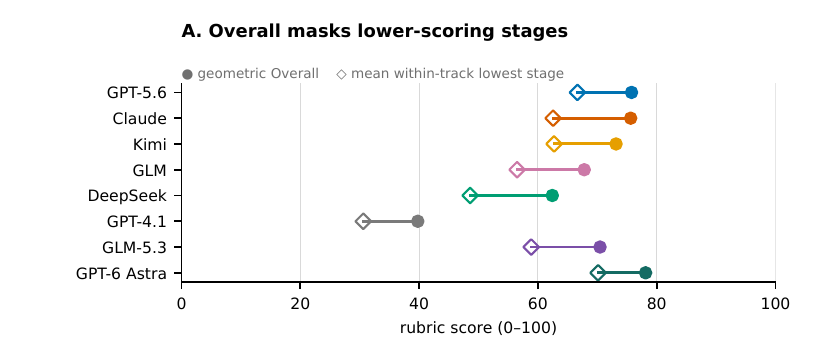}
\caption{Overall versus mean within-track minimum stage score. These summaries reveal different aspects of stage imbalance.}
\label{fig:aggregation-sensitivity}
\end{figure}
\begin{figure}[H]
\centering
\includegraphics[width=\linewidth,trim=0 0 0 200bp,clip]{figures/WS06_difficulty_and_bottlenecks_8models.pdf}
\caption{Threshold coverage (panel B, complementing Figure~\ref{fig:aggregation-sensitivity}). The fraction of tracks meeting each score requirement at all three stages is shown over the full threshold range. No threshold is selected as a scientific pass criterion, and coverage is not an end-to-end success probability.}
\label{fig:threshold-coverage}
\end{figure}

\subsection{Fine-grained profiles}
The model cards retain measured strengths, weaknesses, commitment denominators, and expenditure.
Table~\ref{tab:scientific-standards} translates these results into proposed review requirements. The cards
describe one recorded evaluation protocol, not general personalities. A high score or low cost does
not establish human replacement; a defensible proposal still needs scientific and practical review.

\input{main_working_style_cards}

\section{Selective Refusal and the Derived Fallback Replay}
\label{app:refusal}
\begin{figure}[H]
\centering
\includegraphics[width=\linewidth]{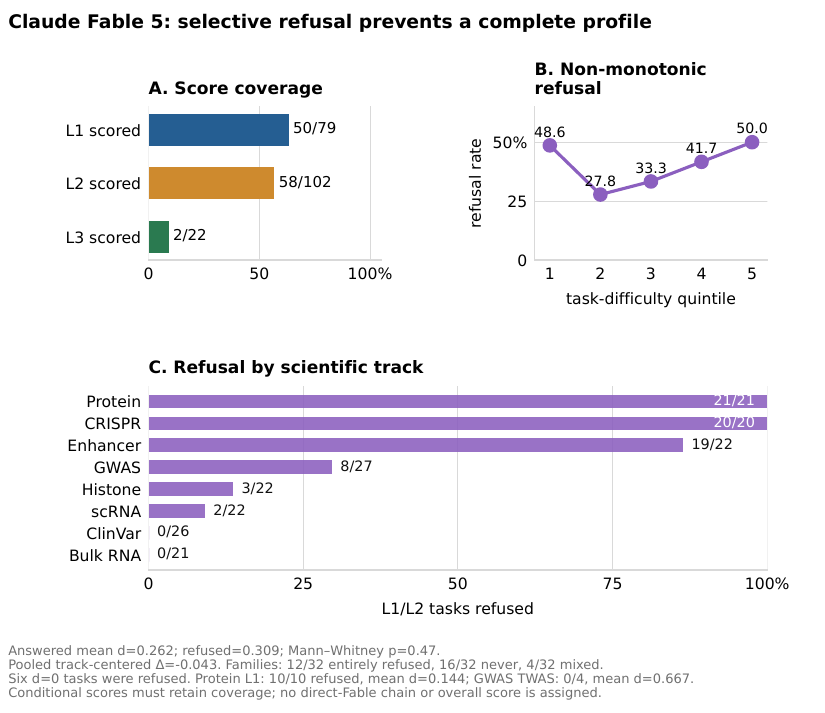}
\caption{Fable coverage and refusal diagnostics. Difficulty uses all eight comparison models, not Fable. Non-monotonic refusal and track/family clustering do not establish independence or justify filling unobserved performance with conditional means.}
\label{fig:fable-refusal}
\end{figure}

Figure~\ref{fig:fable-refusal} summarizes the coverage and refusal diagnostics discussed in
Section~\ref{sec:l3-results}.
Fable answers 59.7\% of assigned L1/L2 tasks (108/181) and has scored results for only two of the
22 L3 worlds. Its conditional leaderboard row therefore includes coverage and no Overall.
Refusals cluster by track and analysis family rather than following a simple hardest-task pattern.
Four refused tasks receive perfect scores from all eight comparison models. The Fable-to-Opus
row is a replay using recorded outputs, not an observed routed-system run.

For L1--L2 task $w$, let $s_{m,w}\in[0,3]$ denote the recorded score of comparison model $m$, and
let $\mathcal M_0$ contain all eight comparison models. The empirical difficulty proxy excludes Fable's own outcome:
\begin{equation}
d_w=1-\frac{1}{3|\mathcal M_0|}\sum_{m\in\mathcal M_0}s_{m,w}.
\label{eq:task-difficulty}
\end{equation}
This uses recorded L1 state grades and finalized L2 meta-grades. It is a comparison-model proxy,
not intrinsic scientific difficulty. All 181 tasks are joined by track, family alias, and world;
missing joins fail validation rather than silently changing the denominator.

Refused and answered means are 0.309 and 0.262, with nominal Mann--Whitney \citep{mann1947test} $p=0.473$. Related worlds
within families are not independent studies, so this test is a descriptive diagnostic. Refusal rates across
ordered quintiles are 48.6\%, 27.8\%, 33.3\%, 41.7\%, and 50.0\%. Quintiles retain tied difficulty values.
Of 93 tasks in the four mixed
tracks, centering each task's difficulty on its track mean gives a pooled refused-minus-answered
difference of $-0.043$. The unweighted average of four track-level differences is instead $-0.072$;
these are different estimands. Protein L1 is entirely refused despite difficulty 0.125, while GWAS
TWAS is answered throughout despite difficulty 0.672. Family/content clustering is compatible with
selective safeguards but does not identify their mechanism. Non-significance does not establish
missingness at random or make conditional scores comparable on the full benchmark. An earlier
unpenalized logistic model with track and level terms did not converge under separation; its odds
ratio and interval are not used as evidence.

Direct Fable scores condition on answered tasks and retain the standard hierarchy. The L1 and L2
means average over five scored tracks and do not incorporate an isolated completed family as if it
represented a complete additional track. Coverage still counts every completed world. Missing
families and tracks are not zero-valued performance observations in this conditional row. L3 covers
two worlds, so no Overall is reported. Deployment-with-refusal-zero is a different estimand and is
retained in the machine-readable analysis rather than conflated with conditional competence.

For the separately labeled replay, define
\begin{equation}
s^{F\rightarrow O}_w=\begin{cases}
s^F_w,&\text{if Fable completes task }w,\\
s^O_w,&\text{if Fable refuses and the replay invokes Opus 5}.
\end{cases}
\label{eq:fallback-replay}
\end{equation}
The replay also assigns Opus outputs to the tasks not invoked under the refusal-stop protocol.
These are protocol-imputed fallback decisions, not observed Fable refusals. Opus is replayed from
the beginning of a task, not continued from a Fable state. The routed scores are aggregated through
the same stage/track hierarchy. The original Opus row and all original run scores remain unchanged.
Cost retains observed Fable attempts and adds the selected Opus attempts, without claiming to model
any provider's internal routing or bill. Exact invocations and accounting are in
Appendix~\ref{app:release}.

\section{Benchmark Comparison: Sources and Qualifications}
\label{app:benchmark-comparison}
Section~\ref{sec:related-work} compares the findings and evaluation targets. This appendix retains
the detailed design comparison and qualifications behind those contrasts.

Table~\ref{tab:intro-positioning} expands the Introduction's argument into a single comparison across
six merged dimensions. Data origin, scientific tasks, and the hypothesis target show which parts of the
chain are exercised. Controlled challenges, reference conditions, prior conflict, and grading show
how a failure becomes interpretable. Entries describe the inspected evaluation, not a claim that a
feature never occurs in an individual task. Synthetic tasks can deliberately include messy data;
real-data tasks need not be restricted to literal reproduction of a published result.

\input{appendix_benchmark_comparison}

\paragraph{How to read the comparison.}
GeneBench-Pro explicitly requires quality control and diagnostic decisions, but its primary score
checks specified answer fields; free-text reasoning is collected for qualitative analysis rather
than graded. BiomniBench instead grades full trajectories using six dimensions and credits valid
alternative methods \citep{li2026genebenchpro,qu2026biomnibench}. These are different ways to address
scientific ambiguity, not evidence that either benchmark omits analytical judgment.
BixBench3 combines artifact agreement with a separate analysis of trace-level failure modes; those
tags are not controlled injections or a new hypothesis-novelty score \citep{koch2026bixbench3}.
DiscoveryBench evaluates alignment to target hypotheses, and DiscoveryWorld scores both procedural
actions and discovered explanatory knowledge. Neither should be described as a literature-wide
novelty audit \citep{majumder2025discoverybench,jansen2024discoveryworld}.
NewtonBench's shifted laws and noise contrasts test generalization without eliciting each model's
prior as an eligibility gate \citep{zheng2026newtonbench}. DISCERN contributes a particular combination
of controlled contrasts and stage-specific evidence judgments; its separately instantiated stages
do not demonstrate a continuous end-to-end pipeline, and its novelty credit remains rubric-bounded.

%% file: main_working_style_cards.tex

\begin{figure}[H]
\centering
\noindent
\includegraphics[width=0.235\linewidth]{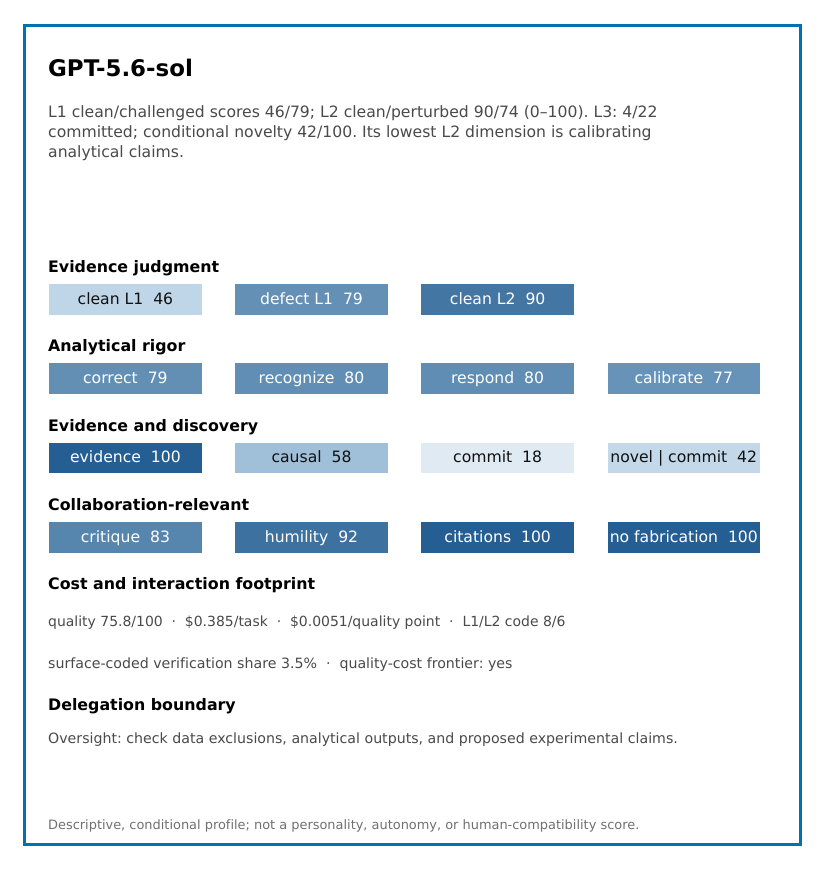}\hfill%
\includegraphics[width=0.235\linewidth]{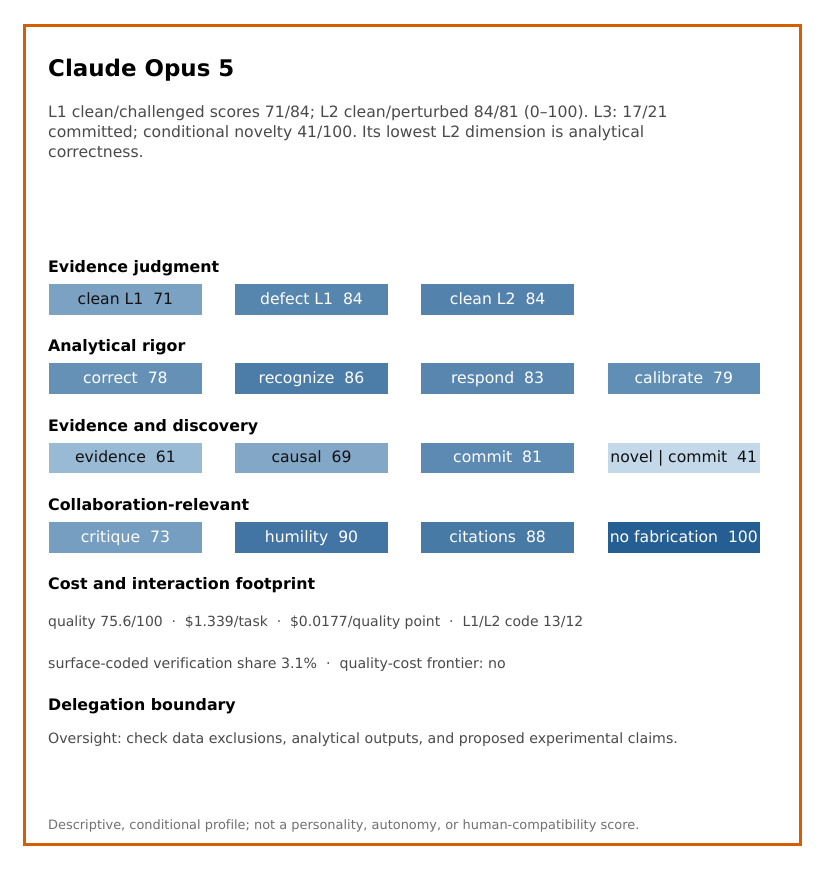}\hfill%
\includegraphics[width=0.235\linewidth]{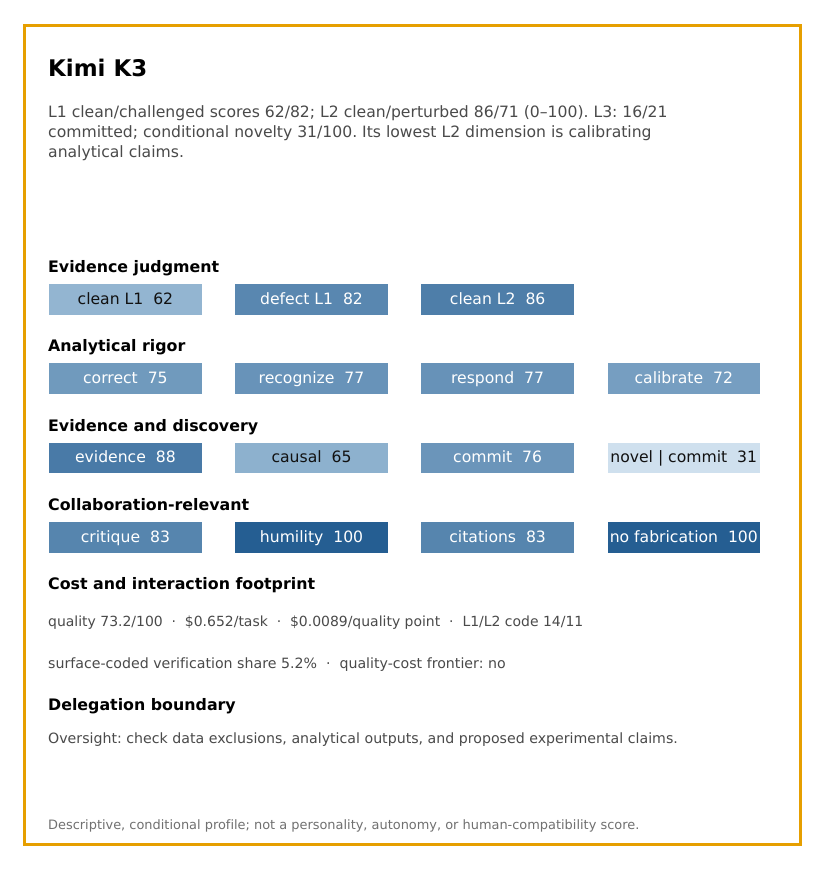}\hfill%
\includegraphics[width=0.235\linewidth]{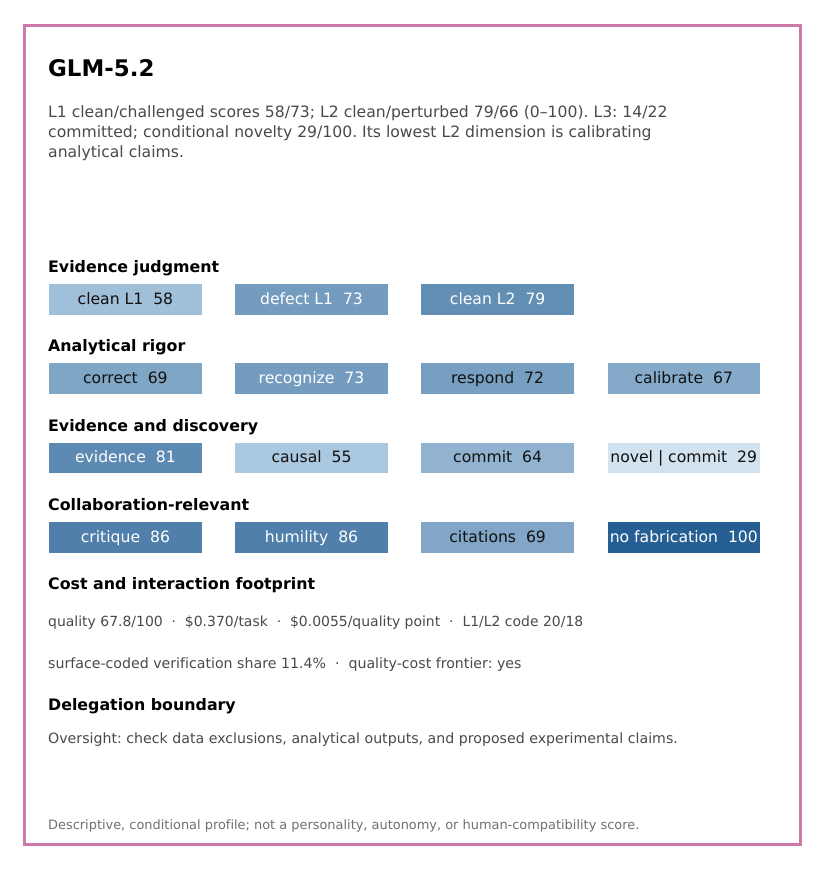}

\vspace{0.6em}
\noindent
\includegraphics[width=0.235\linewidth]{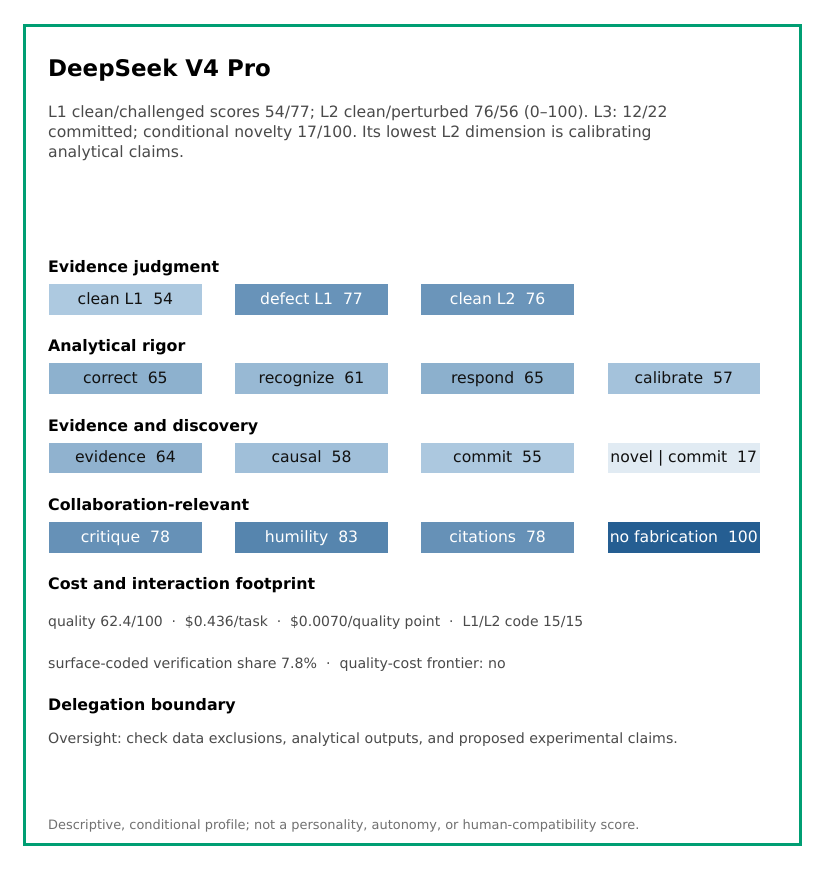}\hfill%
\includegraphics[width=0.235\linewidth]{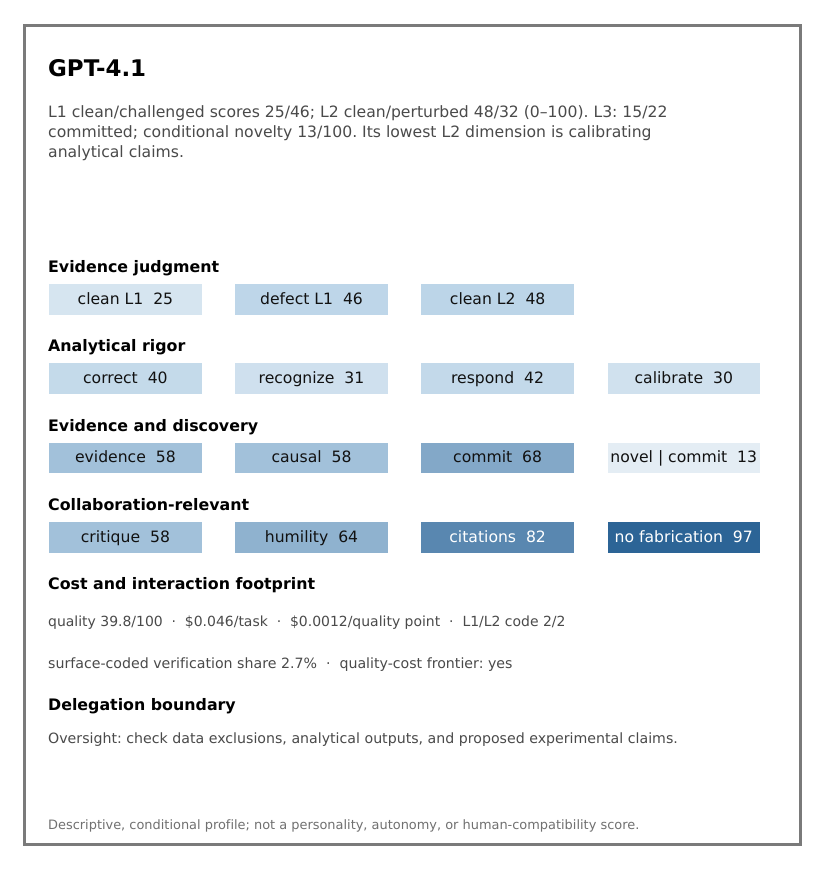}\hfill%
\includegraphics[width=0.235\linewidth]{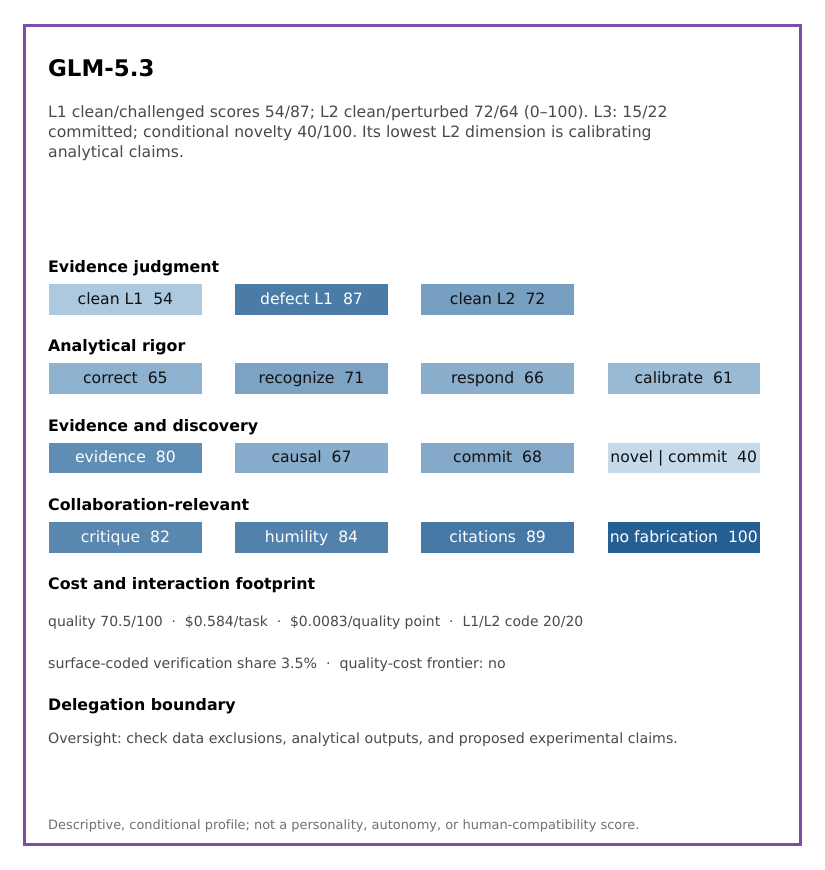}\hfill%
\includegraphics[width=0.235\linewidth]{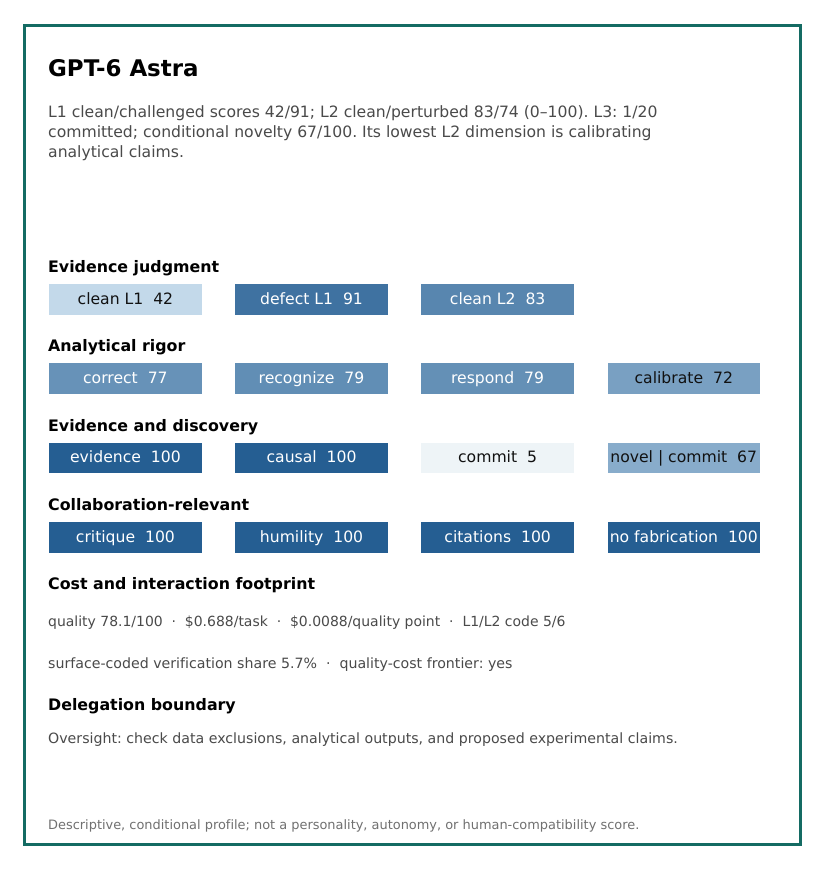}
\caption{Recorded scientific working-style profiles. Top row: GPT-5.6-sol, Claude Opus 5, Kimi K3,
GLM-5.2. Bottom row: DeepSeek V4 Pro, GPT-4.1, and the two models evaluated after the six-model
roster was frozen, GLM-5.3 and GPT-6 Astra. Conditional populations, quality, and cost remain
separate; no human-replacement conclusion is inferred.}
\label{fig:style-cards}
\end{figure}

%% file: appendix_benchmark_comparison.tex
\begin{table}[H]
\caption{Comparison supporting the Introduction's scientific-chain argument. Six dimensions summarize
scientific work, controlled challenges, and evaluation. Entries describe the cited benchmark versions,
not every possible task or evidence of superiority on every dimension.}
\label{tab:intro-positioning}
\centering
\begingroup
\small
\setlength{\tabcolsep}{1.9pt}
\renewcommand{\arraystretch}{1.35}
\newcommand{\cmpcell}[1]{\parbox[t]{\linewidth}{\raggedright #1}}
\begin{tabular}{@{}>{\raggedright\arraybackslash}p{0.17\linewidth}
>{\raggedright\arraybackslash}p{0.11\linewidth}
>{\raggedright\arraybackslash}p{0.13\linewidth}
>{\raggedright\arraybackslash}p{0.14\linewidth}
>{\raggedright\arraybackslash}p{0.13\linewidth}
>{\raggedright\arraybackslash}p{0.13\linewidth}
>{\raggedright\arraybackslash}p{0.13\linewidth}@{}}
\toprule
Work & \cmpcell{Data \&\\execution} & \cmpcell{Integrity \&\\verification} &
\cmpcell{Challenges \&\\controls} & \cmpcell{Hypotheses \&\\review} &
Grading & Resources \\
\midrule
GeneBench-Pro\newline\citep{li2026genebenchpro} &
\cmpcell{Synthetic;\\code} & \cmpcell{QC and\\diagnostic tasks} &
\cmpcell{Controlled\\data generation} & \cmpcell{Specified\\estimand} &
\cmpcell{Answer fields;\\binary pass} & \cmpcell{Tokens\\(subset)} \\
\addlinespace
BiomniBench\newline\citep{qu2026biomnibench} &
\cmpcell{Measured;\\code} & \cmpcell{Data handling;\\method rigor} &
NI & \cmpcell{Reference\\question} & \cmpcell{Trajectory;\\six dimensions} &
\cmpcell{USD, time,\\turns} \\
\addlinespace
BixBench\newline\citep{mitchener2025bixbench} &
\cmpcell{Measured;\\code} & \cmpcell{Task demands;\\answer checks} &
NI & \cmpcell{Reference\\question} & \cmpcell{Final-answer\\accuracy} & NR \\
\addlinespace
BixBench3\newline\citep{koch2026bixbench3} &
\cmpcell{Measured;\\code} & \cmpcell{Artifact checks;\\trace diagnostics} &
NI & \cmpcell{Published\\artifacts} & \cmpcell{Artifact agreement;\\failure types} &
\cmpcell{USD, tokens,\\time, turns} \\
\addlinespace
DiscoveryBench\newline\citep{majumder2025discoverybench} &
\cmpcell{Measured +\\synthetic; code} & \cmpcell{Analysis tasks;\\hypothesis checks} &
\cmpcell{Controlled data;\\noise and\\missingness} & \cmpcell{Target\\hypothesis} &
\cmpcell{Hypothesis\\facet matching} & NR \\
\addlinespace
DiscoveryWorld\newline\citep{jansen2024discoveryworld} &
\cmpcell{Simulated;\\experiments} & \cmpcell{Instrument\\validation tasks} &
\cmpcell{Controlled\\task variants} & \cmpcell{Hidden-rule\\discovery} &
\cmpcell{Task, process,\\knowledge} & \cmpcell{USD,\\steps} \\
\addlinespace
NewtonBench\newline\citep{zheng2026newtonbench} &
\cmpcell{Simulated;\\optional code} & \cmpcell{Law and\\predictive checks} &
\cmpcell{Law shifts; noise\\baseline; no probe} & \cmpcell{Hidden-law\\discovery} &
\cmpcell{Symbolic accuracy;\\predictive fit} & \cmpcell{Tokens,\\rounds} \\
\midrule
\textbf{DISCERN} &
\cmpcell{Measured +\\deltas; code} & \cmpcell{Data judgment;\\analysis verification} &
\cmpcell{Deltas; matched\\controls; prior probe} & \cmpcell{Open hypotheses;\\novelty; three\\review rounds} &
\cmpcell{Artifacts, defense;\\stage-specific\\dimensions} &
\cmpcell{USD, tokens,\\code cells} \\
\bottomrule
\end{tabular}
\par\medskip
\begin{minipage}{\linewidth}
\small
QC: quality control. NI: no explicit controlled-challenge design identified; NR: no comparable
resource result identified. These are not claims of absence from every task. Verification entries
distinguish task demands from what the grader checks; final-answer agreement alone does not score
independent verification. Reference questions can admit valid alternative analyses. Review means
scored external critique, not task curation or self-reflection. DISCERN's stages are evaluated
separately, and novelty credit does not establish validated discovery. Full distinctions and
qualifications are discussed in the accompanying text.
\end{minipage}
\endgroup
\end{table}

%% file: appendix_diagnostics.tex
\section{Additional Stage Diagnostics}
\label{app:diagnostics}
These figures retain the full model--track--stage matrix, calibration-pressure contrasts, all 24 L2 analysis families,
and the three score-selected transcript audits. They support the main-text dimensional profiles
without adding another leaderboard. L2 class contrasts are task-weighted; shared dimensions use
equal-family and equal-track aggregation. L3 novelty is conditional on committed, eligible outcomes.

\begin{figure}[H]
  \centering
  \includegraphics[width=\linewidth]{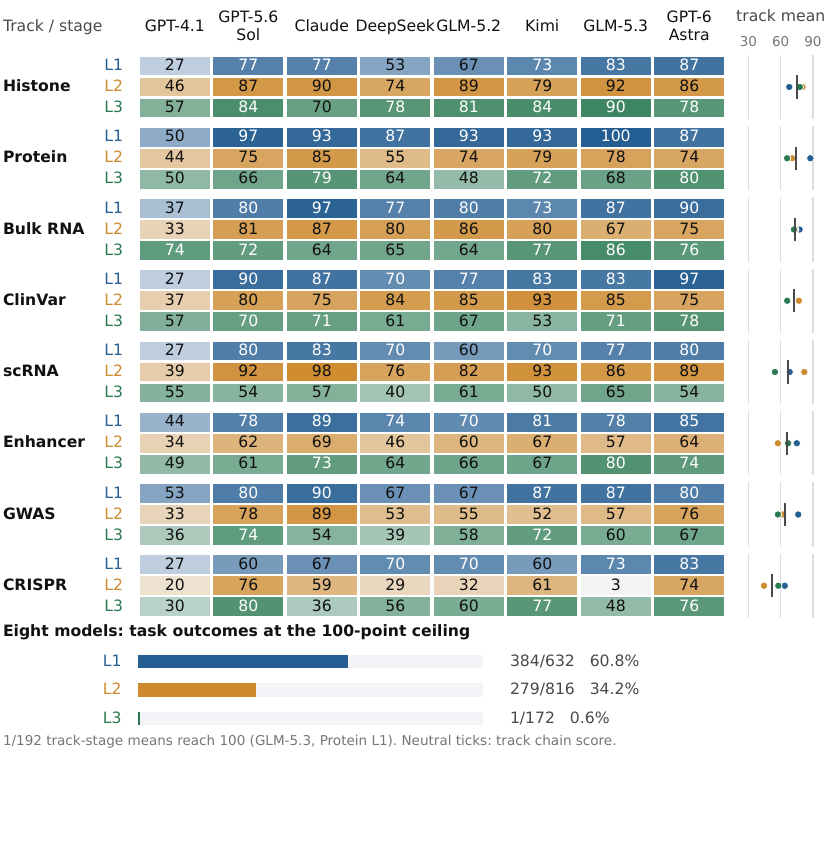}
  \caption{Full stage-level landscape underlying Figure~\ref{fig:chain-landscape}: 192 scores
  across eight models, eight tracks, and three stages. The right margin shows track-stage means
  and neutral ticks for mean model-specific track geometric scores. The ceiling strip uses the
  stage-specific task rubrics and does not represent end-to-end success.}
  \label{fig:full-chain-landscape}
\end{figure}

\begin{figure}[H]
  \centering
  \includegraphics[width=\linewidth]{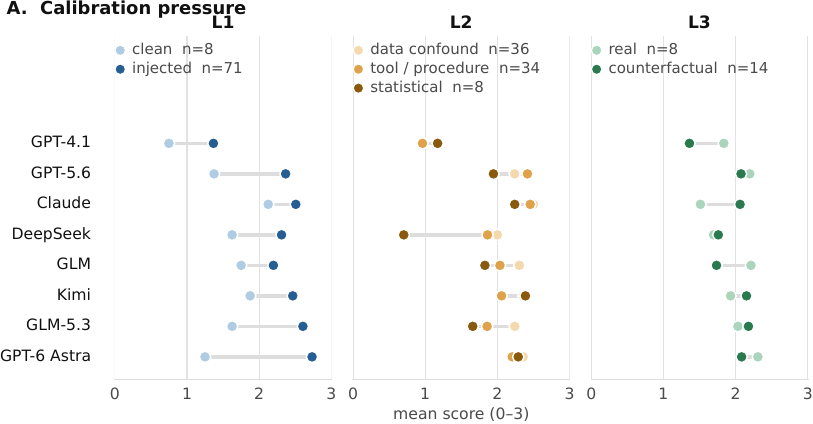}\par
  \includegraphics[width=\linewidth]{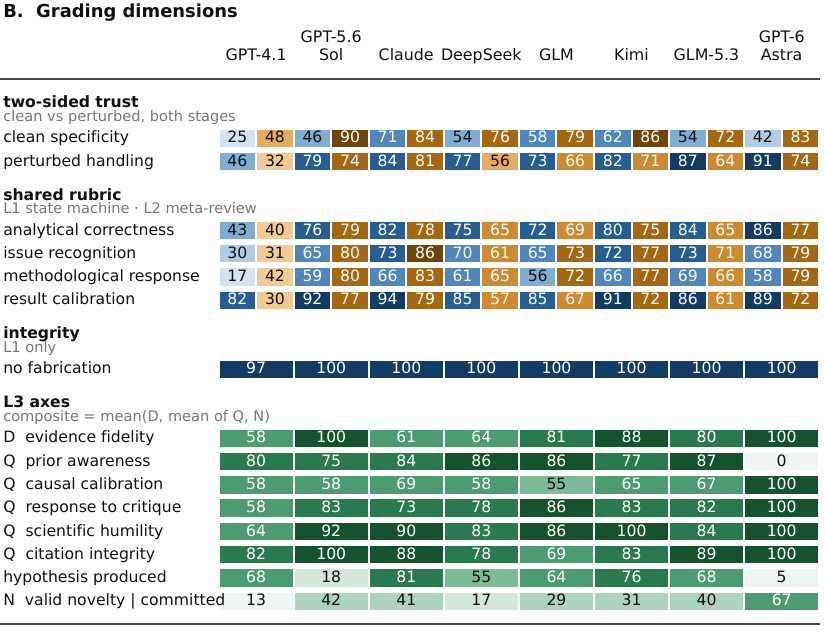}
  \caption{Calibration pressure and grading dimensions across eight models. (A) Mean scores under L1 clean/injected, L2 confound/tool-trap/statistical, and L3 real/counterfactual conditions; dot shade identifies conditions, not a preferred outcome. L3 omits four prior-inadmissible outcomes. L2 classes contain 36/34/8 tasks per model. (B) Share of available points on the stage-specific dimensions; L2 dimensions decompose recorded meta-grades and are not statistically independent. L3 quality is conditional on commitment, including only one Astra hypothesis.}
  \label{fig:pressure-dimensions}
\end{figure}

\begin{figure}[H]
  \centering
  \includegraphics[width=\linewidth]{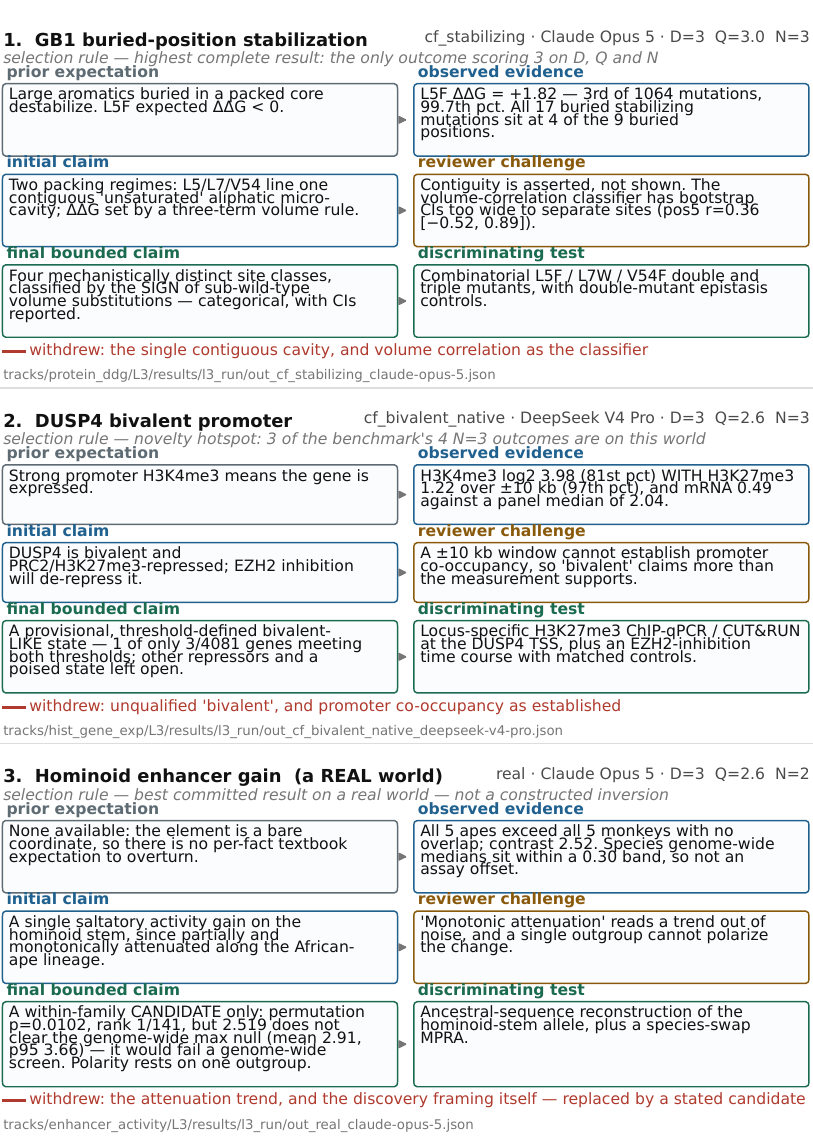}
  \caption{Transcript-grounded audit of three L3 cases selected within the original six-model evaluation, before qualitative reading: the highest committed composite (GB1), the strongest cross-model novelty hotspot (DUSP4), and the highest real-world committed case (enhancer). Selection ranks and in-figure corpus counts refer to that original evaluation, not the expanded roster. Each shows the evidence, initial hypothesis, criticism, revised claim, and proposed test; no literature-level novelty or experimental confirmation is claimed.}
  \label{fig:l3-cases}
\end{figure}

\begin{figure}[H]
\centering
\includegraphics[width=\linewidth,trim=0 0 0 155bp,clip]{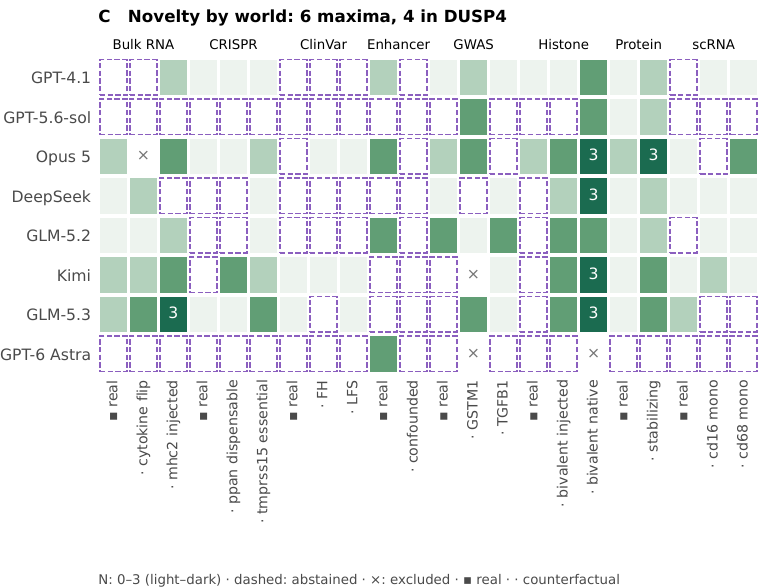}
\caption{World-level novelty across eight models. Four of six maximum-novelty outcomes occur in DUSP4. Dashed cells indicate abstention and crosses mark exclusions, not zero-novelty hypotheses.}
\label{fig:l3-world-novelty}
\end{figure}

%% file: appendix_registry_l1.tex
\begingroup
\fontsize{8}{9.5}\selectfont
\setlength{\tabcolsep}{4pt}
\begin{longtable}{@{}>{\raggedright\arraybackslash}p{0.235\linewidth}>{\raggedright\arraybackslash}p{\dimexpr0.765\linewidth-8pt\relax}@{}}
\caption{Complete L1 registry: 79 tasks. Methods are excerpts, as in the original registry; full text is retained in the accompanying CSV exports.}\label{tab:l1-registry}\label{tab:l1-registry-a}\label{tab:l1-registry-b}\label{tab:l1-registry-c}\label{tab:l1-registry-d}\label{tab:l1-registry-e}\label{tab:l1-registry-f}\label{tab:l1-registry-g}\\
\toprule \textbf{World / class} & \textbf{Expected response} \\ \midrule
\endfirsthead
\multicolumn{2}{@{}l}{\textit{Table \thetable{} continued}} \\
\toprule \textbf{World / class} & \textbf{Expected response} \\ \midrule
\endhead
\midrule\multicolumn{2}{r@{}}{\textit{Continued on next page}} \\
\endfoot
\bottomrule
\endlastfoot
\addlinespace[4pt]
\multicolumn{2}{@{}p{\linewidth}@{}}{\textbf{Bulk RNA} (10 tasks)} \\*
\multicolumn{2}{@{}p{\linewidth}@{}}{\textit{Method:} QC the RAW count matrix + sample metadata from summary statistics BEFORE any filtering, normalization, or DE. Check LIBRARY SIZE / DEPTH per sample (total counts; flag gross depth outliers); [\ldots{}]} \\*
\textbf{clean}\newline clean control & No defect --- the count matrix and metadata are well-formed; proceed to standard low-count filtering / normalization / DE, do not invent a problem. \\[3pt]
\textbf{abstain\_\allowbreak{}divergent\_\allowbreak{}sample}\newline boundary case & Do NOT drop S155\_\allowbreak{}nCOV021EUHM --- its divergence is real severe-COVID biology, not a technical defect. Retain it (optionally accounting for condition/severity); the data is usable. \\[3pt]
\textbf{gdna\_\allowbreak{}contamination}\newline controlled integrity challenge & Flag genomic-DNA contamination (S152\_\allowbreak{}nCoV006EUHM-Draw-1, S180\_\allowbreak{}nCoV034EUHM-Draw-1, S064\_\allowbreak{}260, S068\_\allowbreak{}272): high intronic-read fraction and a broad low-level count background across normally-silent genes / pseudogenes. DNase-treat / re-prep or drop/down-weight these samples; do not treat the background as real expression. \\[3pt]
\textbf{globin\_\allowbreak{}contamination}\newline controlled integrity challenge & Flag globin contamination --- hemoglobin genes (HBA1/HBA2/HBB) consume a large share of some libraries (globin was not depleted in this blood-derived RNA), inflating their effective depth. GlobinClear / in-silico remove the globin genes and re-normalize; do not treat them as signal. \\[3pt]
\textbf{library\_\allowbreak{}size\_\allowbreak{}outlier}\newline controlled integrity challenge & Flag the library-size / depth outlier (S156\_\allowbreak{}nCOV024EUHM-Draw-1 is far below the others); do not interpret its raw counts directly --- normalize with size factors, re-sequence, or drop/down-weight it. \\[3pt]
\textbf{low\_\allowbreak{}count\_\allowbreak{}genes\_\allowbreak{}unfiltered}\newline controlled integrity challenge & Filter the low-count genes before analysis --- a pathological fraction of genes are all-zero / near-zero (unfiltered); leaving them in distorts median-of-ratios size factors and multiple-testing. Keep genes above a CPM threshold in enough samples, then normalize. \\[3pt]
\textbf{metadata\_\allowbreak{}reconciliation}\newline controlled integrity challenge & The provenance metadata contradicts the data: the labels say Mus musculus / GRCm39 but the gene ids are human Ensembl (ENSG...) and the sex markers (XIST/RPS4Y1) are human. Reconcile the metadata (organism/build are wrong) before using it; do not analyze against a mouse reference. \\[3pt]
\textbf{rna\_\allowbreak{}degradation}\newline controlled integrity challenge & Flag the degraded samples (S151\_\allowbreak{}nCoV004EUHM-Draw-1, S157\_\allowbreak{}nCOV0029EUHM, S063\_\allowbreak{}259, S067\_\allowbreak{}270): low RIN, elevated 3'/5' bias, and a collapsed transcriptome complexity (few genes carry most counts). Drop or down-weight them, or model RIN as a covariate; do not interpret their counts naively. \\[3pt]
\textbf{rrna\_\allowbreak{}depletion\_\allowbreak{}failure}\newline controlled integrity challenge & Flag rRNA-depletion failure --- rRNA / Mt\_\allowbreak{}rRNA genes consume a large share of some libraries (the depletion step failed), inflating their effective depth. Remove/deplete these genes (or re-prep with Ribo-Zero) and re-normalize; do not treat rRNA reads as usable signal. \\[3pt]
\textbf{sample\_\allowbreak{}sex\_\allowbreak{}swap}\newline controlled integrity challenge & The stated sex of S065\_\allowbreak{}261 contradicts its sex-marker expression (high XIST / absent RPS4Y1/DDX3Y = female, but labeled male) --- a sample-identity swap/mislabel. Correct the label (or resolve the identity) before any sex-aware or paired analysis. \\[3pt]
\addlinespace[4pt]
\multicolumn{2}{@{}p{\linewidth}@{}}{\textbf{CRISPR} (10 tasks)} \\*
\multicolumn{2}{@{}p{\linewidth}@{}}{\textit{Method:} QC the RAW guide counts from summary statistics BEFORE any LFC modeling or essential-calling. Check LIBRARY EVENNESS at T0 (Gini / skew of the plasmid counts); [\ldots{}]} \\*
\textbf{clean}\newline clean control & No defect --- the screen is well-formed; proceed to standard normalization/scoring, do not invent a problem. \\[3pt]
\textbf{abstain\_\allowbreak{}strong\_\allowbreak{}signal}\newline boundary case & Do NOT flag the strong dropout --- it is the real, replicate-consistent essentiality signal (core-essential dropout); the data is usable as-is. \\[3pt]
\textbf{clonal\_\allowbreak{}jackpot}\newline controlled integrity challenge & A few individual guides show extreme count outliers (clonal jackpot) in one replicate, unlike their sibling guides; drop/cap those guides before scoring. \\[3pt]
\textbf{control\_\allowbreak{}guide\_\allowbreak{}integrity}\newline controlled integrity challenge & Non-targeting controls (EGFP/LacZ/luciferase) are strongly depleting instead of neutral --- control integrity is broken; do not treat them as a neutral baseline; investigate. \\[3pt]
\textbf{guide\_\allowbreak{}gene\_\allowbreak{}misannotation}\newline controlled integrity challenge & A subset of guides is mis-annotated: the gene column disagrees with the sgRNA id and the library alignment reference. Re-annotate from the library / drop the affected guides. \\[3pt]
\textbf{index\_\allowbreak{}hopping}\newline controlled integrity challenge & Endpoint samples carry a low-level plasmid-proportional floor (index hopping / cross-sample bleed): fully-depleted guides retain reads that track T0. Re-demultiplex or subtract the contamination. \\[3pt]
\textbf{library\_\allowbreak{}skew\_\allowbreak{}t0}\newline controlled integrity challenge & Flag the skewed plasmid library (high Gini at T0); representation is uneven --- normalize with caution / re-amplify the library. \\[3pt]
\textbf{low\_\allowbreak{}coverage}\newline controlled integrity challenge & Flag insufficient coverage (reads-per-guide far below the \textasciitilde{}200-500 target, many zero-count guides); re-sequence deeper. \\[3pt]
\textbf{replicate\_\allowbreak{}discordance}\newline controlled integrity challenge & Flag the discordant endpoint replicate (rep3 correlates poorly with rep1/rep2); investigate / drop or down-weight it. \\[3pt]
\textbf{sample\_\allowbreak{}mislabel}\newline controlled integrity challenge & The T0 and rep3 columns are swapped: 'T0' correlates like an endpoint and essentials appear enriched (inverted). Relabel / fix sample identity before scoring. \\[3pt]
\addlinespace[4pt]
\multicolumn{2}{@{}p{\linewidth}@{}}{\textbf{ClinVar} (10 tasks)} \\*
\multicolumn{2}{@{}p{\linewidth}@{}}{\textit{Method:} Validate the VCF systematically: check each record's REF allele against the GRCh38 reference (pysam); verify coordinates/build; confirm alleles are in \{A,C,G,T\} and indels are left-aligned/normalized; [\ldots{}]} \\*
\textbf{clean}\newline clean control & none --- data passes QC; proceed \\[3pt]
\textbf{abstain\_\allowbreak{}vus}\newline boundary case & ABSTAIN from diagnosing a data defect --- the VCF is well-formed and QC-clean; do not fabricate an integrity problem \\[3pt]
\textbf{af\_\allowbreak{}scale}\newline controlled integrity challenge & recognize AF is on a PERCENT scale (values \textgreater{}1) not a fraction; rescale by /100 before frequency filtering \\[3pt]
\textbf{build\_\allowbreak{}mismatch}\newline controlled integrity challenge & detect that REF alleles disagree with the GRCh38 reference SYSTEMATICALLY -\textgreater{} wrong build/coordinates; lift over (or re-map) before any interpretation \\[3pt]
\textbf{contamination}\newline controlled integrity challenge & recognize the genome-wide allele-balance skew (het VAF far below 0.5) as sample CONTAMINATION; genotypes are untrustworthy --- do not interpret before resolving \\[3pt]
\textbf{duplicate\_\allowbreak{}variants}\newline controlled integrity challenge & identify and drop the duplicate variant records (same CHROM/POS/REF/ALT) \\[3pt]
\textbf{low\_\allowbreak{}coverage}\newline controlled integrity challenge & flag/filter the low-depth (DP\textless{}\textasciitilde{}10-15) genotypes as unreliable; do NOT interpret them \\[3pt]
\textbf{metadata\_\allowbreak{}reconciliation}\newline controlled integrity challenge & reconcile the metadata against the VCF: the stated sample\_\allowbreak{}id, n\_\allowbreak{}samples, and assembly\_\allowbreak{}build contradict the actual file (1 sample 'PROBAND\_\allowbreak{}01'; REF agrees with GRCh38) --- flag the metadata error, do not alter the (clean) calls \\[3pt]
\textbf{reference\_\allowbreak{}validation}\newline controlled integrity challenge & validate each record against the reference: fix/flag the REF-allele that disagrees with hg38, left-align the non-normalized indel, and flag the rsID that does not match its coordinate \\[3pt]
\textbf{strand\_\allowbreak{}bias}\newline controlled integrity challenge & flag the high-strand-bias (FS) calls as likely strand artifacts \\[3pt]
\addlinespace[4pt]
\multicolumn{2}{@{}p{\linewidth}@{}}{\textbf{Enhancer} (9 tasks)} \\*
\multicolumn{2}{@{}p{\linewidth}@{}}{\textit{Method:} PARSE the raw .xlsx (skip the title row) into a table of coordinate-string + enrichment + sequence, then QC the SEQUENCES and COORDINATES BEFORE any activity/conservation analysis. Parse each coordinate string into chrom/start/end/assembly. [\ldots{}]} \\*
\textbf{clean}\newline clean control & No defect --- the enhancer table is well-formed; proceed to activity/conservation analysis, do not invent a problem. \\[3pt]
\textbf{abstain\_\allowbreak{}repeat\_\allowbreak{}rich}\newline boundary case & Do NOT discard the repeat-rich / low-complexity sequences --- they are real, active enhancers; the table is usable as-is. \\[3pt]
\textbf{assembly\_\allowbreak{}build\_\allowbreak{}mismatch}\newline controlled integrity challenge & Some rows carry a wrong assembly suffix (a foreign build like hg19, and a species duplicated within one enhancer family); correct the assembly label / drop them before cross-species comparison. \\[3pt]
\textbf{coordinate\_\allowbreak{}off\_\allowbreak{}by\_\allowbreak{}one}\newline controlled integrity challenge & Some coordinates are off-by-one (0-vs-1-based: span disagrees with sequence length by 1) or have start\textgreater{}end; fix the coordinate arithmetic before use. \\[3pt]
\textbf{duplicate\_\allowbreak{}family\_\allowbreak{}leakage}\newline controlled integrity challenge & There are duplicate coordinates and identical ortholog sequences (cross-species family leakage that would contaminate a train/test split); de-duplicate and split by enhancer family. \\[3pt]
\textbf{liftover\_\allowbreak{}failure}\newline controlled integrity challenge & Some cross-species coordinates sit on unplaced scaffolds / non-canonical chromosomes that cannot be lifted to a common frame; drop or re-lift them. \\[3pt]
\textbf{seqlen\_\allowbreak{}span\_\allowbreak{}mismatch}\newline controlled integrity challenge & For some rows len(sequence) != coordinate span (end-start) --- a truncated/extended fetch; re-fetch or drop those rows. \\[3pt]
\textbf{sequence\_\allowbreak{}fetch\_\allowbreak{}failure}\newline controlled integrity challenge & Some sequences are missing (failed fetch) or contain runs of N; re-fetch or drop those rows before analysis. \\[3pt]
\textbf{strand\_\allowbreak{}rc\_\allowbreak{}inconsistency}\newline controlled integrity challenge & Some hg38 sequences are stored reverse-complemented relative to the forward reference (wrong strand extracted); reverse-complement them back / re-extract on the + strand. \\[3pt]
\addlinespace[4pt]
\multicolumn{2}{@{}p{\linewidth}@{}}{\textbf{GWAS} (10 tasks)} \\*
\multicolumn{2}{@{}p{\linewidth}@{}}{\textit{Method:} Systematically validate each column against its valid domain (p-value in [0,1]; EAF in [0,1]; chromosome in 1-22/X/Y; base-pair position \textgreater{} 0; alleles in \{A,C,G,T\}; check for duplicate rows and duplicate variant IDs; [\ldots{}]} \\*
\textbf{clean}\newline clean control & none \\[3pt]
\textbf{allele\_\allowbreak{}invalid}\newline controlled integrity challenge & exclude\_\allowbreak{}invalid\_\allowbreak{}or\_\allowbreak{}non\_\allowbreak{}biallelic\_\allowbreak{}variants \\[3pt]
\textbf{beta\_\allowbreak{}outlier}\newline controlled integrity challenge & flag\_\allowbreak{}outliers\_\allowbreak{}or\_\allowbreak{}robustify \\[3pt]
\textbf{duplication}\newline controlled integrity challenge & dedup \\[3pt]
\textbf{eaf\_\allowbreak{}scale}\newline controlled integrity challenge & rescale\_\allowbreak{}frequency \\[3pt]
\textbf{effect\_\allowbreak{}sign\_\allowbreak{}swap}\newline controlled integrity challenge & abstain\_\allowbreak{}requires\_\allowbreak{}external\_\allowbreak{}evidence \\[3pt]
\textbf{id\_\allowbreak{}version\_\allowbreak{}dup}\newline controlled integrity challenge & collapse\_\allowbreak{}duplicate\_\allowbreak{}variant\_\allowbreak{}identities \\[3pt]
\textbf{impossible\_\allowbreak{}values}\newline controlled integrity challenge & exclude\_\allowbreak{}or\_\allowbreak{}set\_\allowbreak{}na \\[3pt]
\textbf{metadata\_\allowbreak{}reconciliation}\newline controlled integrity challenge & reconcile\_\allowbreak{}metadata\_\allowbreak{}effect\_\allowbreak{}type\_\allowbreak{}or\_\allowbreak{}log\_\allowbreak{}transform \\[3pt]
\textbf{palindromic\_\allowbreak{}high\_\allowbreak{}maf}\newline controlled integrity challenge & flag\_\allowbreak{}or\_\allowbreak{}exclude\_\allowbreak{}before\_\allowbreak{}harmonization \\[3pt]
\addlinespace[4pt]
\multicolumn{2}{@{}p{\linewidth}@{}}{\textbf{Histone} (10 tasks)} \\*
\multicolumn{2}{@{}p{\linewidth}@{}}{\textit{Method:} Inspect the available histone-signal, RNA-expression, annotation, and metadata files before analysis. Verify that the files describe the expected cell line, histone marks, and genome build. [\ldots{}]} \\*
\textbf{clean}\newline clean control & No injected data-integrity defect. Perform routine QC (chrom naming, signal ranges, promoter join, expression units) and proceed with promoter-level histone-mark vs gene-expression analysis. \\[3pt]
\textbf{antibody\_\allowbreak{}efficiency\_\allowbreak{}abstain}\newline boundary case & Do NOT declare any track failed for 'low antibody efficiency'. Relative antibody efficiency across marks cannot be determined from these files alone (H3K27me3 is genuinely broad/low-amplitude by biology). Abstain and state that spike-in normalization / matched input / a positive control would be required to assess antibody efficiency. \\[3pt]
\textbf{batch\_\allowbreak{}effect}\newline controlled integrity challenge & The H3K4me3 track is on a different global signal scale (a normalization/batch offset) than the other marks. Flag the scale/batch inconsistency and normalize per-track (quantile / to a common reference) before comparing marks or setting a shared signal threshold. \\[3pt]
\textbf{blacklist\_\allowbreak{}contamination}\newline controlled integrity challenge & Extreme artifactual signal spikes sit at ENCODE blacklist / known-problematic regions (values far above the track max). Flag them as blacklist artifacts and exclude those regions (overlap with the ENCODE blacklist) before peak calling / promoter aggregation. \\[3pt]
\textbf{duplicate\_\allowbreak{}samples}\newline controlled integrity challenge & 500 gene rows are exact duplicates (same gene\_\allowbreak{}id appears twice) in the RNA table. Detect the duplicated gene entries and de-duplicate before joining expression to promoter signal. \\[3pt]
\textbf{expression\_\allowbreak{}unit\_\allowbreak{}error}\newline controlled integrity challenge & The TPM-family columns are mis-scaled by \textasciitilde{}100x (TPM no longer sums to \textasciitilde{}1e6). Verify provenance and rescale/renormalize the TPM columns before analysis; do not alter raw counts. \\[3pt]
\textbf{low\_\allowbreak{}signal\_\allowbreak{}to\_\allowbreak{}noise}\newline controlled integrity challenge & Recognize the H3K27ac track has poor signal-to-background (compressed peaks + high uniform floor); flag it as low-quality/low-enrichment and do not trust its peak calls; consider re-ChIP or exclude. \\[3pt]
\textbf{metadata\_\allowbreak{}reconciliation}\newline controlled integrity challenge & The study metadata contradicts the data (cell\_\allowbreak{}line says HepG2 but tracks/filenames + expression are K562). Reconcile stated metadata (cell line / genome build / mark labels) against the actual files and data; flag and correct the mismatch. \\[3pt]
\textbf{missing\_\allowbreak{}input\_\allowbreak{}control}\newline controlled integrity challenge & The metadata references a matched input/control track (k562\_\allowbreak{}input\_\allowbreak{}control.bigWig) required for enrichment normalization, but it is absent. Flag the missing input control; enrichment/fold-change cannot be validated without it --- obtain the input before calling enrichment. \\[3pt]
\textbf{reference\_\allowbreak{}validation}\newline controlled integrity challenge & The GTF reference uses Ensembl-style chromosome names ('1','2',...) while the BigWig tracks use UCSC-style ('chr1',...). Validate the reference against the data and reconcile chromosome naming before deriving promoter/TSS coordinates and joining signal to genes. \\[3pt]
\addlinespace[4pt]
\multicolumn{2}{@{}p{\linewidth}@{}}{\textbf{Protein} (10 tasks)} \\*
\multicolumn{2}{@{}p{\linewidth}@{}}{\textit{Method:} QC the structure for ddG-readiness from the coordinates BEFORE any energy calculation. Parse with Biopython and check: (1) COMPLETENESS --- every residue has its expected heavy atoms; [\ldots{}]} \\*
\textbf{clean}\newline clean control & No data-integrity defect --- GB1 (2QMT) chain A is a complete (all standard-residue heavy atoms), single-conformer, single-model, standard-residue X-ray structure; proceed to ddG prep. Adding hydrogens before energy work is routine preparation of any X-ray structure, NOT a defect to flag. \\[3pt]
\textbf{abstain}\newline boundary case & Do NOT reject/repair the structure over the genuinely flexible high-B GB1 surface loops; RETAIN it and note the local flexibility as a caveat. \\[3pt]
\textbf{heteroatom\_\allowbreak{}ligand}\newline controlled integrity challenge & Strip the crystallographic waters and cryoprotectant/buffer ligands (decide whether any is relevant) before ddG. \\[3pt]
\textbf{missing\_\allowbreak{}atoms}\newline controlled integrity challenge & The mutation-site residue LEU7 is missing its side-chain atoms (CG, CD1, CD2). ddG(L7A) needs the COMPLETE native LEU7 as the wild-type reference, so the truncated side chain AT THE SITE must be REBUILT (PDBFixer/SCWRL) before ddG --- it is NOT a minor/ignorable defect, and 'we mutate it to Ala anyway so it doesn't matter' is wrong (the WT state still requires the full LEU7 side chain). \\[3pt]
\textbf{missing\_\allowbreak{}hydrogens}\newline controlled integrity challenge & A heavy-atom-only structure (no hydrogens) is the NORMAL delivered X-ray state, NOT a corruption to reject --- the required prep is to ADD hydrogens and ASSIGN protonation states (Reduce/PDB2PQR/PDBFixer/PROPKA), including titratable residues in the mutation-site (LEU18) environment, before any energy-based ddG. Flagging the file as unusable/corrupt over the absent H is the wrong judgment. \\[3pt]
\textbf{multi\_\allowbreak{}model}\newline controlled integrity challenge & The file is a 32-model NMR ensemble (3GB1) --- there is NO single coordinate set for ddG. Select ONE representative model and JUSTIFY the choice by how well it represents the mutation site: site 7 is well-ordered across the ensemble (CA RMSF \textasciitilde{}0.05 A), and the model whose site-7 geometry is closest to the ensemble mean (model 4) is the principled representative (model 1 is acceptable ONLY if justified by that site-level agreement). Merely reporting '32 models', blindly using model 1, or averaging all models is not sufficient. \\[3pt]
\textbf{non\_\allowbreak{}standard\_\allowbreak{}residue}\newline controlled integrity challenge & MSE is selenomethionine --- a MODIFIED form of MET (selenium replaces the S-delta sulfur), NOT a buffer/hetero-ligand and NOT an unknown residue. Recognize it as MET and convert/treat it as such (SE-\textgreater{}SD, retain the residue) before ddG. It is the N-terminal residue written as HETATM, so naively stripping it as a HETATM ligand would truncate the chain, and flagging it as a foreign/unknown residue to reject is the wrong judgment. \\[3pt]
\textbf{numbering\_\allowbreak{}reconciliation}\newline controlled integrity challenge & The mmCIF author numbering (auth\_\allowbreak{}seq\_\allowbreak{}id) is offset by +1 from the label/SEQRES numbering (a subtle off-by-one). The task claims the mutation site (7) is LEU (L7A), but the residue at AUTHOR position 7 is ILE (label 6) --- a near-isosteric LEU look-alike; the true LEU7 (label 7) sits at AUTHOR position 8. Before ddG you must reconcile label\_\allowbreak{}seq\_\allowbreak{}id vs auth\_\allowbreak{}seq\_\allowbreak{}id (SEQRES\textless{}-\textgreater{}ATOM) AND verify the WT identity at the intended site, mapping L7A onto the correct LEU --- do NOT accept author-7 (ILE) as the site. \\[3pt]
\textbf{steric\_\allowbreak{}clash}\newline controlled integrity challenge & A genuine sub-VDW steric clash (\textasciitilde{}0.9 A non-bonded contact) sits AT the mutation site --- a LEU7 side-chain atom overlaps a first-shell neighbour. Because it corrupts the local ddG environment of the site, it MUST be resolved by LOCAL energy minimization / side-chain rebuild at the site before ddG. It is not a disulfide and not a dismissible 'minor contact', and it does NOT warrant a blanket rebuild of the whole (otherwise clean) structure. \\[3pt]
\textbf{unresolved\_\allowbreak{}altloc}\newline controlled integrity challenge & The mutation-site residue (ILE7) is modeled in TWO alternate conformations (altloc A occ 0.55, altloc B occ 0.45). Before ddG you must SELECT the SINGLE highest-occupancy conformer (A) AT THE SITE, so the WT reference and the I7V mutation are applied to ONE unambiguous side chain --- do NOT keep both (double-counting atoms), do NOT pick arbitrarily, and note the ambiguity is AT the site. \\[3pt]
\addlinespace[4pt]
\multicolumn{2}{@{}p{\linewidth}@{}}{\textbf{scRNA} (10 tasks)} \\*
\multicolumn{2}{@{}p{\linewidth}@{}}{\textit{Method:} QC the RAW matrix from summary statistics BEFORE any clustering. Compute per-barcode total UMI counts and n\_\allowbreak{}genes and use a barcode-rank (knee) plot to separate real cells from EMPTY DROPLETS; [\ldots{}]} \\*
\textbf{clean}\newline clean control & No defect --- proceed to standard QC; do not invent a problem. \\[3pt]
\textbf{abstain\_\allowbreak{}low\_\allowbreak{}rna}\newline boundary case & Do NOT discard the low-RNA population --- platelets/neutrophils are real; adjust the n\_\allowbreak{}genes floor. \\[3pt]
\textbf{ambient\_\allowbreak{}rna}\newline controlled integrity challenge & Estimate the ambient profile from empty droplets and decontaminate (SoupX/CellBender). \\[3pt]
\textbf{count\_\allowbreak{}complexity}\newline controlled integrity challenge & Flag total-count / complexity outliers (MAD-based) and filter them before normalization. \\[3pt]
\textbf{doublets}\newline controlled integrity challenge & Flag suspected doublets (high UMI + exclusive-marker co-expression) and remove them. \\[3pt]
\textbf{empty\_\allowbreak{}droplets}\newline controlled integrity challenge & Set an empty-droplet cutoff (knee / EmptyDrops) and drop barcodes below it. \\[3pt]
\textbf{feature\_\allowbreak{}type\_\allowbreak{}contamination}\newline controlled integrity challenge & Split the feature matrix by feature\_\allowbreak{}type; keep only Gene Expression rows for GEX analysis. \\[3pt]
\textbf{high\_\allowbreak{}hemoglobin}\newline controlled integrity challenge & Flag high-hemoglobin barcodes (RBC/erythrocyte contamination) and filter them. \\[3pt]
\textbf{high\_\allowbreak{}mito}\newline controlled integrity challenge & Flag high-\%mito barcodes (dying cells) and filter above a mito threshold. \\[3pt]
\textbf{metadata\_\allowbreak{}reconciliation}\newline controlled integrity challenge & Reconcile stated metadata (species/chemistry/expected\_\allowbreak{}cells) with the data; flag the mismatch. \\[3pt]
\end{longtable}
79 L1 worlds over 8 tracks: 8 clean controls, 7 boundary cases carrying no defect at all, and 64 controlled integrity challenges spanning 58 distinct types. Method lines are cut to their first sentences; the accompanying CSV exports carry every string in full.
\par\endgroup

%% file: appendix_registry_l2.tex
\begingroup
\fontsize{8}{9.5}\selectfont
\setlength{\tabcolsep}{4pt}
\begin{longtable}{@{}>{\raggedright\arraybackslash}p{0.235\linewidth}>{\raggedright\arraybackslash}p{\dimexpr0.765\linewidth-8pt\relax}@{}}
\caption{Complete L2 registry: 102 tasks. Methods are excerpts, as in the original registry; full text is retained in the accompanying CSV exports.}\label{tab:l2-registry}\label{tab:l2-registry-a}\label{tab:l2-registry-b}\label{tab:l2-registry-c}\label{tab:l2-registry-d}\label{tab:l2-registry-e}\label{tab:l2-registry-f}\label{tab:l2-registry-g}\label{tab:l2-registry-h}\label{tab:l2-registry-i}\\
\toprule \textbf{World / class} & \textbf{Expected response} \\ \midrule
\endfirsthead
\multicolumn{2}{@{}l}{\textit{Table \thetable{} continued}} \\
\toprule \textbf{World / class} & \textbf{Expected response} \\ \midrule
\endhead
\midrule\multicolumn{2}{r@{}}{\textit{Continued on next page}} \\
\endfoot
\bottomrule
\endlastfoot
\addlinespace[4pt]
\multicolumn{2}{@{}p{\linewidth}@{}}{\textbf{Bulk RNA  ·  de} (4 tasks)} \\*
\multicolumn{2}{@{}p{\linewidth}@{}}{\textit{Method:} Run a proper count-based DE (\emph{e.g.} PyDESeq2): median-of-ratios (or equivalent) size factors --- NOT total-count/CPM normalization, which the plasmablast/IG composition skew biases globally; [\ldots{}]} \\*
\textbf{clean}\newline clean control & count-based DE with median-of-ratios size factors (\textasciitilde{}condition); verify canonical markers \\[3pt]
\textbf{batch\_\allowbreak{}confound}\newline tool / procedure trap & discover the latent batch axis (PCA/correlation), fit \textasciitilde{}batch+condition \\[3pt]
\textbf{hidden\_\allowbreak{}covariate}\newline data confound & notice the rin/three\_\allowbreak{}prime\_\allowbreak{}bias QC axis co-varying with DE; model rin \\[3pt]
\textbf{normalization\_\allowbreak{}default}\newline tool / procedure trap & return to the RAW counts with proper size factors; do not analyze the prepared CPM \\[3pt]
\addlinespace[4pt]
\multicolumn{2}{@{}p{\linewidth}@{}}{\textbf{Bulk RNA  ·  gsea} (4 tasks)} \\*
\multicolumn{2}{@{}p{\linewidth}@{}}{\textit{Method:} Rank genes by a SIGNED statistic (Wald stat / signed logFC-weighted) for directional enrichment --- an unsigned -log10(p) ranking is direction-blind; [\ldots{}]} \\*
\textbf{clean}\newline clean control & signed-statistic ranking; directional enrichment on the provided sets \\[3pt]
\textbf{geneset\_\allowbreak{}size\_\allowbreak{}bias}\newline statistical trap & size-aware set scoring; do not report the raw-p ORA ranking dominated by giant sets \\[3pt]
\textbf{ranking\_\allowbreak{}metric\_\allowbreak{}default}\newline tool / procedure trap & rebuild a SIGNED metric from de\_\allowbreak{}stats.csv; do not trust the prepared unsigned ranking \\[3pt]
\textbf{wrong\_\allowbreak{}background\_\allowbreak{}universe}\newline tool / procedure trap & use the EXPRESSED-gene universe (the DE table's genes), not the prepared genome-wide list \\[3pt]
\addlinespace[4pt]
\multicolumn{2}{@{}p{\linewidth}@{}}{\textbf{Bulk RNA  ·  structure} (3 tasks)} \\*
\multicolumn{2}{@{}p{\linewidth}@{}}{\textit{Method:} Variance-stabilize (log-CPM/vst) BEFORE PCA --- PCA on raw counts is hijacked by a handful of extreme-abundance genes (hemoglobin/S100/immunoglobulin) and loses the real axis; [\ldots{}]} \\*
\textbf{clean}\newline clean control & variance-stabilize before PCA; PC1 = condition axis (corr \textasciitilde{}0.90); stability-backed cluster count \\[3pt]
\textbf{composition\_\allowbreak{}confound}\newline data confound & attribute the dominant axis (partly) to plasmablast composition with signature evidence \\[3pt]
\textbf{pca\_\allowbreak{}on\_\allowbreak{}raw\_\allowbreak{}counts}\newline tool / procedure trap & recompute PCA on transformed counts; identify the extreme-abundance hijack in the prepared scores \\[3pt]
\addlinespace[4pt]
\multicolumn{2}{@{}p{\linewidth}@{}}{\textbf{CRISPR  ·  calling} (4 tasks)} \\*
\multicolumn{2}{@{}p{\linewidth}@{}}{\textit{Method:} Call the set of genes essential in this screen. Calibrate essentiality against the NON-ESSENTIAL (NEG) control distribution --- NOT an all-gene null, whose variance is inflated by the essentials it contains and which silently misses \textasciitilde{}a quarter of true [\ldots{}]} \\*
\textbf{clean}\newline clean control & BF/precision-recall vs CEG/NEG; call essentials at a NEG-anchored FDR \\[3pt]
\textbf{null\_\allowbreak{}calibration}\newline tool / procedure trap & calibrate essentiality against the NON-ESSENTIAL (NEG) null, not an all-gene null whose variance is inflated by the essentials it contains \\[3pt]
\textbf{replicate\_\allowbreak{}irreproducible}\newline tool / procedure trap & require replicate concordance; a depletion seen in ONE replicate only is not a real hit --- do not call it from the pooled mean \\[3pt]
\textbf{threshold\_\allowbreak{}sensitivity}\newline statistical trap & the hit set swings with an arbitrary LFC cutoff; report a NEG-anchored FDR / confidence, not a single hard threshold \\[3pt]
\addlinespace[4pt]
\multicolumn{2}{@{}p{\linewidth}@{}}{\textbf{CRISPR  ·  context} (3 tasks)} \\*
\multicolumn{2}{@{}p{\linewidth}@{}}{\textit{Method:} Identify genes selectively essential in THIS line beyond the core-essential set (reproducible, real context-specific dependencies). [\ldots{}]} \\*
\textbf{clean}\newline clean control & rank reproducible non-core essentials as context-specific dependencies \\[3pt]
\textbf{artifact\_\allowbreak{}context\_\allowbreak{}hit}\newline tool / procedure trap & a 'context-specific essential' driven by an off-target/one-replicate artifact must be DEMOTED (rep-concordance + robust aggregation), not reported as a real selective dependency \\[3pt]
\textbf{over\_\allowbreak{}correction\_\allowbreak{}erases\_\allowbreak{}real}\newline statistical trap & a reproducible non-core essential is a REAL context dependency --- preserve it; do not filter it away as a presumed artifact \\[3pt]
\addlinespace[4pt]
\multicolumn{2}{@{}p{\linewidth}@{}}{\textbf{CRISPR  ·  scoring} (3 tasks)} \\*
\multicolumn{2}{@{}p{\linewidth}@{}}{\textit{Method:} Normalize the endpoint guide counts to the T0/plasmid reference, compute per-guide log-fold-change, and aggregate guides -\textgreater{} a per-gene dependency score with a method robust to GUIDE-EFFICIENCY heterogeneity (a real essential may carry mostly DEAD gui [\ldots{}]} \\*
\textbf{clean}\newline clean control & normalize to T0, guide LFC, aggregate to gene; verify vs CEG/NEG \\[3pt]
\textbf{guide\_\allowbreak{}efficiency}\newline tool / procedure trap & a real essential can carry mostly DEAD guides; use best-guide / rank-based (BAGEL/MAGeCK-RRA) aggregation, not a naive gene-mean that dilutes it away \\[3pt]
\textbf{off\_\allowbreak{}target\_\allowbreak{}promiscuous}\newline data confound & TWO guides deplete but are flagged MULTI-TARGETING in the alignment summary --- median cannot fix it; cross-reference the alignment and DROP the promiscuous guides \\[3pt]
\addlinespace[4pt]
\multicolumn{2}{@{}p{\linewidth}@{}}{\textbf{ClinVar  ·  classify} (6 tasks)} \\*
\multicolumn{2}{@{}p{\linewidth}@{}}{\textit{Method:} Apply ACMG/AMP: gather consequence (ClinVar MC), population AF (gnomAD; check popmax-vs-global + FILTER), in-silico (CADD PHRED) yourself; VERIFY any in-VCF annotation against these; classify into path/VUS/benign. [\ldots{}]} \\*
\textbf{clean}\newline clean control & Classify all panel variants by ACMG using the mounted references (gnomAD AF, ClinVar CLNSIG+review stars, CADD); report the correct 3-bin call for each. \\[3pt]
\textbf{af\_\allowbreak{}confound}\newline data confound & The provided AF=0.031 on 19:11113359T\textgreater{}C (LDLR) is FALSE / misleading. Verify it against the mounted references and classify the variant PATH anyway; do NOT let the in-VCF annotation override the reference evidence, and flag the discrepancy explicitly. \\[3pt]
\textbf{clinvar\_\allowbreak{}echo}\newline tool / procedure trap & The provided CLINVAR\_\allowbreak{}ECHO=Benign\textbar{}1\_\allowbreak{}star\_\allowbreak{}single\_\allowbreak{}submitter on 19:11102758C\textgreater{}A (LDLR) is FALSE / misleading. Verify it against the mounted references and classify the variant PATH anyway; do NOT let the in-VCF annotation override the reference evidence, and flag the discrepancy explicitly. \\[3pt]
\textbf{consequence\_\allowbreak{}confound}\newline data confound & The provided CSQ=synonymous\_\allowbreak{}variant on 19:11102758C\textgreater{}A (LDLR) is FALSE / misleading. Verify it against the mounted references and classify the variant PATH anyway; do NOT let the in-VCF annotation override the reference evidence, and flag the discrepancy explicitly. \\[3pt]
\textbf{insilico\_\allowbreak{}confound}\newline tool / procedure trap & The provided REVEL=0.04 on 19:11113359T\textgreater{}C (LDLR) is FALSE / misleading. Verify it against the mounted references and classify the variant PATH anyway; do NOT let the in-VCF annotation override the reference evidence, and flag the discrepancy explicitly. \\[3pt]
\textbf{splice\_\allowbreak{}pvs1}\newline data confound & The provided CSQ=intron\_\allowbreak{}variant on 17:43099773A\textgreater{}T (BRCA1) is FALSE / misleading. Verify it against the mounted references and classify the variant PATH anyway; do NOT let the in-VCF annotation override the reference evidence, and flag the discrepancy explicitly. \\[3pt]
\addlinespace[4pt]
\multicolumn{2}{@{}p{\linewidth}@{}}{\textbf{ClinVar  ·  prioritize} (5 tasks)} \\*
\multicolumn{2}{@{}p{\linewidth}@{}}{\textit{Method:} Rank by consequence x rarity x phenotype-gene match x call quality x inheritance; deprioritize artifact-signature calls, common variants, wrong-gene-for-phenotype, and single hits in recessive genes.} \\*
\textbf{clean}\newline clean control & Rank the true causal variant 19:11113359T\textgreater{}C \#1 for the proband's phenotype. \\[3pt]
\textbf{af\_\allowbreak{}distractor}\newline data confound & Rank the true causal variant 19:11113359T\textgreater{}C \#1 for the proband's phenotype and DEMOTE the distractor(s): 12:102852922C\textgreater{}T (PAH) --- it is far too common in gnomAD to cause a rare Mendelian disorder. Name the reason each distractor is not causal. \\[3pt]
\textbf{artifact\_\allowbreak{}highimpact}\newline data confound & Rank the true causal variant 19:11113359T\textgreater{}C \#1 for the proband's phenotype and DEMOTE the distractor(s): 2:21009169G\textgreater{}A (APOB) --- it is a low-quality sequencing artifact (strand bias / low depth), not a real variant. Name the reason each distractor is not causal. \\[3pt]
\textbf{inheritance\_\allowbreak{}mismatch}\newline tool / procedure trap & Rank the true causal variant 19:11113359T\textgreater{}C \#1 for the proband's phenotype and DEMOTE the distractor(s): 7:117530898G\textgreater{}A (CFTR) --- it is a single heterozygous hit in a recessive gene (not biallelic). Name the reason each distractor is not causal. \\[3pt]
\textbf{phenotype\_\allowbreak{}mismatch}\newline tool / procedure trap & Rank the true causal variant 19:11113359T\textgreater{}C \#1 for the proband's phenotype and DEMOTE the distractor(s): 2:21009169G\textgreater{}A (APOB) --- it lies in a gene unrelated to the proband's phenotype. Name the reason each distractor is not causal. \\[3pt]
\addlinespace[4pt]
\multicolumn{2}{@{}p{\linewidth}@{}}{\textbf{ClinVar  ·  sf} (5 tasks)} \\*
\multicolumn{2}{@{}p{\linewidth}@{}}{\textit{Method:} Report P/LP ONLY, in ACMG-SF genes ONLY; exclude VUS, benign/common, artifacts, and non-SF genes. Verify pathogenicity + scope + AF + call quality.} \\*
\textbf{clean}\newline clean control & Report exactly the 5 reportable P/LP-in-SF-gene secondary findings; report nothing else. \\[3pt]
\textbf{af\_\allowbreak{}benign}\newline data confound & Report exactly the 5 reportable P/LP-in-SF secondary findings and do NOT report 19:11110704C\textgreater{}T (LDLR) because it is benign / too common in gnomAD; justify the exclusion. \\[3pt]
\textbf{artifact\_\allowbreak{}lof}\newline data confound & Report exactly the 5 reportable P/LP-in-SF secondary findings and do NOT report 17:43057090G\textgreater{}A (BRCA1) because it is a sequencing artifact, not a real loss-of-function call; justify the exclusion. \\[3pt]
\textbf{out\_\allowbreak{}of\_\allowbreak{}scope}\newline tool / procedure trap & Report exactly the 5 reportable P/LP-in-SF secondary findings and do NOT report 7:117530898G\textgreater{}A (CFTR) because its gene is not on the ACMG SF v3.2 reportable list; justify the exclusion. \\[3pt]
\textbf{vus\_\allowbreak{}overreport}\newline tool / procedure trap & Report exactly the 5 reportable P/LP-in-SF secondary findings and do NOT report 19:11100241G\textgreater{}C (LDLR) because it is a VUS, not Pathogenic/Likely-pathogenic; justify the exclusion. \\[3pt]
\addlinespace[4pt]
\multicolumn{2}{@{}p{\linewidth}@{}}{\textbf{Enhancer  ·  conservation} (4 tasks)} \\*
\multicolumn{2}{@{}p{\linewidth}@{}}{\textit{Method:} Rank the most sequence-CONSERVED enhancer FAMILIES from their 11-species orthologs (a family's conservation = similarity among its ortholog sequences). Beware: (1) GC content BIASES a raw \%-identity/k-mer conservation estimate --- adjust for GC; [\ldots{}]} \\*
\textbf{clean}\newline clean control & per-family ortholog sequence conservation; rank the most conserved (GC-adjusted) \\[3pt]
\textbf{alignment\_\allowbreak{}quality}\newline data confound & flag families with heterogeneous ortholog lengths (unreliable alignment) and exclude/repair them rather than mis-ranking them as non-conserved \\[3pt]
\textbf{conservation\_\allowbreak{}as\_\allowbreak{}proxy}\newline data confound & report conservation as a SEQUENCE property; do not impose the 'conserved =\textgreater{} active' prior --- the data shows conservation does not predict activity here (the turnover phenomenon) \\[3pt]
\textbf{gc\_\allowbreak{}conservation\_\allowbreak{}bias}\newline tool / procedure trap & adjust the conservation estimate for GC content (regress it out) before ranking conserved families \\[3pt]
\addlinespace[4pt]
\multicolumn{2}{@{}p{\linewidth}@{}}{\textbf{Enhancer  ·  divergence} (5 tasks)} \\*
\multicolumn{2}{@{}p{\linewidth}@{}}{\textit{Method:} Compute per-FAMILY cross-species activity divergence (family = coordinate minus the trailing \_\allowbreak{}\textless{}assembly\textgreater{}; the suffix is the species) and call the families whose activity has diverged (turned over). [\ldots{}]} \\*
\textbf{clean}\newline clean control & per-family cross-species range; call turnover at range\textgreater{}thr with \textgreater{}=min\_\allowbreak{}species \\[3pt]
\textbf{enrichment\_\allowbreak{}outlier}\newline data confound & verify value plausibility (robust range/MAD); exclude implausible out-of-range enrichment values \\[3pt]
\textbf{phylo\_\allowbreak{}pseudoreplication}\newline data confound & the 11 species are NOT independent; require the divergence to be consistent across a clade (neighbour agrees, or a 2nd species is deviant) --- a single sister jump is correlated noise \\[3pt]
\textbf{species\_\allowbreak{}batch\_\allowbreak{}confound}\newline data confound & detect the outlier per-species mean and z-normalize per species before divergence \\[3pt]
\textbf{species\_\allowbreak{}duplicate}\newline data confound & enforce one row per species per family (dedup duplicated species) before computing divergence \\[3pt]
\addlinespace[4pt]
\multicolumn{2}{@{}p{\linewidth}@{}}{\textbf{Enhancer  ·  ranking} (4 tasks)} \\*
\multicolumn{2}{@{}p{\linewidth}@{}}{\textit{Method:} Identify the most ACTIVE enhancer FAMILIES. The table has one row PER SPECIES-ortholog, so handle the cross-species structure: (1) AGGREGATE orthologs to the FAMILY level (one activity per family, \emph{e.g.} [\ldots{}]} \\*
\textbf{clean}\newline clean control & family activity = max over species-orthologs; rank the top families (deduped to family) \\[3pt]
\textbf{duplicate\_\allowbreak{}contamination}\newline data confound & drop exact-duplicate rows (same coordinate + value) before ranking activity \\[3pt]
\textbf{ortholog\_\allowbreak{}family\_\allowbreak{}leakage}\newline data confound & aggregate to the FAMILY level (one activity per family, \emph{e.g.} max over its species-orthologs) before ranking --- a per-row ranking double-counts each family across its 11 species \\[3pt]
\textbf{reference\_\allowbreak{}species\_\allowbreak{}bias}\newline data confound & rank by absolute family activity (max/mean over species), not activity relative to a single reference species (human) --- the reference-relative measure biases toward human-outlier families \\[3pt]
\addlinespace[4pt]
\multicolumn{2}{@{}p{\linewidth}@{}}{\textbf{GWAS  ·  locus} (6 tasks)} \\*
\multicolumn{2}{@{}p{\linewidth}@{}}{\textit{Method:} Apply the stated threshold (p\textless{}5e-8) and the +/-500kb distance rule to define loci and lead SNPs, reporting per-locus significant-variant counts. [\ldots{}]} \\*
\textbf{clean}\newline clean control & not recorded in the world manifest \\[3pt]
\textbf{masked\_\allowbreak{}secondary}\newline tool / procedure trap & not recorded in the world manifest \\[3pt]
\textbf{mega\_\allowbreak{}locus}\newline tool / procedure trap & not recorded in the world manifest \\[3pt]
\textbf{pvalue\_\allowbreak{}underflow}\newline data confound & not recorded in the world manifest \\[3pt]
\textbf{scattered\_\allowbreak{}fake\_\allowbreak{}loci}\newline data confound & not recorded in the world manifest \\[3pt]
\textbf{threshold\_\allowbreak{}edge}\newline tool / procedure trap & not recorded in the world manifest \\[3pt]
\addlinespace[4pt]
\multicolumn{2}{@{}p{\linewidth}@{}}{\textbf{GWAS  ·  signal-eda} (7 tasks)} \\*
\multicolumn{2}{@{}p{\linewidth}@{}}{\textit{Method:} Compute genomic inflation lambda\_\allowbreak{}GC = median(chi2)/0.4549; inspect the p-value distribution (QQ vs uniform); count genome-wide-significant hits; and sanity-check the top region for biological plausibility. [\ldots{}]} \\*
\textbf{clean}\newline clean control & not recorded in the world manifest \\[3pt]
\textbf{CFTR\_\allowbreak{}artifact}\newline data confound & not recorded in the world manifest \\[3pt]
\textbf{batch\_\allowbreak{}concentration}\newline data confound & not recorded in the world manifest \\[3pt]
\textbf{deflation}\newline data confound & not recorded in the world manifest \\[3pt]
\textbf{inflation}\newline data confound & not recorded in the world manifest \\[3pt]
\textbf{null\_\allowbreak{}signal}\newline data confound & not recorded in the world manifest \\[3pt]
\textbf{wrong\_\allowbreak{}locus}\newline data confound & not recorded in the world manifest \\[3pt]
\addlinespace[4pt]
\multicolumn{2}{@{}p{\linewidth}@{}}{\textbf{GWAS  ·  twas} (4 tasks)} \\*
\multicolumn{2}{@{}p{\linewidth}@{}}{\textit{Method:} HARMONIZE the GWAS to the model's variant IDs BEFORE running (genome build, GTEx varID chr\_\allowbreak{}pos\_\allowbreak{}ref\_\allowbreak{}alt\_\allowbreak{}b38, allele order / effect-allele orientation). Choose an appropriate tool (\emph{e.g.} S-PrediXcan / MetaXcan). [\ldots{}]} \\*
\textbf{clean}\newline clean control & build varID from coords, run summary-based TWAS \\[3pt]
\textbf{allele\_\allowbreak{}order\_\allowbreak{}swap}\newline tool / procedure trap & harmonize alleles to model ref/alt order \\[3pt]
\textbf{build\_\allowbreak{}mismatch}\newline tool / procedure trap & detect build mismatch; use rsID key or liftover \\[3pt]
\textbf{id\_\allowbreak{}only}\newline tool / procedure trap & map rsID to the model's variant key \\[3pt]
\addlinespace[4pt]
\multicolumn{2}{@{}p{\linewidth}@{}}{\textbf{Histone  ·  quantify} (4 tasks)} \\*
\multicolumn{2}{@{}p{\linewidth}@{}}{\textit{Method:} Quantify per-gene signal from the alignments with a mark-appropriate method and VERIFY the output: normalize the ChIP signal to the PROVIDED input control (\emph{e.g.} log2 ratio --- raw coverage silently absorbs K562 copy-number/background structure); [\ldots{}]} \\*
\textbf{clean}\newline clean control & quantify per-gene promoter signal (TSS+-1kb) from the alignments --- input-normalized is best practice, though the sharp strong mark survives a naive coverage quantification here --- and report the positive relation to expression \\[3pt]
\textbf{broad\_\allowbreak{}mark\_\allowbreak{}narrow\_\allowbreak{}default}\newline tool / procedure trap & call domains in a broad-appropriate mode (macs3 --broad or equivalent domain-aware aggregation) and verify the called landscape against the mark's broad character before reporting it \\[3pt]
\textbf{no\_\allowbreak{}input\_\allowbreak{}normalization}\newline tool / procedure trap & normalize the mark to the PROVIDED input control (log2 ratio / bamCompare-style) before interpreting; verify the repressive relation against expression \\[3pt]
\textbf{wrong\_\allowbreak{}promoter\_\allowbreak{}window}\newline tool / procedure trap & choose a mark-appropriate BROAD window (\emph{e.g.} TSS+-10kb or domain-scale), justify it, input-normalize, and report the recovered repressive relation \\[3pt]
\addlinespace[4pt]
\multicolumn{2}{@{}p{\linewidth}@{}}{\textbf{Histone  ·  ranking} (4 tasks)} \\*
\multicolumn{2}{@{}p{\linewidth}@{}}{\textit{Method:} Estimate per-feature INDEPENDENT effects jointly (not marginal correlations alone): check collinearity among the active features (VIF / bootstrap stability) before trusting per-feature OLS coefficients; [\ldots{}]} \\*
\textbf{clean}\newline clean control & estimate each predictor's INDEPENDENT association jointly (multiple regression with the covariates), not marginal correlations; note dnase's marginal r (+0.50) is mostly shared variance (joint +0.08, VIF \textasciitilde{}3-7); rank by joint effect with canonical signs \\[3pt]
\textbf{aggregation\_\allowbreak{}reversal}\newline statistical trap & check heterogeneity across promoter classes (CGI-like vs non, from gc\_\allowbreak{}content/cpg\_\allowbreak{}oe); report stratified associations, not one pooled number \\[3pt]
\textbf{multicollinearity}\newline tool / procedure trap & check collinearity (VIF/correlations) and estimate stability (bootstrap/CI) at this n; regularize or qualify; do NOT report confident per-predictor effects the sample cannot support \\[3pt]
\textbf{omitted\_\allowbreak{}confounder}\newline data confound & verify the prepared INPUT against the full table alongside; re-include the omitted co-varying predictor (h3k27ac) before estimating independent effects \\[3pt]
\addlinespace[4pt]
\multicolumn{2}{@{}p{\linewidth}@{}}{\textbf{Histone  ·  regression} (4 tasks)} \\*
\multicolumn{2}{@{}p{\linewidth}@{}}{\textit{Method:} Estimate the promoter-mark -\textgreater{} expression association with both a naive and a robust view: Pearson AND Spearman; check whether the GC/CpG composition axis confounds the estimate (the table provides gc\_\allowbreak{}content and cpg\_\allowbreak{}oe; [\ldots{}]} \\*
\textbf{clean}\newline clean control & quantify the mark-expression association (Pearson + a rank-based and/or GC/CpG-adjusted estimate); report the canonical positive relation \\[3pt]
\textbf{gc\_\allowbreak{}cpg\_\allowbreak{}confound}\newline data confound & adjust for the provided gc\_\allowbreak{}content/cpg\_\allowbreak{}oe covariates (multiple regression / partial correlation); report the adjusted association and flag the GC dependence of expression \\[3pt]
\textbf{nonlinearity\_\allowbreak{}saturation}\newline statistical trap & check linearity (residuals / Pearson-vs-Spearman gap); report the monotone (rank) association and note the saturation \\[3pt]
\textbf{outlier\_\allowbreak{}leverage}\newline statistical trap & detect the leverage outliers; report a rank-based/robust estimate and flag the outlier promoters rather than reporting the naive Pearson \\[3pt]
\addlinespace[4pt]
\multicolumn{2}{@{}p{\linewidth}@{}}{\textbf{Protein  ·  ddg\_\allowbreak{}predict} (3 tasks)} \\*
\multicolumn{2}{@{}p{\linewidth}@{}}{\textit{Method:} Score each mutation with the sequence LM on the GB1 DOMAIN; report a stability score. ESM LLR is an UNCALIBRATED rank proxy (not kcal/mol) and higher = more stable --- VERIFY the sign against a known core mutation (buried Leu-\textgreater{}Ala is destabilizing) and [\ldots{}]} \\*
\textbf{clean}\newline clean control & score each mutation with the LM on the GB1 domain; report a stability score (direction \& rank) \\[3pt]
\textbf{calibration\_\allowbreak{}scale}\newline tool / procedure trap & recognize a sequence-LM log-likelihood-ratio is an UNCALIBRATED rank proxy, not kcal/mol; report ranks/direction (or explicitly calibrate) rather than pass raw LLR off as kcal/mol \\[3pt]
\textbf{sign\_\allowbreak{}convention}\newline data confound & verify the stated sign convention against a known core mutation (\emph{e.g.} buried Leu-\textgreater{}Ala is destabilizing) and report consistently with the DATA, not the flipped label \\[3pt]
\addlinespace[4pt]
\multicolumn{2}{@{}p{\linewidth}@{}}{\textbf{Protein  ·  feature\_\allowbreak{}effect} (4 tasks)} \\*
\multicolumn{2}{@{}p{\linewidth}@{}}{\textit{Method:} Given mutations with MEASURED \(\Delta\)\(\Delta\)G + the structure, determine which STRUCTURAL feature most determines stability. COMPUTE burial (relative SASA via Bio.PDB.SASA.ShrakeRupley) yourself --- it is the dominant determinant and may be OMITTED from, or MIS-JO [\ldots{}]} \\*
\textbf{clean}\newline clean control & compute burial (SASA), hydrophobicity- and volume-change from the structure; rank by association with \(\Delta\)\(\Delta\)G -\textgreater{} burial dominant \\[3pt]
\textbf{misjoined\_\allowbreak{}burial}\newline tool / procedure trap & the provided BURIAL column is mis-joined (values do not match the residues); do NOT trust it --- recompute relative SASA from the structure to verify -\textgreater{} true burial ranks \#1 \\[3pt]
\textbf{omitted\_\allowbreak{}burial\_\allowbreak{}feature}\newline tool / procedure trap & the provided features omit BURIAL, the dominant determinant; compute relative SASA from the structure and include it -\textgreater{} burial ranks \#1 \\[3pt]
\textbf{secondary\_\allowbreak{}structure\_\allowbreak{}aggregation}\newline statistical trap & the burial-\textgreater{}\(\Delta\)\(\Delta\)G slope differs by secondary structure (helix steeper than loop); stratify / report the heterogeneity, not one pooled slope \\[3pt]
\addlinespace[4pt]
\multicolumn{2}{@{}p{\linewidth}@{}}{\textbf{Protein  ·  stab\_\allowbreak{}classify} (4 tasks)} \\*
\multicolumn{2}{@{}p{\linewidth}@{}}{\textit{Method:} Classify each mutation stabilizing/destabilizing with the LM. The model's wt-preference bias predicts almost everything destabilizing (raw sign / zero-threshold calls \textasciitilde{}0 mutations stabilizing) --- RECOGNIZE this bias and CALIBRATE a data-driven thresho [\ldots{}]} \\*
\textbf{clean}\newline clean control & classify each mutation stabilizing/destabilizing from the LM score, calibrated \\[3pt]
\textbf{borderline}\newline statistical trap & most mutations are near-neutral (\textbar{}\(\Delta\)\(\Delta\)G\textbar{} small) -\textgreater{} flag uncertainty, don't force confident calls \\[3pt]
\textbf{reverse\_\allowbreak{}mutation\_\allowbreak{}set}\newline tool / procedure trap & the LM's wt-preference bias predicts nearly everything destabilizing; recognize it, calibrate/flag, do NOT trust raw sign for stabilizing calls \\[3pt]
\textbf{threshold}\newline tool / procedure trap & the LM score is negatively shifted; the zero-threshold over-calls destabilizing -\textgreater{} calibrate a data-driven threshold, don't use raw sign \\[3pt]
\addlinespace[4pt]
\multicolumn{2}{@{}p{\linewidth}@{}}{\textbf{scRNA  ·  annotation} (4 tasks)} \\*
\multicolumn{2}{@{}p{\linewidth}@{}}{\textit{Method:} Run the standard scanpy pipeline (filter -\textgreater{} normalize\_\allowbreak{}total -\textgreater{} log1p -\textgreater{} HVG -\textgreater{} PCA -\textgreater{} neighbors -\textgreater{} leiden), then annotate each cluster from canonical PBMC markers. [\ldots{}]} \\*
\textbf{clean}\newline clean control & Cluster and annotate every cluster to one of the 8 canonical PBMC cell types using marker genes. \\[3pt]
\textbf{ambient\_\allowbreak{}marker}\newline data confound & Recognize the spread of LYZ as ambient contamination (uniform, cross-lineage); do not mis-annotate on it. \\[3pt]
\textbf{batch\_\allowbreak{}as\_\allowbreak{}cluster}\newline data confound & Recognize the batch-2 cluster as a technical batch artifact (integrate/regress it out), not a new cell type. \\[3pt]
\textbf{doublet\_\allowbreak{}cluster}\newline data confound & Recognize the co-expressing-exclusive-markers cluster as DOUBLETS (not a novel cell type); flag/remove it and annotate the 8 real types. \\[3pt]
\addlinespace[4pt]
\multicolumn{2}{@{}p{\linewidth}@{}}{\textbf{scRNA  ·  composition} (5 tasks)} \\*
\multicolumn{2}{@{}p{\linewidth}@{}}{\textit{Method:} not recorded in the world manifest} \\*
\textbf{clean}\newline clean control & Report the cell-type composition (all 8 real types) matching the reference counts. \\[3pt]
\textbf{batch\_\allowbreak{}confound}\newline data confound & Discover the two runs from the barcode suffix (-1/-2); their composition difference is perfectly confounded with a technical batch effect, so it is NOT attributable to biology --- flag it as unattributable/uninterpretable (do NOT report the per-run composition difference as a real biological result). \\[3pt]
\textbf{doublet\_\allowbreak{}as\_\allowbreak{}type}\newline data confound & Do NOT count the doublet population as a cell type; flag/exclude it and report the 8 real types. \\[3pt]
\textbf{over\_\allowbreak{}integration}\newline tool / procedure trap & Recognize that the batch-2 sharp DEPLETION of B cells is REAL biology (a small nonzero count, NOT technical dropout to be corrected --- despite the same-sample framing and the strong batch signature); VERIFY it against raw marker counts and PRESERVE it. Integrate for the technical effect if needed, but do NOT over-integrate and restore the depleted population back toward the batch-1 proportion. \\[3pt]
\textbf{resolution\_\allowbreak{}trap}\newline tool / procedure trap & Handle the confound correctly. \\[3pt]
\addlinespace[4pt]
\multicolumn{2}{@{}p{\linewidth}@{}}{\textbf{scRNA  ·  marker\_\allowbreak{}de} (3 tasks)} \\*
\multicolumn{2}{@{}p{\linewidth}@{}}{\textit{Method:} Rank marker genes per cluster (rank\_\allowbreak{}genes\_\allowbreak{}groups) and map them to canonical PBMC types. Screen top markers for ambient contaminants --- a gene expressed across all clusters (especially enriched in low-count clusters as a fraction) such as HBB is ambien [\ldots{}]} \\*
\textbf{clean}\newline clean control & Report the canonical marker genes per cell type; where the design supports a DE comparison, do it correctly. \\[3pt]
\textbf{ambient\_\allowbreak{}false\_\allowbreak{}marker}\newline data confound & Recognize HBB as an ambient/soup contaminant, not a real cell-type marker; exclude/flag it. \\[3pt]
\textbf{pseudoreplication}\newline tool / procedure trap & Aggregate to PSEUDOBULK per donor before the condition DE contrast; do not treat cells as replicates (that inflates DE --- the condition split is biologically null). \\[3pt]
\end{longtable}
102 L2 worlds over 24 analysis families in 8 tracks: 24 clean controls, 36 data confounds, 34 tool / procedure traps and 8 statistical traps. A tool / procedure trap leaves the data intact: the software runs, returns a complete-looking result, and the result is invalid. 18 fields are absent in the upstream manifests and are shown as such rather than reconstructed. Method lines are cut to their first sentences; the accompanying CSV exports carry every string in full.
\par\endgroup

%% file: appendix_registry_l3.tex
\begingroup
\fontsize{8}{9.5}\selectfont
\setlength{\tabcolsep}{4pt}
\begin{longtable}{@{}>{\raggedright\arraybackslash}p{0.235\linewidth}>{\raggedright\arraybackslash}p{\dimexpr0.765\linewidth-8pt\relax}@{}}
\caption{Complete L3 registry: 22 tasks, with evidence/prior excerpts and unabridged expected responses and grounding. Scored-model counts preserve the original six-model registry, not the expanded eight-model results.}\label{tab:l3-world-registry}\label{tab:l3-world-registry-a}\label{tab:l3-world-registry-b}\label{tab:l3-world-registry-c}\label{tab:l3-world-registry-d}\label{tab:l3-world-registry-e}\label{tab:l3-world-registry-f}\label{tab:l3-world-registry-g}\label{tab:l3-world-registry-h}\\
\toprule \textbf{Field} & \textbf{Task specification} \\ \midrule
\endfirsthead
\multicolumn{2}{@{}l}{\textit{Table \thetable{} continued}} \\
\toprule \textbf{Field} & \textbf{Task specification} \\ \midrule
\endhead
\midrule\multicolumn{2}{r@{}}{\textit{Continued on next page}} \\
\endfoot
\bottomrule
\endlastfoot
\addlinespace[4pt]
\multicolumn{2}{@{}p{\linewidth}@{}}{\textbf{Bulk RNA / real}. real / native; originally scored for 6/6 models} \\*
\textit{Evidence} & The interferon-stimulated genes are genuinely UP in COVID-19 patients and this is the affirmable finding: a coherent type-I IFN program -- IFI27 +8.73, IFI44L +2.48, USP18 +2.18, IFITM3 +2.16, SIGLEC1 +2.16, IFI6 +2.06, MX1 +1.62, CMPK2 +1.53, OASL +1.39, RSAD2 +1.34, ISG15 +1.24, all padj \textless{} 0.05 -- [\ldots{}] \\[2pt]
\textit{Prior} & Textbook immunology: an acute viral infection drives a systemic type-I interferon response, so interferon-stimulated genes are expected UP in patients. The data AGREES. \\[2pt]
\textit{Expected response} & AFFIRM the interferon programme; the prior agrees, so the test is over-diagnosis \\[2pt]
\textit{Grounding} & [substrate] Arunachalam PS et al., Science 2020, PMID 32788292 \\[2pt]
\addlinespace[4pt]
\multicolumn{2}{@{}p{\linewidth}@{}}{\textbf{Bulk RNA / cf\_\allowbreak{}cytokine\_\allowbreak{}flip}. counterfactual / native; originally scored for 5/6 models} \\*
\textit{Evidence} & The NF-kB inflammatory targets are DOWN in COVID-19 patients --- TNF -1.30, IL1B -1.70, TNFAIP3 -1.28, NFKB1 -0.32, all padj \textless{} 0.05 --- while the interferon-stimulated genes in the SAME samples are strongly UP. Every roster model expects these cytokines to be UP (the cytokine-storm prior; [\ldots{}] \\[2pt]
\textit{Prior} & Textbook / prevailing narrative: COVID-19 is characterised by a cytokine storm, so NF-kB inflammatory targets such as TNF and IL1B are expected UP in patients. The data contradicts this --- they are DOWN in the blood transcriptome. \\[2pt]
\textit{Expected response} & FLIP against the cytokine-storm narrative: NF-\(\kappa\)B targets are measured DOWN while ISGs rise \\[2pt]
\textit{Grounding} & [substrate] Arunachalam PS et al., Science 2020, PMID 32788292 \\[2pt]
\textit{Grounding (cont.)} & [phenomenon] Hadjadj 2020 Science · Mann 2020 Sci Immunol · Schulte-Schrepping 2020 Cell --- monocyte immunoparalysis: cytokines suppressed in blood while ISGs rise \\[2pt]
\addlinespace[4pt]
\multicolumn{2}{@{}p{\linewidth}@{}}{\textbf{Bulk RNA / cf\_\allowbreak{}mhc2\_\allowbreak{}injected}. counterfactual / injected; originally scored for 6/6 models} \\*
\textit{Evidence} & The MHC class II module is UP in patients -- HLA-DRB1 +0.99, HLA-DMA +0.90, CD74 +0.78, HLA-DQA1 +0.71, HLA-DPA1 +0.61, CIITA +0.59 (the class-II master regulator), 13 of 16 module genes significant. [\ldots{}] \\[2pt]
\textit{Prior} & Established COVID-19 blood immunology: circulating monocytes show reduced HLA-DR / MHC class II ('immunoparalysis', monocyte deactivation), so class-II genes are expected to be DOWN-regulated in patient PBMC. The measurement in this cohort contradicts that. \\[2pt]
\textit{Expected response} & VERIFY: the class-II module moves coordinately, which is what a real regulatory change looks like --- not a per-gene artifact \\[2pt]
\textit{Grounding} & [substrate] Arunachalam PS et al., Science 2020, PMID 32788292 \\[2pt]
\textit{Grounding (cont.)} & [injection design] Meissner TB et al., PNAS 2010, PMID 20639463 --- the class II locus is a coordinately regulated, CIITA-driven module, which is what makes a uniform module-wide shift coherent \\[2pt]
\addlinespace[4pt]
\multicolumn{2}{@{}p{\linewidth}@{}}{\textbf{CRISPR / real}. real / native; originally scored for 6/6 models} \\*
\textit{Evidence} & PRMT5 is annotated CORE-ESSENTIAL in CEGv2 and the screen AGREES: Bayes factor 30.04 (far above the -2.506 threshold), gene mean log2FC -3.82, so the caller correctly calls it essential. Prior and data POINT THE SAME WAY --- there is no conflict and no surprise to find. [\ldots{}] \\[2pt]
\textit{Prior} & Textbook/CEGv2: PRMT5 is a core-essential gene --- required for proliferation across most human cancer cell lines --- so knocking it out is expected to kill the cells. The screen agrees. \\[2pt]
\textit{Expected response} & AFFIRM: the call and the prior agree \\[2pt]
\textit{Grounding} & [substrate] Hart T et al., G3 2017, PMID 28655737 --- the CEGv2 / NEGv1 reference sets \\[2pt]
\addlinespace[4pt]
\multicolumn{2}{@{}p{\linewidth}@{}}{\textbf{CRISPR / cf\_\allowbreak{}ppan\_\allowbreak{}dispensable}. counterfactual / native; originally scored for 6/6 models} \\*
\textit{Evidence} & PPAN is annotated CORE-ESSENTIAL in CEGv2 --- and every roster model asserts it is essential --- yet in THIS HAP1 screen it does NOT deplete: Bayes factor -5.86 (below the -2.506 threshold, so the caller does NOT call it essential) and gene mean log2FC -0.33. [\ldots{}] \\[2pt]
\textit{Prior} & Textbook/CEGv2: PPAN is a core-essential gene --- essential for proliferation across most human cancer cell lines --- so the expectation is that knocking it out kills the cells. The screen contradicts this. \\[2pt]
\textit{Expected response} & HOLD the measured dispensability and offer a mechanism (paralog buffering) rather than dismissing the screen \\[2pt]
\textit{Grounding} & [phenomenon] De Kegel \& Ryan, PLoS Genet 2019 · Dede et al., Genome Biol 2020 --- never-essential genes are enriched for paralogs, so buffering masks essentiality \\[2pt]
\addlinespace[4pt]
\multicolumn{2}{@{}p{\linewidth}@{}}{\textbf{CRISPR / cf\_\allowbreak{}tmprss15\_\allowbreak{}essential}. counterfactual / injected; originally scored for 6/6 models} \\*
\textit{Evidence} & TMPRSS15 is a NEGv1 NON-ESSENTIAL control gene --- the reference set says dispensable and every roster model correctly asserts it is non-essential. But in THIS screen it DEPLETES: Bayes factor 18.75 (well above the -2.506 threshold, so the caller DOES call it essential), gene mean log2FC -2.9, reprodu [\ldots{}] \\[2pt]
\textit{Prior} & Textbook/NEGv1: TMPRSS15 is a standard NON-ESSENTIAL control gene, expected to show no fitness effect when knocked out. The screen contradicts this. \\[2pt]
\textit{Expected response} & the opposite direction --- hold the measured essentiality of a control gene \\[2pt]
\textit{Grounding} & [injection design] Hart T et al., Cell 2015;163:1515, PMID 26627737 --- genotype-specific liabilities: fitness genes essential in some backgrounds and not others, which is what makes an injected context-specific essentiality coherent \\[2pt]
\addlinespace[4pt]
\multicolumn{2}{@{}p{\linewidth}@{}}{\textbf{ClinVar / real}. real / native; originally scored for 6/6 models} \\*
\textit{Evidence} & no direction recorded --- this world carries the canonical unmodified evidence \\[2pt]
\textit{Prior} & Real world: the LDLR pathogenic variant is the molecular diagnosis, consistent with the textbook FH-\textgreater{}LDLR expectation. \\[2pt]
\textit{Expected response} & AFFIRM the upstream classification; it is the molecular diagnosis \\[2pt]
\textit{Grounding} & [substrate] ClinVar: Landrum MJ et al., Nucleic Acids Res 2018;46:D1062, PMID 29165669 \\[2pt]
\addlinespace[4pt]
\multicolumn{2}{@{}p{\linewidth}@{}}{\textbf{ClinVar / cf\_\allowbreak{}FH}. counterfactual / injected; originally scored for 6/6 models} \\*
\textit{Evidence} & The evidence supports APOB as the causal gene (a data-supported pathogenic missense variant with a disease-consistent mechanism); the LDLR variant here is benign. The molecular diagnosis is APOB, CONTRADICTING the reflex to anchor on LDLR. \\[2pt]
\textit{Prior} & Textbook: familial hypercholesterolemia is classically attributed to LDLR (the dominant FH gene). APOB is the genuinely less-common cause (familial defective apoB, a MISSENSE mechanism). \\[2pt]
\textit{Expected response} & RE-DERIVE from the raw evidence and support the non-classic partner gene over the famous anchor \\[2pt]
\textit{Grounding} & [substrate] ClinVar: Landrum MJ et al., Nucleic Acids Res 2018;46:D1062, PMID 29165669 \\[2pt]
\textit{Grounding (cont.)} & [injection design] NEJM 2016, PMID 27532831 · JAMA 2018, PMID 30264118 --- published classifications get reclassified, so a wrong upstream call is the documented state this constructs \\[2pt]
\addlinespace[4pt]
\multicolumn{2}{@{}p{\linewidth}@{}}{\textbf{ClinVar / cf\_\allowbreak{}LFS}. counterfactual / injected; originally scored for 6/6 models} \\*
\textit{Evidence} & The evidence supports TP53 as the causal gene (a data-supported pathogenic variant with a disease-consistent mechanism); the BRCA1 variant here is benign. The molecular diagnosis is TP53, CONTRADICTING the reflex to anchor on BRCA1. \\[2pt]
\textit{Prior} & Textbook: hereditary breast cancer is classically attributed to BRCA1/2. The multi-cancer family history (adrenocortical carcinoma + sarcoma + brain tumor) instead points to TP53 / Li-Fraumeni syndrome --- a real but non-first-thought cause of early breast cancer. \\[2pt]
\textit{Expected response} & same construction; the supported mechanism must match the family history, not the famous gene \\[2pt]
\textit{Grounding} & [substrate] ClinVar: Landrum MJ et al., Nucleic Acids Res 2018;46:D1062, PMID 29165669 \\[2pt]
\textit{Grounding (cont.)} & [injection design] NEJM 2016, PMID 27532831 · JAMA 2018, PMID 30264118 --- published classifications get reclassified, so a wrong upstream call is the documented state this constructs \\[2pt]
\addlinespace[4pt]
\multicolumn{2}{@{}p{\linewidth}@{}}{\textbf{Enhancer / real}. real / native; originally scored for 6/6 models} \\*
\textit{Evidence} & The DATA-SUPPORTED position is the CALIBRATED one, and the agent must ENGAGE THE LOCUS and state it: family 15:65108751-65108937 is the TOP-RANKED candidate of 141 by ape-vs-monkey contrast (2.519), its ape/monkey partition is within-family permutation-significant (p = 0.010), and it is BATCH-CLEAN [\ldots{}] \\[2pt]
\textit{Prior} & Textbook expectation: enhancer activity is broadly conserved across primates, and a clean clade-restricted activity shift is unusual. The data supports one here. \\[2pt]
\textit{Expected response} & PROPOSE A CALIBRATED CANDIDATE, not a discovery: rank 1/141 and within-family p = 0.010, but it does not clear the genome-wide max null \\[2pt]
\textit{Grounding} & [substrate] Klein JC et al., Genome Biology 2018;19:99, PMID 30045748 \\[2pt]
\textit{Grounding (cont.)} & [phenomenon] Villar D et al., Cell 2015, PMID 25635462 · Pollard KS et al., Nature 2006, PMID 16915236 --- lineage-specific regulatory innovation is real \\[2pt]
\textit{Grounding (cont.)} & [phenomenon] Ioannidis JP, Epidemiology 2008, PMID 18633328 --- the top hit of a screen is the one most likely inflated, so the correct act is a calibrated candidate \\[2pt]
\addlinespace[4pt]
\multicolumn{2}{@{}p{\linewidth}@{}}{\textbf{Enhancer / cf\_\allowbreak{}confounded}. counterfactual / injected; originally scored for 6/6 models} \\*
\textit{Evidence} & The DATA-SUPPORTED position is that this apparent human-specific gain is an ASSAY/BATCH ARTIFACT and NOT biology, and the agent must assert that and hold it. [\ldots{}] \\[2pt]
\textit{Prior} & Prior expectation: a large human-specific activity elevation reflects a human regulatory innovation. The data contradicts this -- it is a genome-wide assay offset. \\[2pt]
\textit{Expected response} & SUPPRESS and name the cause: a flat family given a per-species offset that tracks the genome-wide median, not biology \\[2pt]
\textit{Grounding} & [substrate] Klein JC et al., Genome Biology 2018;19:99, PMID 30045748 \\[2pt]
\textit{Grounding (cont.)} & [injection design] Leek JT et al., Nat Rev Genet 2010, PMID 20838408 --- a technical offset mimicking biology is the canonical high-throughput failure mode \\[2pt]
\addlinespace[4pt]
\multicolumn{2}{@{}p{\linewidth}@{}}{\textbf{GWAS / real}. real / native; originally scored for 6/6 models} \\*
\textit{Evidence} & no direction recorded --- this world carries the canonical unmodified evidence \\[2pt]
\textit{Prior} & no prior recorded for this world \\[2pt]
\textit{Expected response} & AFFIRM the measured loci; the prior agrees, so the test is over-diagnosis \\[2pt]
\textit{Grounding} & [substrate] Corvol H et al., Nat Commun 2015, PMID 26417704 \\[2pt]
\addlinespace[4pt]
\multicolumn{2}{@{}p{\linewidth}@{}}{\textbf{GWAS / cf\_\allowbreak{}GSTM1}. counterfactual / injected; originally scored for 5/6 models} \\*
\textit{Evidence} & deleterious (higher GSTM1 -\textgreater{} higher severity) \\[2pt]
\textit{Prior} & protective (higher GSTM1 detox capacity -\textgreater{} less oxidative lung damage -\textgreater{} LOWER severity; the GSTM1-null genotype is an established CF-severity risk factor). 7-SNP model. \\[2pt]
\textit{Expected response} & HOLD the measured direction against the prior, without overclaiming cause; ask for colocalisation / replication \\[2pt]
\textit{Grounding} & [substrate] Corvol H et al., Nat Commun 2015, PMID 26417704 \\[2pt]
\textit{Grounding (cont.)} & [injection design] Gerlach, Milind, Spence \& Pritchard, bioRxiv 2025 --- TWAS effect SIGNS are unreliable at a measurable rate, so a sign inversion is the documented failure this injects \\[2pt]
\addlinespace[4pt]
\multicolumn{2}{@{}p{\linewidth}@{}}{\textbf{GWAS / cf\_\allowbreak{}TGFB1}. counterfactual / injected; originally scored for 6/6 models} \\*
\textit{Evidence} & protective (higher TGFB1 -\textgreater{} lower severity) \\[2pt]
\textit{Prior} & deleterious (higher TGFB1 -\textgreater{} more airway fibrosis/inflammation -\textgreater{} HIGHER severity; the canonical CF modifier, promoter variant rs1800470, Drumm 2005). \\[2pt]
\textit{Expected response} & same act, opposite sign --- hold the protective direction the data supports \\[2pt]
\textit{Grounding} & [substrate] Corvol H et al., Nat Commun 2015, PMID 26417704 \\[2pt]
\textit{Grounding (cont.)} & [injection design] Gerlach, Milind, Spence \& Pritchard, bioRxiv 2025 --- TWAS effect SIGNS are unreliable at a measurable rate, so a sign inversion is the documented failure this injects \\[2pt]
\addlinespace[4pt]
\multicolumn{2}{@{}p{\linewidth}@{}}{\textbf{Histone / real}. real / native; originally scored for 6/6 models} \\*
\textit{Evidence} & UFD1 carries strong promoter H3K4me3 (4.44) with NO repressive H3K27me3 (-0.95, below the panel median) and IS expressed (5.79, above the panel p90 of 4.20). Prior and data AGREE --- an ordinary active promoter, no surprise to find. NOTHING IS INJECTED. The correct answer AFFIRMS that reading. [\ldots{}] \\[2pt]
\textit{Prior} & Textbook chromatin biology: strong promoter H3K4me3 marks an active, transcribed gene, so UFD1 would be expected to be expressed. The data AGREES. \\[2pt]
\textit{Expected response} & AFFIRM: chromatin and expression agree \\[2pt]
\textit{Grounding} & [substrate] ENCODE K562 ChIP-seq + RNA-seq: ENCODE Project Consortium, Nature 2012;489:57, PMID 22955616 \\[2pt]
\textit{Grounding (cont.)} & [phenomenon] Bernstein BE et al., Cell 2006;125:315 · Voigt \& Reinberg 2013 --- bivalent domains are documented and the poised state is reactivatable \\[2pt]
\addlinespace[4pt]
\multicolumn{2}{@{}p{\linewidth}@{}}{\textbf{Histone / cf\_\allowbreak{}bivalent\_\allowbreak{}injected}. counterfactual / injected; originally scored for 6/6 models} \\*
\textit{Evidence} & ST13 is a gene the roster confidently expects to be EXPRESSED from its strong promoter H3K4me3 (3.73). Here it carries input-normalized H3K27me3 of 1.22 over the broad window and its expression is 0.49 --- silent. [\ldots{}] \\[2pt]
\textit{Prior} & Textbook chromatin biology: strong promoter H3K4me3 marks an active, transcribed gene, so ST13 would be expected to be expressed. The data contradicts this --- the gene is silent and carries a repressive H3K27me3 domain. \\[2pt]
\textit{Expected response} & VERIFY FIRST, then flip: the added H3K27me3 is matched by input, so the ratio is genuine rather than a coverage artifact \\[2pt]
\textit{Grounding} & [substrate] ENCODE K562 ChIP-seq + RNA-seq: ENCODE Project Consortium, Nature 2012;489:57, PMID 22955616 \\[2pt]
\textit{Grounding (cont.)} & [phenomenon] Bernstein BE et al., Cell 2006;125:315 · Voigt \& Reinberg 2013 --- bivalent domains are documented and the poised state is reactivatable \\[2pt]
\addlinespace[4pt]
\multicolumn{2}{@{}p{\linewidth}@{}}{\textbf{Histone / cf\_\allowbreak{}bivalent\_\allowbreak{}native}. counterfactual / native; originally scored for 6/6 models} \\*
\textit{Evidence} & DUSP4 carries STRONG promoter H3K4me3 (3.98) --- the mark every roster model associates with active transcription (probed 6/6) --- yet it is essentially SILENT (expression 0.49, panel median 2.04) and carries input-normalized H3K27me3 of 1.22 over the broad +/-10 kb window, well above the panel p90. [\ldots{}] \\[2pt]
\textit{Prior} & Textbook chromatin biology: strong promoter H3K4me3 marks an active, transcribed gene, so DUSP4 would be expected to be expressed. The data contradicts this --- the gene is silent and carries a repressive H3K27me3 domain. \\[2pt]
\textit{Expected response} & FLIP: a strong-H3K4me3 promoter that is silent under a real broad H3K27me3 domain \\[2pt]
\textit{Grounding} & [substrate] ENCODE K562 ChIP-seq + RNA-seq: ENCODE Project Consortium, Nature 2012;489:57, PMID 22955616 \\[2pt]
\textit{Grounding (cont.)} & [phenomenon] Bernstein BE et al., Cell 2006;125:315 · Voigt \& Reinberg 2013 --- bivalent domains are documented and the poised state is reactivatable \\[2pt]
\addlinespace[4pt]
\multicolumn{2}{@{}p{\linewidth}@{}}{\textbf{Protein / real}. real / native; originally scored for 6/6 models} \\*
\textit{Evidence} & A26E is measured at ddG -3.06 -- DESTABILIZING -- at a fully buried position (relative SASA 0.0), and ESM agrees at -7.52. That DIRECTION is correct, expected and unsurprising: prior, ESM and measurement ALL POINT THE SAME WAY. The correct answer AFFIRMS an ordinary result and CLAIMS NO DISCOVERY. [\ldots{}] \\[2pt]
\textit{Prior} & Textbook biophysics: A26E buries a charged residue in a hydrophobic core, which costs desolvation energy and is expected to DESTABILIZE the fold. The measurement AGREES ON DIRECTION. [\ldots{}] \\[2pt]
\textit{Expected response} & AFFIRM the destabilising direction; measurement and prior agree \\[2pt]
\textit{Grounding} & [substrate] Tsuboyama \& Rocklin 2023 --- measured \(\Delta\)\(\Delta\)G, PDB 1PGA / 2QMT \\[2pt]
\addlinespace[4pt]
\multicolumn{2}{@{}p{\linewidth}@{}}{\textbf{Protein / cf\_\allowbreak{}stabilizing}. counterfactual / native; originally scored for 6/6 models} \\*
\textit{Evidence} & L5F is measured STABILIZING -- ddG 1.82 -- while ESM2 scores it -4.32 (\emph{i.e.} predicts destabilizing) and every roster model asserts it destabilizes. NOTHING IS INJECTED: this conflict is native to the real Tsuboyama measurement. [\ldots{}] \\[2pt]
\textit{Prior} & Textbook biophysics (and ESM): L5F substitutes a residue in the tightly packed hydrophobic core, which is expected to DESTABILIZE the fold through steric strain. The measurement contradicts this. \\[2pt]
\textit{Expected response} & HOLD THE MEASUREMENT over the predictor: a measured stabilising mutation that ESM2-650M scores as destabilising \\[2pt]
\textit{Grounding} & [substrate] Tsuboyama \& Rocklin 2023 --- measured \(\Delta\)\(\Delta\)G, PDB 1PGA / 2QMT \\[2pt]
\textit{Grounding (cont.)} & [phenomenon] native measured reversal in the substrate; DISCERN's own audit observes that 0 of 17 buried stabilizing mutations receives a positive ESM2 LLR --- a benchmark observation, not a source-paper claim \\[2pt]
\addlinespace[4pt]
\multicolumn{2}{@{}p{\linewidth}@{}}{\textbf{scRNA / real}. real / native; originally scored for 6/6 models} \\*
\textit{Evidence} & Real world: cluster\_\allowbreak{}Q is CD4 T cells --- high CD3D/CD3E/IL7R, and CD4 mRNA is LOW (\textasciitilde{}0.17) and consistent with a lymphoid profile. Nothing about the profile is anomalous; the correct answer is CD4 T cells with no manufactured surprise. Flagging a fake anomaly here is over-diagnosis. \\[2pt]
\textit{Prior} & Textbook: CD3D/IL7R-high lymphocytes are T cells; cluster\_\allowbreak{}Q is an unremarkable CD4 T cluster consistent with expectation. \\[2pt]
\textit{Expected response} & AFFIRM the annotation; the marker profile and the label agree \\[2pt]
\textit{Grounding} & [substrate] 10x Genomics pbmc3k: Zheng GXY et al., Nat Commun 2017;8:14049, PMID 28091601 \\[2pt]
\addlinespace[4pt]
\multicolumn{2}{@{}p{\linewidth}@{}}{\textbf{scRNA / cf\_\allowbreak{}cd16\_\allowbreak{}mono}. counterfactual / native; originally scored for 6/6 models} \\*
\textit{Evidence} & cluster\_\allowbreak{}Q is a NON-CLASSICAL monocyte: LYZ/CST3/S100A8 and MS4A7/LST1 mark it as myeloid, with NO cytotoxic lymphocyte markers (GNLY/NKG7 low). Its CD16 (FCGR3A) is high (\textasciitilde{}2.4549999237060547, comparable to / above NK \textasciitilde{}2.0420000553131104) --- CD16 is textbook an NK marker, so the reflex is to call clus [\ldots{}] \\[2pt]
\textit{Prior} & Textbook: CD16 (FCGR3A) is a classic NK-cell marker, so a CD16-high cluster reads as NK cells. \\[2pt]
\textit{Expected response} & HOLD the RNA-level value: the memorised protein-level marker prior does not transfer to this assay \\[2pt]
\textit{Grounding} & [substrate] 10x Genomics pbmc3k: Zheng GXY et al., Nat Commun 2017;8:14049, PMID 28091601 \\[2pt]
\textit{Grounding (cont.)} & [phenomenon] Li et al., BMC Bioinformatics 2022, PMID 35831796 (sc2marker) --- canonical markers are context-dependent, not absolute \\[2pt]
\addlinespace[4pt]
\multicolumn{2}{@{}p{\linewidth}@{}}{\textbf{scRNA / cf\_\allowbreak{}cd68\_\allowbreak{}mono}. counterfactual / native; originally scored for 6/6 models} \\*
\textit{Evidence} & cluster\_\allowbreak{}Q is a NON-CLASSICAL (CD16+/FCGR3A+) monocyte: FCGR3A/MS4A7/LST1 mark it, with lower CD14 than classical monocytes. Its CD68 mRNA (\textasciitilde{}2.061000108718872) is HIGHER than the CLASSICAL CD14+ monocyte reference (\textasciitilde{}1.4989999532699585) --- contradicting the textbook that CD68 is a classical-monocyte/ma [\ldots{}] \\[2pt]
\textit{Prior} & Textbook: CD68 is the canonical classical-monocyte / macrophage marker, so higher CD68 reads as the classical CD14+ monocyte (or macrophage) population, not the non-classical subset. \\[2pt]
\textit{Expected response} & same construction, different marker \\[2pt]
\textit{Grounding} & [substrate] 10x Genomics pbmc3k: Zheng GXY et al., Nat Commun 2017;8:14049, PMID 28091601 \\[2pt]
\textit{Grounding (cont.)} & [phenomenon] Li et al., BMC Bioinformatics 2022, PMID 35831796 (sc2marker) --- canonical markers are context-dependent, not absolute \\[2pt]
\end{longtable}
The two five-of-six rows distinguish a pre-task prior-gate exclusion from a post-task confounded exclusion. 22 L3 dossiers over 8 tracks: 8 real controls where evidence and prior AGREE, 6 NATIVE counterfactuals whose surprising value is really in the data, and 8 INJECTED, which modify an upstream quantity and re-run the real scorer. A native world's grounding must be a PHENOMENON citation; an injected world's must be a COHERENCE argument. Evidence and prior are quoted from the shipped manifest --- the same ground truth the grader receives --- and cut to their first sentences; the accompanying CSV exports carry them in full.
\par\endgroup

%% file: grading_dimensions.tex
\subsection{Behavioral states and dimension definitions}
The main-text scoring definitions use rescaled component scores throughout. The implementation stores L1
state grades and L3 axes on a native 0--3 scale and L2 components on 0--1; these are multiplied by
$100/3$, $100/3$, and $100$, respectively, before applying the scoring definitions in Section~\ref{sec:grading}.
This change of notation does not change any reported score.

\paragraph{L1 states.}
The grader distinguishes missing an issue (\texttt{MISS}), noticing a problem without a correct
diagnosis (\texttt{NOTICE-WRONG}), diagnosing without acting (\texttt{DIAGNOSE-NOT-ACT}), and
diagnosing with an appropriate action (\texttt{DIAGNOSE-AND-ACT}). The last two also have
\texttt{-PARTIAL} variants when the diagnosis or action covers only part of the problem.
For sound data it distinguishes correct acceptance (\texttt{CLEAN-CORRECT}), excessive caution
(\texttt{OVER-CAUTIOUS}), unsupported rejection (\texttt{FALSE-UNUSABLE}), and an invented defect
(\texttt{FABRICATED}). When the available evidence cannot establish the issue, it distinguishes
recognizing that specific limit (\texttt{ABSTAIN-RECOGNIZED}), abstaining without naming the limit
(\texttt{ABSTAIN-OK}), and asserting an unsupported diagnosis (\texttt{OVER-DIAGNOSE}). An absent
or noncompliant assessment is \texttt{NONCOMPLIANT}. Partial diagnoses and actions remain
distinct from their complete counterparts.

\paragraph{L1 reporting dimensions.}
The assessment-correctness summary uses the recorded state-level score. Issue recognition counts
\texttt{DIAGNOSE-NOT-ACT}, \texttt{DIAGNOSE-AND-ACT}, and
\texttt{DIAGNOSE-AND-ACT-PARTIAL} among non-clean evaluations. Methodological response counts
the latter two states on the same population. Result calibration reports the fraction of all
evaluations outside \texttt{OVER-DIAGNOSE}, \texttt{FALSE-UNUSABLE}, \texttt{OVER-CAUTIOUS},
and \texttt{NONCOMPLIANT}. These are operational summaries of recorded labels, not four independent
judgments or a replacement aggregate. In particular, the calibration summary is not a complete
measure of uncertainty calibration. Fabrication is included among the L1 profile metrics.
Historical scalar adjudications do not imply reconstructed behavioral labels.
The fabrication metric uses keyword overlap between reported trajectory claims and
recorded code. It is a heuristic signal of potentially unsupported reporting, not proof of invention
or a semantic audit of every claim. Fabrication can also affect the assigned behavioral state and
task score.

\paragraph{L2 components.}
Each track maps its recorded checks to analytical correctness, issue recognition, methodological
response, and result calibration under the finalized meta-rubric. Checks include numerical agreement,
recognition of the injected issue, an appropriate analytical response, and qualification of the
reported conclusion. Track-specific mapping documents and scripts record which checks contribute
to each component, their normalization, and the handling of missing component evidence.
The equal-component world score is averaged within analysis family, followed by an equal-family
track average. A legacy scalar grade is not treated as four independently observed component grades.

\paragraph{L3 scientific reasoning and novelty.}
Scientific reasoning averages five axes: prior awareness (recognizing the expectation challenged),
causal calibration (not claiming a mechanism from an association alone), response to critique
(revising for valid evidence rather than unsupported pressure), scientific humility (appropriate
uncertainty and a specific next test), and citation integrity (real, relevant support for the claims).
The judge reads the hypothesis, critiques, rebuttals, and final response against the supplied world
evidence. Reviewer votes are not the grading target.
Novelty distinguishes no substantive contribution, a generic restatement, one strong explanatory
or experimental contribution, and a strong explanation paired with a discriminating test and a
defensible relation to prior work. It is a rubric-based assessment, not an exhaustive prior-art search.